\documentclass[journal,twoside]{IEEEtran}

\ifCLASSINFOpdf
  \usepackage[pdftex]{graphicx}
  \graphicspath{{figures/}}
  \DeclareGraphicsExtensions{.pdf,.jpeg,.png}
\else
  \usepackage[dvips]{graphicx}
  \graphicspath{{figures/}}
  \DeclareGraphicsExtensions{.eps}
\fi

\usepackage{diagbox}
\usepackage{multirow}

\usepackage{cite}

\usepackage{amsmath}
\usepackage{amsfonts}

\usepackage{algorithm}
\usepackage{algorithmic}

\usepackage{array}

\ifCLASSOPTIONcompsoc
  \usepackage[caption=false,font=normalsize,labelfont=sf,textfont=sf]{subfig}
\else
  \usepackage[caption=false,font=footnotesize]{subfig}
\fi
\usepackage{float}

\usepackage{threeparttable}

\begin{document}
\title{A Robust Visual System for Small Target Motion Detection Against Cluttered Moving Backgrounds}

\author{Hongxin Wang, Jigen Peng, Xuqiang Zheng and Shigang Yue, \IEEEmembership{Senior Member,~IEEE} \thanks{Manuscript received August 21, 2018; revised January 14, 2019; accepted April 5, 2019. This work was supported in part by EU HORIZON 2020 Project STEP2DYNA under Grant 691154, in part by EU HORIZON 2020 Project ULTRACEPT under Grant 778062, and in part by the National Natural Science Foundation of China under Grant 11771347. \emph{(Hongxin Wang and Jigen Peng contributed equally to this work.)}\emph{(Corresponding author: Shigang Yue.)}}
	\thanks{H. Wang and S. Yue are with the Machine Life and Intelligence Research Center, Guangzhou University, Guangzhou 510006, China, and also with the Computational Intelligence Lab, School of Computer Science, University of Lincoln, Lincoln LN6 7TS, U.K. (email: syue@lincoln.ac.uk).}
	\thanks{J. Peng is with the School of Mathematics and Information Science, Guangzhou University, Guangzhou 510006, China (email: jgpeng@gzhu.edu.cn).}
	\thanks{X. Zheng is with the Institute of Microelectronics of the Chinese Academy of Sciences, Beijing 100029, China.}
	\thanks{Color versions of one or more figures in this paper are available at https://ieeexplore.ieee.org.}
	\thanks{Digital Object Identifier 10.1109/TNNLS.2019.2910418}
}
   

\markboth{IEEE TRANSACTIONS ON NEURAL NETWORKS AND LEARNING SYSTEMS}
{Wang \MakeLowercase{\textit{et al.}}: A Robust Visual System for Small Target Motion Detection}
%



\maketitle

\begin{abstract}
	Monitoring small objects against cluttered moving backgrounds is a huge challenge to future robotic vision systems. As a source of inspiration, insects are quite apt at searching for mates and tracking prey -- which always appear as small dim speckles in the visual field. The exquisite sensitivity of insects for small target motion, as revealed recently, is coming from a class of specific neurons called small target motion detectors (STMDs). Although a few STMD-based models have been proposed, these existing models only use motion information for small target detection and cannot discriminate small targets from small-target-like background features (named as fake features). To address this problem, this paper proposes a novel visual system model (STMD+) for small target motion detection, which is composed of four subsystems -- ommatidia, motion pathway, contrast pathway and mushroom body. Compared to existing STMD-based models, the additional contrast pathway extracts directional contrast from luminance signals to eliminate false positive background motion. The directional contrast and the extracted motion information by the motion pathway are integrated in the mushroom body for small target discrimination. Extensive experiments showed the significant and consistent improvements of the proposed visual system model over existing STMD-based models against fake features.	
\end{abstract}

\begin{IEEEkeywords}
Visual system model, neural modeling, small target motion detector (STMD), cluttered natural environment, background motion.
\end{IEEEkeywords}

%
\IEEEpeerreviewmaketitle

\section{Introduction}

\IEEEPARstart{T}{he} dynamic visual world is often complex, with many motion cues at different speeds, directions, distances and orientations, exhibiting various physical characteristics such as size, colour, texture and shape \cite{yue2006bio,yue2006collision,wozniak2018adaptive,Yan2018AFast,yan2018effective,yan2018supervised}. Being able to detect target motion in the distance and early would put an entity (a robot or an animal) in a good position to prepare for interaction/competition, for example, a flying insect searching for mates in the distance. In the visual world, detecting visual motion in the distance and early often means dealing with small targets with only one or a few pixels in size let alone other physical characteristics. Small target motion detection has a wide variety of applications in defences, surveillance, security and road safety. However, detecting small targets against cluttered moving backgrounds is always a challenge for artificial visual systems due to limited physical cues of small targets, free motion of camera, and extremely cluttered backgrounds. 

How to detect small target motion in cluttered moving backgrounds robustly with limited resources? Research in insects' visual system have revealed one effective solution. Insects show exquisite sensitivity for small target motion \cite{barnett2007retinotopic} and are able to pursue small flying targets with high capture rates \cite{mischiati2015internal}. Biological research demonstrates that a class of specific neurons, called small target motion detectors (STMDs), can account for insects' exquisite sensitivity for small target motion \cite{kelecs2017object,nordstrom2006insect,barnett2007retinotopic}. These STMD neurons give peak responses to small targets subtending $1-3^{\circ}$ of the visual field, with no response to large bars (typically $>10^{\circ}$) or to background movements represented by wide-field  grating stimuli \cite{nordstrom2012neural}. Build a quantitative STMD model is the first step for not only further understanding of the biological visual system but also providing robust and economic solutions of small target detection for an artificial vision system.

\begin{figure}[t!]
	\centering
	\includegraphics[width=0.35\textwidth]{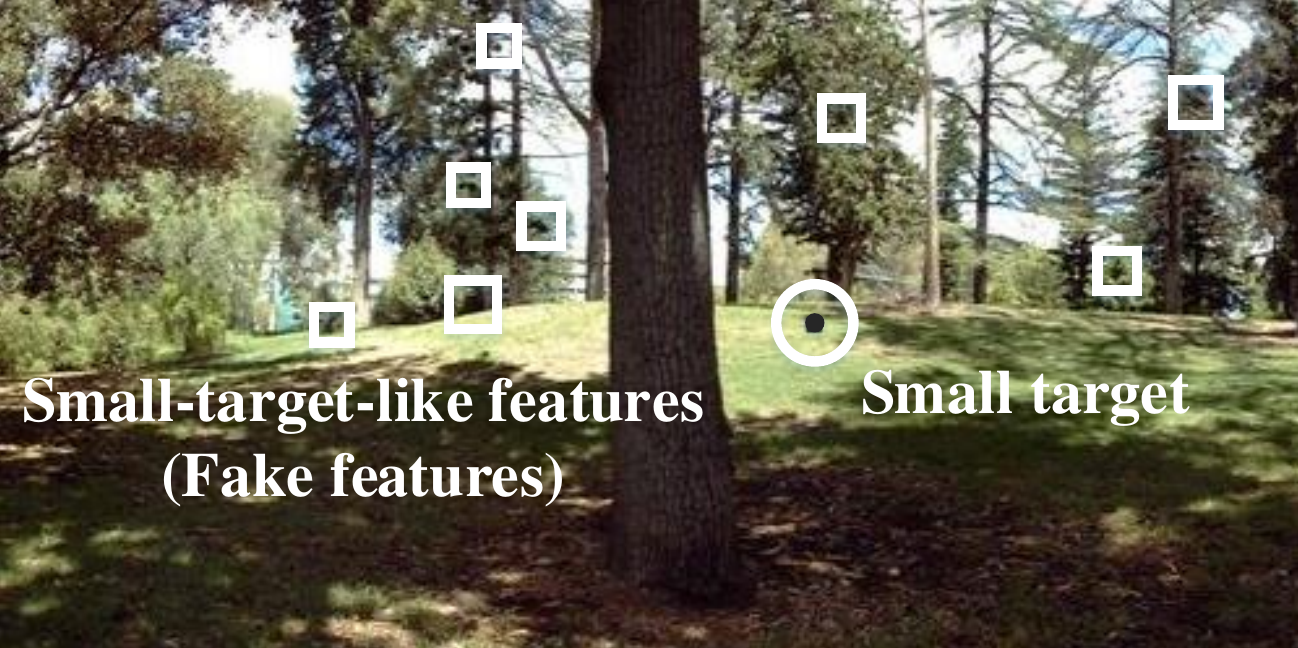}
	\caption{A small target is moving in the cluttered natural background which contains a number of small-target-like features (or called fake features). The small target and fake features all appear as small dim speckles whose sizes vary from one pixel to a few pixels, since they are far away from the animal/camera.}
	\label{Introduction-Schematic-Illustration-of-Small-Target-and-Background-Noise}
\end{figure}

The electrophysiological knowledge about the STMD neurons revealed in the past few decades, makes it possible to propose quantitative models, such as elementary small target motion detector (ESTMD) \cite{wiederman2008model} and directionally selective small target motion detector (DSTMD) \cite{wang2018directionally}. Using motion information\footnote{Motion information refers to luminance changes of a pixel with respect to time. From the view of mathematics, it is equivalent to temporal derivative of a pixel.} extracted by large monopolar cells (LMCs) \cite{freifeld2013gabaergic,behnia2014processing}, these models are able to detect small moving targets in cluttered backgrounds. However, they cannot discriminate small moving targets from small-target-like background features (as shown in Fig. \ref{Introduction-Schematic-Illustration-of-Small-Target-and-Background-Noise}), which means that their detection results may contain a large number of false positives. This is because (1) small-target-like background features are embedded in the cluttered background such as bushes, trees and/or rocks, (2) they are moving with the whole background due to a free flying animal/camera.  In this case, these small-target-like features (named as fake features) cannot be simply filtered out by existing STMD-based models with {\bf{motion information only}} for small target motion detection. To address this problem, other visual information, such as directional contrast\footnote{Directional contrast denotes luminance changes of a pixel along different spatial directions. From the view of mathematics, it corresponds to directional derivatives of a pixel.}, should be combined with motion information for distinguishing small targets from fake features.

In the insects' visual systems, multiple visual cues are extracted by different specialized neural circuits \cite{fu2018shaping,Hu2016A,yue2013redundant}. Multiple neural circuits could be coordinated to discriminate small target motion. For example, in the lamina layer, large monopolar cells (LMCs) \cite{freifeld2013gabaergic,behnia2014processing} have been described as temporal band-pass filters which extract motion information from luminance signals \cite{wiederman2008model,wang2018directionally,bagheri2017performance}; and amacrine cells (AMCs) \cite{St2016Adaptations,Riveraalvidrez2011A,lessios2018multiple} linked to multi adjacent ommatidia with thin extending fibers, may constitute a contrast pathway with their downstream neurons to extract directional contrast from luminance signals. Although the contribution from the AMCs to STMD neural circuits in insects is unknown, it is clear that with directional contrast and motion information together, an artificial vision system could discriminate small moving targets from fake features robustly.

Inspired by the above biological findings, this paper proposes a new visual system model ({\bfseries{STMD+}}) to detect small target motion in cluttered moving backgrounds. The main contribution of this work is combining motion information with directional contrast to successfully discriminate small targets from fake features. The rest of this paper is organized as follows. Section \ref{Related-Work} reviews related work on small target motion detection. In Section \ref{Formulation-of-the-System}, we introduce our proposed visual system model. Section \ref{Results-and-Discussions} provides extensive performance evaluation as well as comparisons against the existing models. Finally, we conclude this paper in Section \ref{Conclusion}.

\section{Related Work}
\label{Related-Work}

Small target motion detection aims to detect objects of interest which move against cluttered natural environments and appear as small dim speckles\footnote{The sizes of small dim speckles vary from $1$ pixel to $10 \times 10$ pixels, whereas other physical characteristics, such as color, shape and texture, are difficult to recognize and cannot be used for motion detection.} in images. Inspired by the insect's motion-sensitive neurons, several models have been developed to detect small target motion. In this section, we firstly review motion-sensitive neural models, then briefly discuss traditional motion detection and small target detection approaches.

\subsection{Motion-sensitive Neural Models}
Small target motion detectors (STMDs) \cite{kelecs2017object,nordstrom2006insect,nordstrom2012neural} and lobula plate tangential cells (LPTCs) \cite{lee2015spatio,li2017local} are widely investigated motion-sensitive neurons, where the former shows exquisite sensitivity to small target motion while the latter responds strongly to wide-field motion. 

Wiederman \emph{et al.} \cite{wiederman2008model} presented a mathematical model called ESTMD to simulate STMD neurons. It can detect the presence of small moving targets, but is unable to estimate motion direction. To address this issue, directional selectivity has been introduced into the ESTMD \cite{wiederman2013biologically,bagheri2017performance,wang2018directionally}. However, these models cannot discriminate small targets from fake features, as they only make use of motion information.

The first LPTC model called elementary motion detector (EMD) \cite{hassenstein1956systemtheoretische}, is originally inferred from the insects' behavior. Following that, several studies have been done to further improve the EMD, such as \cite{eichner2011internal,wang2018improved,clark2011defining}. These models can detect all objects' motion, nevertheless they are unable to distinguish small moving objects from large ones.

\subsection{Traditional Motion Detection Methods}
Traditional motion detection methods such as optical flow \cite{fortun2015optical}, background subtraction \cite{yong2017robust} and temporal differencing \cite{li2016rotation}, have been developed to detect normal-sized objects like pedestrians and vehicles. They utilize physical characteristics including shape, color and texture, to segment regions corresponding
to moving objects from the background. Nonetheless, these methods would be  powerless for objects that are as small as one pixel or a few pixels, because it is difficult to identify objects' physical characteristics in such small sizes. Additionally, the above-mentioned methods may not work for cluttered moving backgrounds, as small moving objects could be submerged among the pixel error when applying background motion compensation \cite{ren2003motion}.

\subsection{Infrared Small Target Detection}
Previous research and application of small target detection has mainly focused on infrared images \cite{gao2013infrared,wei2016multiscale,bai2018derivative}. These infrared-based methods strongly rely on significant temperature differences between the background and objects of interest, such as rockets, jets and missiles. However, such significant temperature difference is rare in natural world. Moreover, the detection environment of these methods were mainly sky and/or ocean, which are much more clear and homogeneous than the cluttered natural environments. These infrared-based methods may not work in a natural environment with lots of bushes, trees, sunlight and shadows, let alone to meet the needs of compact in size and low energy consumption in real applications \cite{hu2016bio,indiveri2000neuromorphic,rind2003locust,he2018adaptive,polap2018multi}.

\section{Formulation of the System}
\label{Formulation-of-the-System}

\begin{figure*}[!t]
	\centering
	\includegraphics[width=1\textwidth]{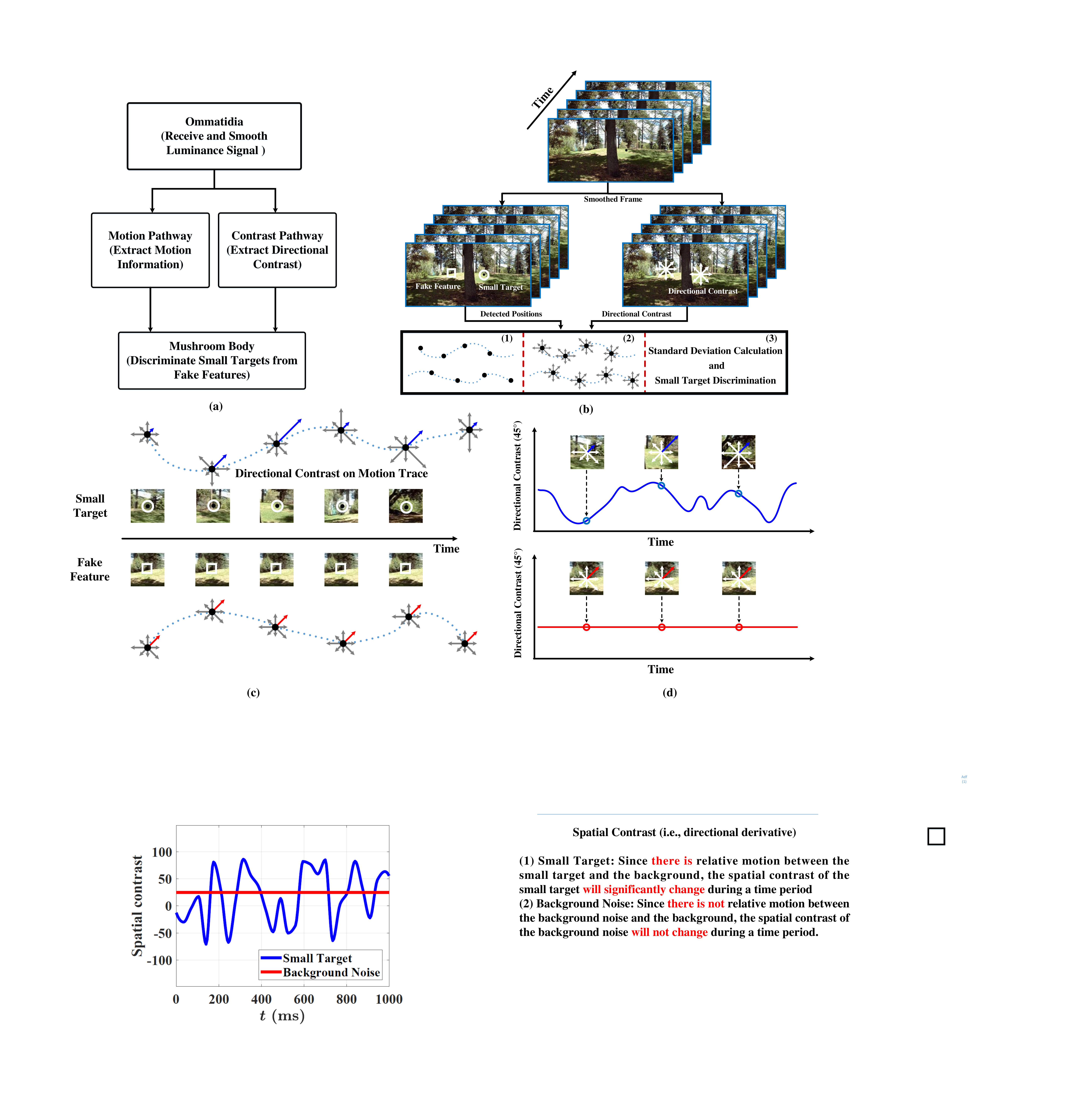}
	\caption{(a) Schematic illustration of the proposed visual system model (STMD+). (b) Image processing of the proposed visual system model. (c) Directional contrast on two motion traces which are caused by the small target and fake feature, respectively. Directional contrast is denoted by arrows along different directions where the arrow's length represents the strength of the directional contrast. For the small target (top), its directional contrast varies significantly with time. However, for the fake feature (bottom), its directional contrast shows little change over time. (d) Directional contrast along $45^{\circ}$ direction of the small target (top) and fake feature (bottom) with respect to time.}
	\label{Schematic-of-Information-Process}
\end{figure*}

In this section, we first illustrate the proposed visual system model schematically, then elaborate on its components in following subsections. 

The proposed visual system model is composed of four subsystems, including ommatidia, motion pathway, contrast pathway and mushroom body \cite{ardin2016using,webb2016neural}, as illustrated in Fig. \ref{Schematic-of-Information-Process}(a). The luminance signals are received and smoothed by the ommatidia, then applied to the motion and contrast pathways. These two pathways separately extract motion information and directional contrast which are finally integrated in the mushroom body to discriminate small targets from fake features.

Fig. \ref{Schematic-of-Information-Process}(b) shows the image processing of the proposed visual system model, where the input image sequence is processed frame by frame. In each frame, both small targets and fake features are located by computing luminance changes of each pixel over time, while directional contrast is obtained by calculating luminance changes of each pixel along different directions. The detected positions and directional contrast are further processed as follows. 
\begin{enumerate}
	\item Successively record the detected positions to infer motion traces.
	\item Extract the directional contrast on each motion trace. 
	\item Compute the standard deviation of directional contrast on each motion trace and compare it with a threshold for distinguishing small targets from fake features.
\end{enumerate}

\begin{figure*}[!t]
	\centering
	\subfloat[]{\includegraphics[width=0.48\textwidth]{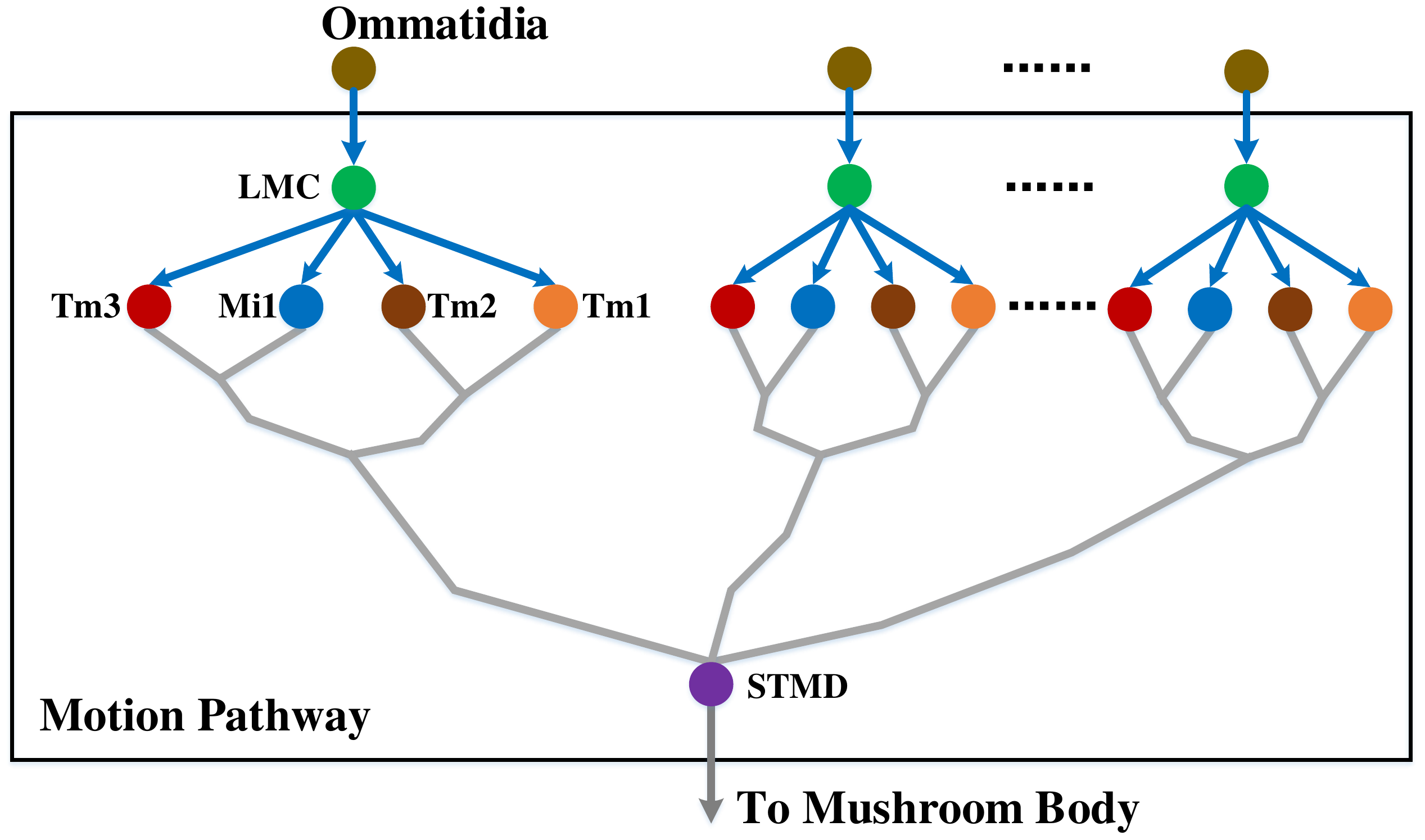}
		\label{Schematic-Motion-Pathway}}
	\hfil
	\subfloat[]{\includegraphics[width=0.48\textwidth]{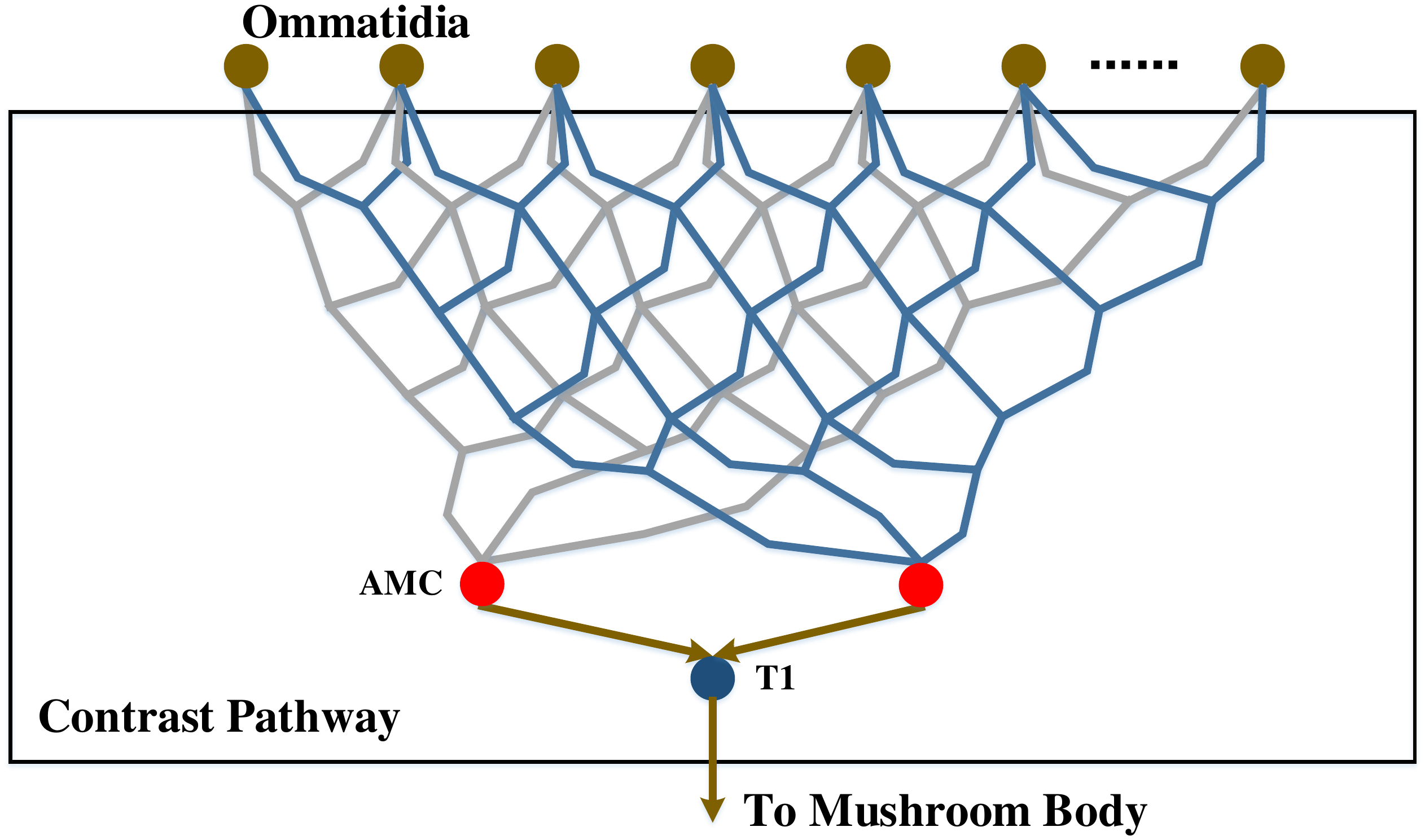}
		\label{Schematic-Contrast-Pathway}}
	\caption{Wiring sketches of motion and contrast pathways. In subplots, each colored node denotes a neuron. For clear illustration, only one STMD and T1 neurons are presented here. (a) Motion pathway. (b) Contrast pathway. Note that each AMC collects signals from multiple ommatidia while each LMC receives signals from a single ommatidium.}
	\label{Schematic-Motion-Contrast-Pathway}
\end{figure*}

\begin{figure*}[!t]
	\centering
	\includegraphics[width=1\textwidth]{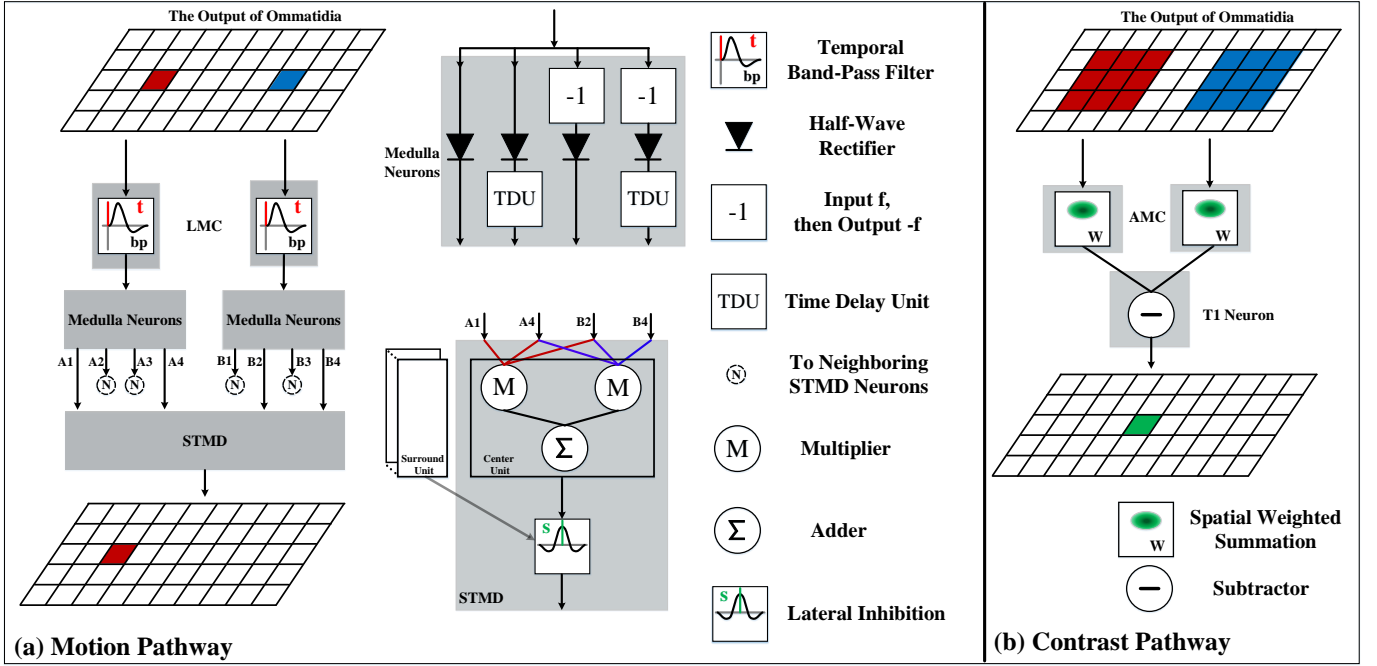}
	\caption{Schematic illustration of models of motion and contrast pathways. For clear illustration, only one STMD and T1 neurons are presented here. However, these types of neurons are all arranged in matrix form in the proposed visual system.}
	\label{Schematic-of-Signal-Process}
\end{figure*}

Our motivation is mainly based on the following observations: the directional contrast of small targets varies significantly with time, since they have relative movement to the background; on the contrary, the directional contrast of fake features shows little change over time, as they are static relative to the background. The variation amount in the  directional contrast with time is represented by the standard deviation, which is taken as the criterion for small target discrimination. Fig. \ref{Schematic-of-Information-Process}(c) visually displays the directional contrast on two typical motion traces that are separately caused by the small target and fake feature. As an example, Fig. \ref{Schematic-of-Information-Process}(d) presents the  directional contrast along $45^{\circ}$ direction, which is used to calculate the standard deviation for this direction.

\subsection{Ommatidia}
Ommatidia act as luminance receptors to perceptive visual stimuli from the natural world  \cite{warrant2016matched}. In the proposed visual system, they are arranged in a matrix and modelled as spatial Gaussian filters, each of which captures and smooths the luminance of each pixel in the input image. Formally, let $I(x,y,t) \in \mathbb{R}$ denote the input image sequence, where $x,y$ and $t$ are spatial and temporal field positions. The output of an ommatidium $P(x,y,t)$ is given by,
\begin{equation}
P(x,y,t) =  \iint I(u,v,t)G_{\sigma_1}(x-u,y-v)dudv
\label{Photoreceptors-Gaussian-Blur}
\end{equation}
where $G_{\sigma_1}(x,y)$ is a Gaussian function, defined as
\begin{equation}
G_{\sigma_1}(x,y)= \frac{1}{2\pi\sigma_1^2}\exp(-\frac{x^2+y^2}{2\sigma_1^2}).
\label{Photoreceptors-Gauss-blur-Kernel}
\end{equation}

\subsection{Motion Pathway}
As shown in Fig. \ref{Schematic-Motion-Contrast-Pathway}(a), the motion pathway consists of large monopolar cells (LMCs) \cite{freifeld2013gabaergic,behnia2014processing}, medulla neurons (i.e., Mi1, Tm1, Tm2 and Tm3)\cite{takemura2013visual,behnia2015visual}, small target motion detectors (STMDs) \cite{kelecs2017object,nordstrom2006insect,nordstrom2012neural}. The output of ommatidia is first fed into LMCs, then processed by medulla neurons and finally integrated by STMDs. Fig. \ref{Schematic-of-Signal-Process}(a) displays the model of the motion pathway, which is elaborated as follows.

\textbf{\textit{1) Large Monopolar Cells (LMCs)}:} Objects' motion can induce luminance changes of pixels with time. 
These luminance changes are extracted by the LMCs, each of  which is modelled by a temporal band-pass filter that is defined as the difference of two Gamma kernels (see Fig. \ref{Schematic-of-Signal-Process}(a)). That is,
\begin{align}
H(t) &= \Gamma_{n_1,\tau_1}(t) - \Gamma_{n_2,\tau_2}(t) \label{BPF-Para}\\
\Gamma_{n,\tau}(t) &= (nt)^n \frac{\exp(-nt/\tau)}{(n-1)!\cdot \tau^{n+1}}
\end{align}
where $H(t)$ denotes the impulse response of the band-pass filter, $\Gamma_{n,\tau}(t)$ stands for the Gamma kernel \cite{de1991theory}, $n$ and $\tau$  refers to the order and time constant of the Gamma kernel $\Gamma_{n,\tau}(t)$. Then the output of each LMC can be calculated by convolving $H(t)$ with the output of ommatidia $P(x,y,t)$,
\begin{equation}
L(x,y,t) = \int P(x,y,s)H(t-s) ds.
\label{LMCs-HPF}
\end{equation}
The  $L(x,y,t)$  reflects luminance changes of pixel $(x,y)$ over time $t$, where a positive $L(x,y,t)$ means luminance increase while a negative $L(x,y,t)$ suggests luminance decrease.

\textbf{\textit{2) Medulla Neurons}:} Medulla neurons including Tm1, Tm2, Tm3 and Mi1, constitute four parallel channels to process the output of LMCs $L(x,y,t)$. The Tm3 and Tm2 are modelled as half-wave rectifiers to separate $L(x,y,t)$ into luminance increase and decrease components. Let $S^{\text{Tm3}}(x,y,t)$ and $S^{\text{Tm2}}(x,y,t)$ denote the output of the Tm3 and Tm2, respectively, then they are given by
\begin{align}
S^{\text{Tm3}}(x,y,t) &= [L(x,y,t)]^{+}  \label{Tm3-Output} \\
S^{\text{Tm2}}(x,y,t) &= [-L(x,y,t)]^{+} \label{Tm2-Output}
\end{align}
where $[x]^+$ denotes $\max (x,0)$. The Mi1 and Tm1 further temporally delay $S^{\text{Tm3}}(x,y,t)$ and $S^{\text{Tm2}}(x,y,t)$ by convolving them with a Gamma kernel. That is, 
\begin{align}
S_{{(n,\tau)}}^{\text{Mi1}}(x,y,t) &= \int [L(x,y,s)]^{+} \cdot \Gamma_{n,\tau}(t-s) ds \label{Mi1-Output}\\
S_{{(n,\tau)}}^{\text{Tm1}}(x,y,t) &= \int [-L(x,y,s)]^{+} \cdot  \Gamma_{n,\tau}(t-s) ds \label{Tm1-Output}
\end{align}
where $S_{{(n,\tau)}}^{\text{Mi1}}(x,y,t)$ and $S_{{(n,\tau)}}^{\text{Tm1}}(x,y,t)$ represent the outputs of the Mi1 and Tm1, respectively; $n$ and $\tau$ are the order and time constant of the Gamma kernel, which separately determine the order and time-delay length of the time delay unit (TDU) (see Fig. \ref{Schematic-of-Signal-Process}(a)).

\begin{figure}[t!]
	\centering
	\subfloat[]{\includegraphics[width=0.1\textwidth]{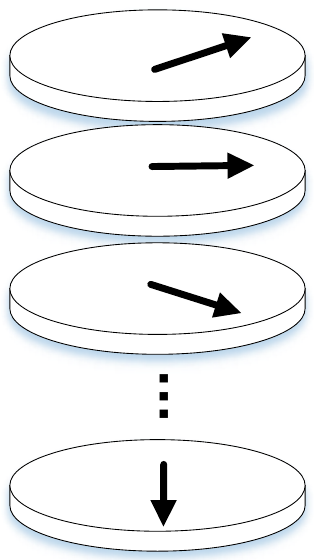}
		\label{Schematic-Illustration-Neurons-Tuned-to-Different-Directions-at-One-Position}}
	\hfil
	\subfloat[]{\includegraphics[width=0.125\textwidth]{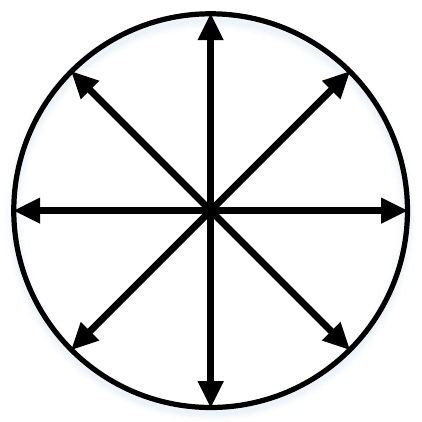}
		\label{Schematic-Illustration-Neurons-Tuned-to-Different-Directions-at-One-Position-2}}
	\caption{(a) Illustration of neurons which are located at the same position, but have different preferred directions. The black arrows denote preferred directions. (b) Illustration of different preferred directions in the x-y plane.}
	\label{Neurons-with-different-preferred-directions}
\end{figure}

\textbf{\textit{3) Small Target Motion Detectors (STMDs)}:} As can be seen from Fig. \ref{Schematic-of-Signal-Process}(a), each STMD collects the outputs of medulla neurons located at two pixels, i.e., $(x,y)$ and $(x'(\theta),y'(\theta))$ which are defined as 
\begin{equation}
\begin{split}
x'(\theta) &= x+\alpha_1\cos\theta \\ 
y'(\theta) &= y + \alpha_1\sin\theta
\end{split}
\label{DSTMD-Signal-Correlation-Distance}
\end{equation}
where $\alpha_1$ is a constant, $\theta$ denotes the preferred direction of the STMD. When a dim object successively moves over pixels $(x,y)$ and $(x'(\theta),y'(\theta))$, a luminance decrease followed by a luminance increase will appear at each of these two pixels. These luminance increase and decrease signals are first aligned in time domain and then multiplied together so as to produce a large response \cite{wang2018directionally}. That is,
\begin{equation}
\begin{split}
D(x, & y,t,\theta) = S^{\text{Tm3}}(x,y,t) \cdot \Big \{S^{\text{Tm1}}_{{(n_{_4},\tau_{_4})}}(x,y,t)\\  &+S^{\text{Mi1}}_{{(n_{_3},\tau_{_3})}}(x'(\theta),y'(\theta),t)\Big\} \cdot S^{\text{Tm1}}_{{(n_{_5},\tau_{_5})}}(x'(\theta),y'(\theta),t)
\label{DSTMD-Signal-Correlation}
\end{split}
\end{equation}
where $D(x,y,t,\theta)$ denotes the output of the STMD neuron with a preferred direction $\theta$. Here, $\theta$ belongs to $\{0, \frac{\pi}{4}, \frac{\pi}{2}, \frac{3\pi}{4}, \pi, \frac{5\pi}{4}, \frac{3\pi}{2}, \frac{7\pi}{4}\}$, corresponding to eight preferred directions of STMD neurons (see Fig. \ref{Neurons-with-different-preferred-directions}). It is worthy to note that $\tau_3$, $\tau_4$ and $\tau_5$ are determined by the different delays among the luminance changes, while $n_3$, $n_4$ and $n_5$ are accordingly tuned to guarantee appropriate Gamma kernel shapes \cite{wang2018directionally}.

So far, the obtained $D(x,y,t,\theta)$ can detect both small and large moving objects in the forms of producing a large response. In order to suppress the responses to large moving objects, the $D(x,y,t,\theta)$ is further laterally inhibited by convolving with an inhibition kernel $W_s(x,y)$. That is,
\begin{equation}
E(x,y,t,\theta) = \iint D(u,v,t,\theta) W_s(x-u,y-v) du dv
\label{DS-STMD-Lateral-Inhibition}
\end{equation}
where $E(x,y,t,\theta)$ represents the inhibited signal; the inhibition kernel $W_s(x,y)$ is defined as
\begin{align}
W_s(x,y) &= A \cdot [g(x,y)]^{+} + B \cdot [g(x,y)]^{-}  \label{Inhibition-Kernel-W2-1}\\
g(x,y)  &= G_{\sigma_2}(x,y) - e \cdot G_{\sigma_3}(x,y) - \rho
\label{DSTMD-Lateral-Inhibition-Kernel-W2-2}
\end{align}
where $[x]^+$ and $[x]^-$ respectively denote $\max (x,0)$ and $\min (x,0)$; $A$, $B$, $e$ and $\rho$ are constant. 

By comparing the $E(x,y,t,\theta)$ with a detection threshold $\beta$, we can find the positions of small moving objects. Specially, if $E(x,y,t,\theta) > \beta$, then we believe that a small object moving along direction $\theta$ is located at pixel $(x,y)$ and time $t$.
However, it cannot distinguish small targets and fake features that can be both recognized as small moving objects. To address this issue, we construct a contrast pathway accounting for directional contrast calculation.   

\subsection{Contrast Pathway}

\begin{figure*}[!t]
	\vspace{-10pt}
	{\centering
	\subfloat[]{\includegraphics[width=0.23\textwidth]{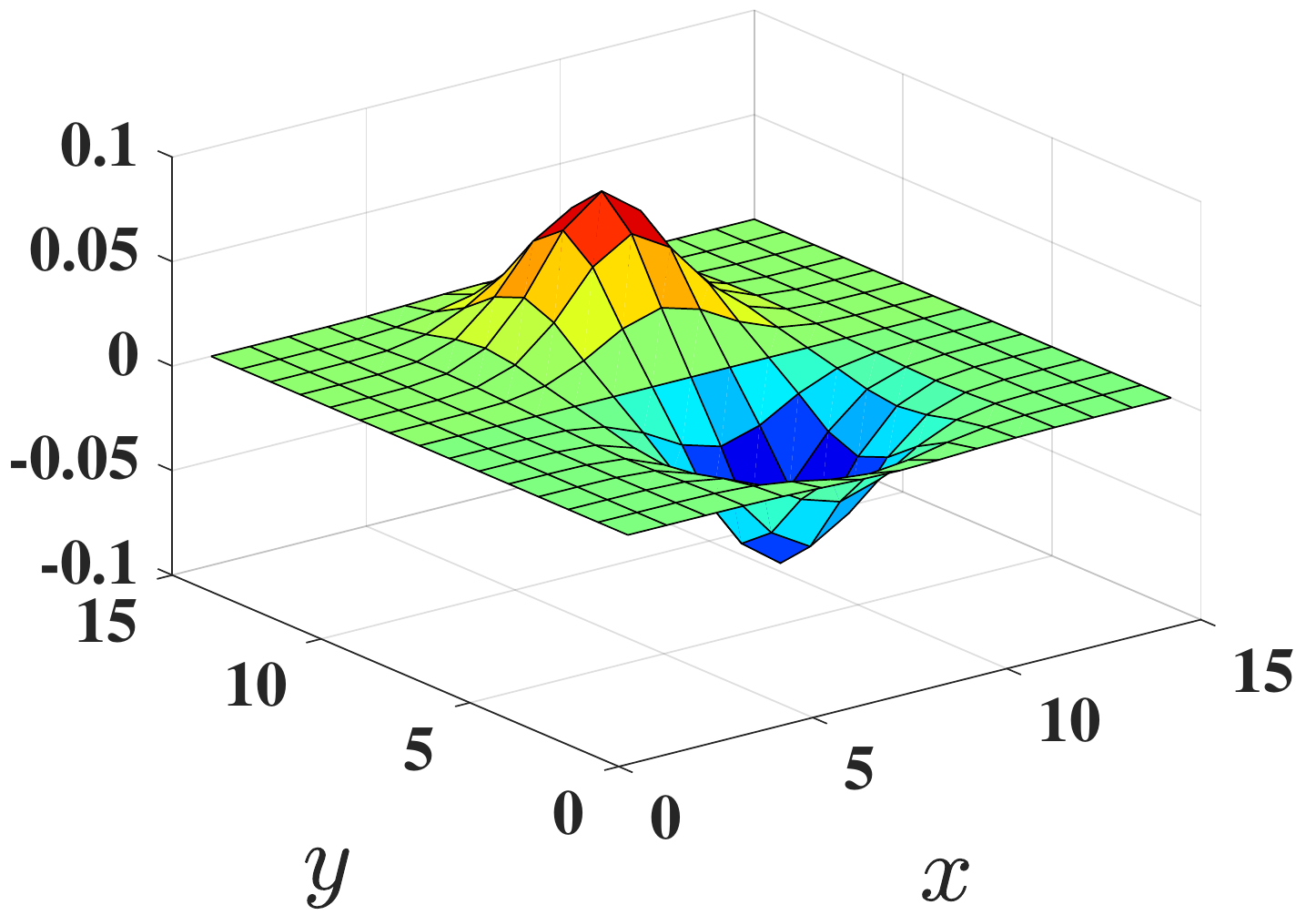}
		\label{Contrast-Pathway-T1-Kernel-1}}
	\hfill
	\subfloat[]{\includegraphics[width=0.223\textwidth]{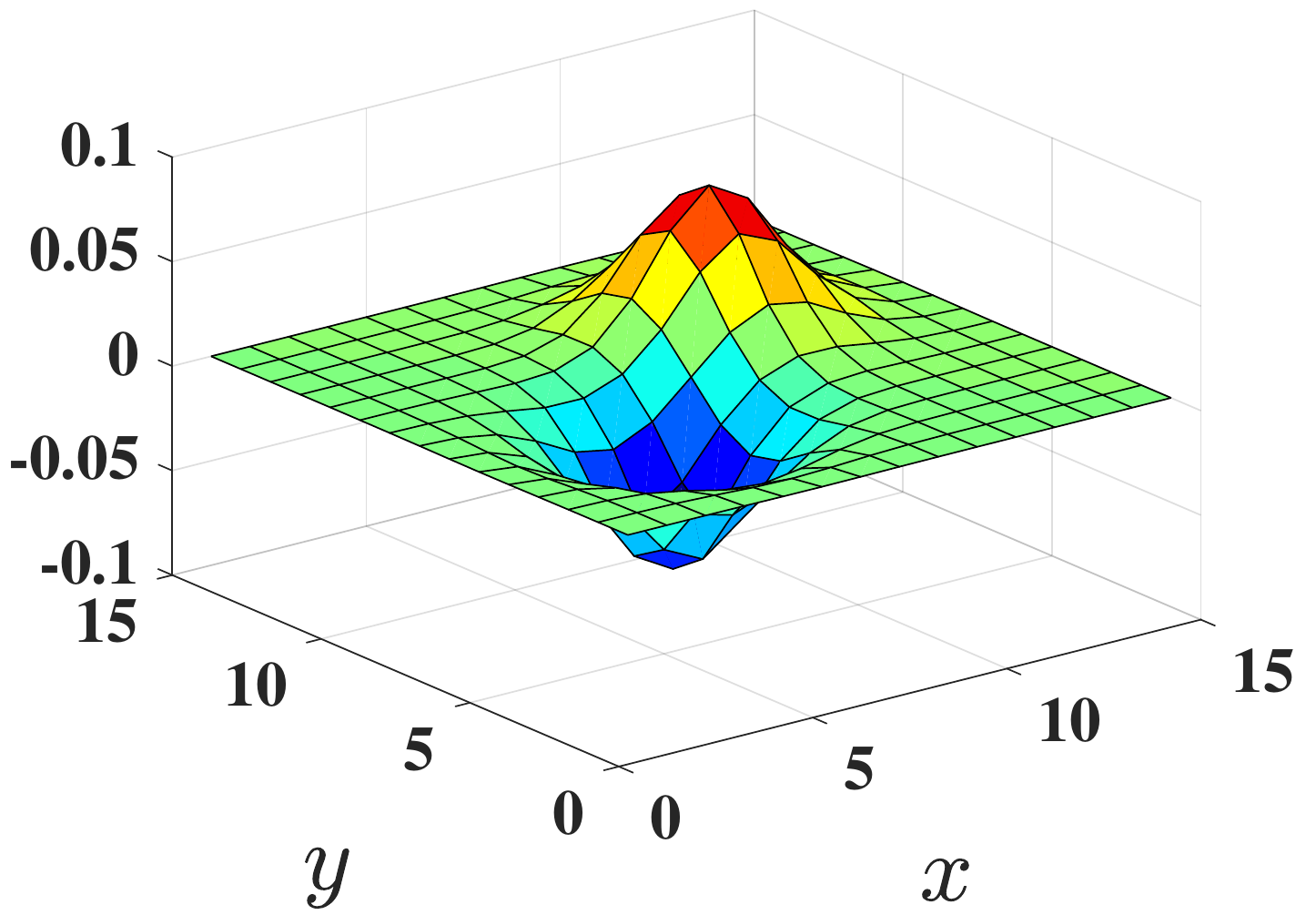}
		\label{Contrast-Pathway-T1-Kernel-2}}
	\hfill
	\subfloat[]{\includegraphics[width=0.23\textwidth]{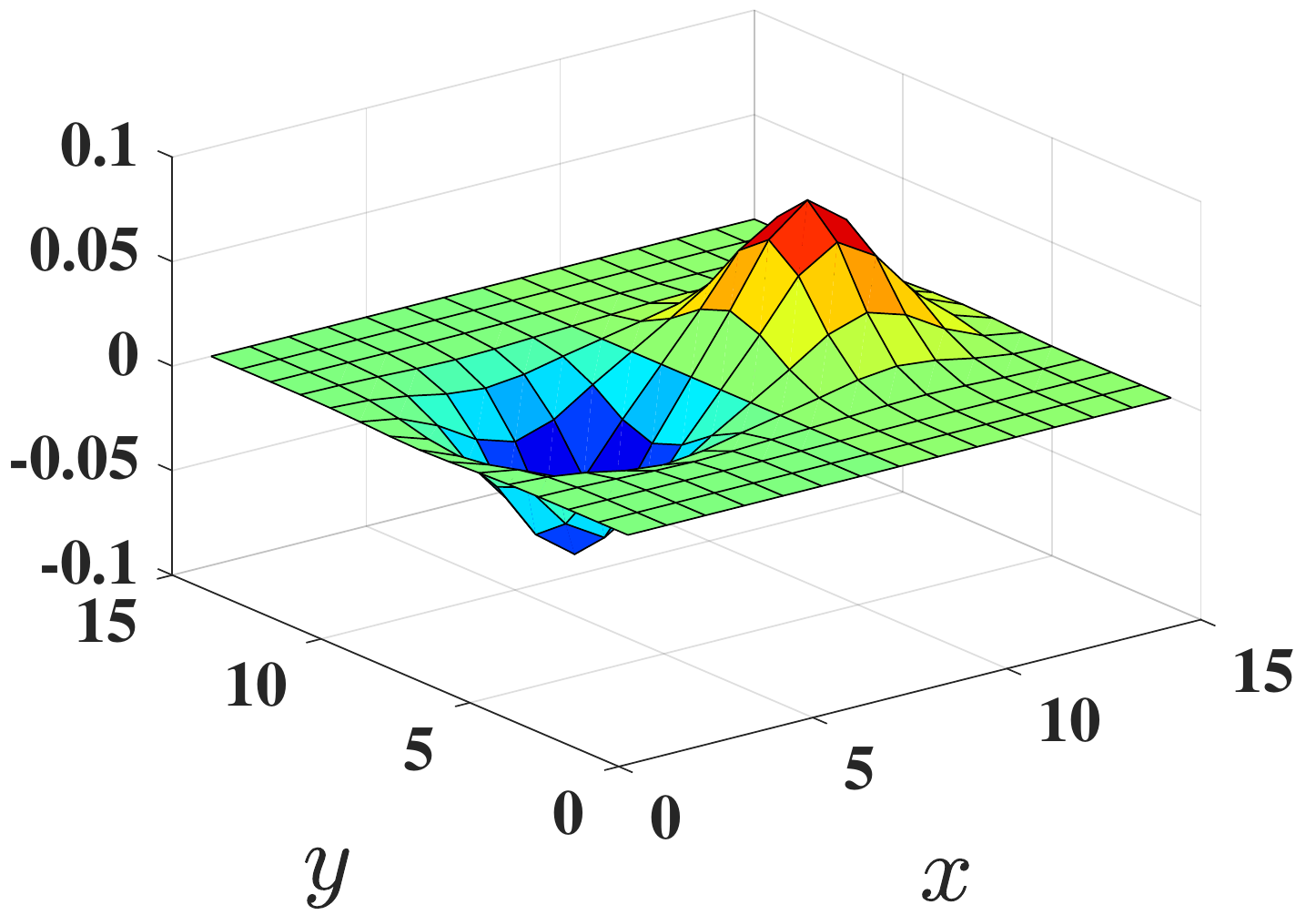}
		\label{Contrast-Pathway-T1-Kernel-3}}
	\hfill
	\subfloat[]{\includegraphics[width=0.23\textwidth]{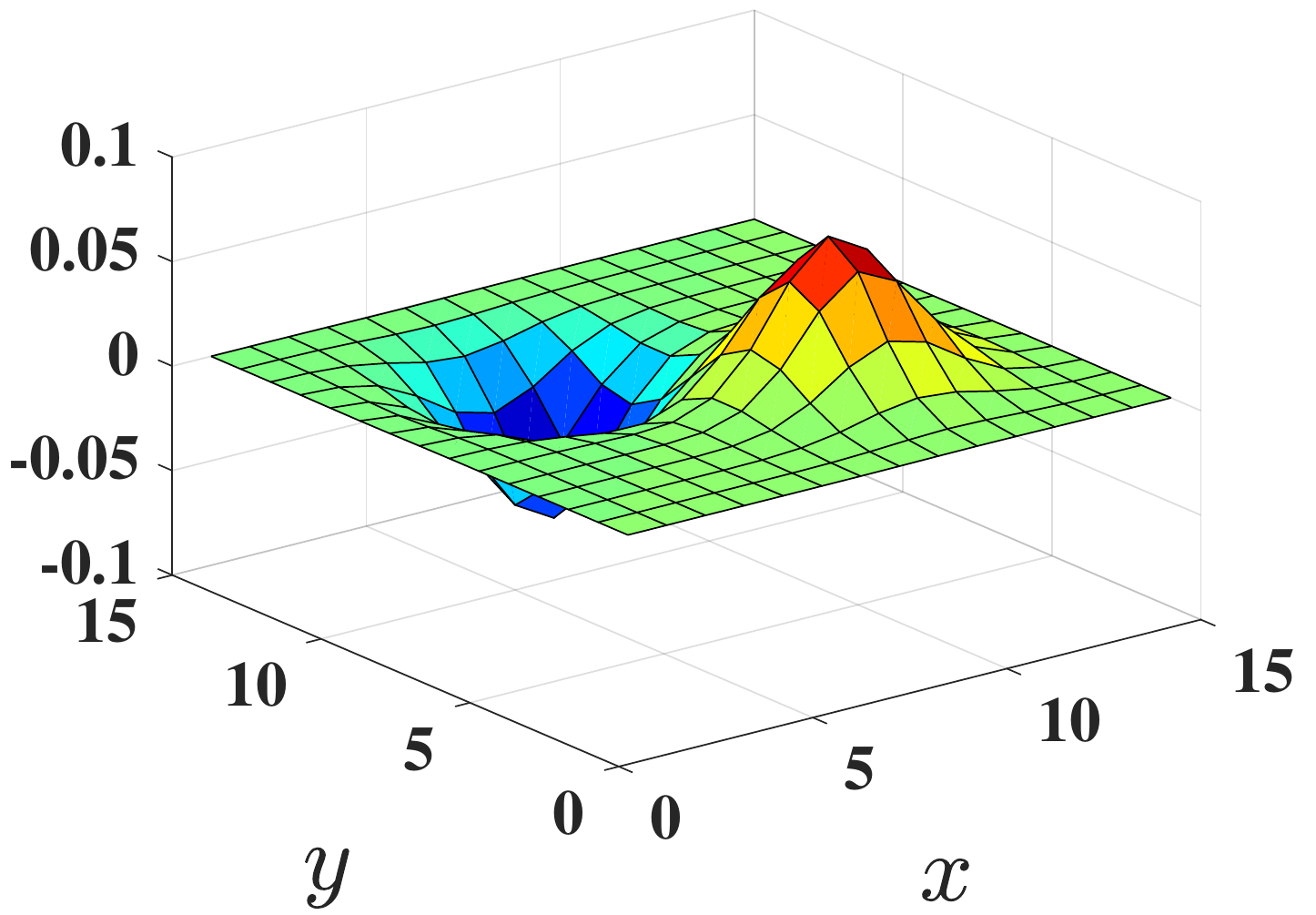}
		\label{Contrast-Pathway-T1-Kernel-4}}
	}
	\caption{Illustration of the convolution kernel $W_{\scriptscriptstyle T}(x,y,\phi)$. (a) $\phi = 0$. (b) $\phi = \frac{\pi}{4}$. (c) $\phi = \frac{\pi}{2}$. (d) $\phi = \frac{3\pi}{4}$.}
	\label{Contrast-Pathway-T1-Kernel}
\end{figure*}

As shown in Fig. \ref{Schematic-Motion-Contrast-Pathway}(b), contrast pathway is composed of amacrine cells (AMCs)\cite{St2016Adaptations,Riveraalvidrez2011A,lessios2018multiple} and T1 neurons \cite{yamaguchi2011photoreceptors,rogers2015differential}. The output of ommatidia is firstly fed into AMCs, then processed by T1 neurons.  Fig. \ref{Schematic-of-Signal-Process}(b) displays the model of the contrast pathway, which is elaborated as follows.

\textbf{\textit{1) Amacrine Cells (AMCs)}:}  Each AMC receives the output of multiple ommatidia located in a small region and serves as a weighted summation unit, as presented in Fig. \ref{Schematic-of-Signal-Process}(b). Here, we define the weight function as 
\begin{equation}
W_{\scriptscriptstyle A}(x,y) = \frac{1}{2\pi\eta^2}\exp(-\frac{x^2+y^2}{2\eta^2})
\label{AMC-1-Kernel}
\end{equation}
where $\eta$ is constant. Then the output of each AMC $A(x,y,t)$ can be given by
\begin{equation}
A(x,y,t) =  \iint P(u,v,t)W_{\scriptscriptstyle A}(x-u,y-v)dudv 
\label{AMC-Output}
\end{equation}
where $P(x,y,t)$ is the output of ommatidia defined in (\ref{Photoreceptors-Gaussian-Blur}).

\textbf{\textit{2) T1 Neurons}:} The T1 neuron layer is adopted to extract the directional contrast along different directions. The directional contrast at $(x,y)$ along direction $\phi$ is defined as the difference between the outputs of two AMCs that are located at $(x+\alpha_2\cos\phi, y+\alpha_2\sin\phi)$ and $(x-\alpha_2\cos\phi, y-\alpha_2\sin\phi)$. Here, $\alpha_2$ is a constant. Let $T(x,y,t,\phi)$ denote the output of a T1 neuron with  a preferred direction $\phi$, then it can be given by
\begin{equation}
\begin{split}
T(x,y,t,\phi) = & A(x+\alpha_2\cos\phi, y+\alpha_2\sin\phi,t) \\
                  & - A(x-\alpha_2\cos\phi, y-\alpha_2\sin\phi,t).
\label{T1-Output}
\end{split}
\end{equation}
Substituting (\ref{AMC-Output}) in (\ref{T1-Output}), we have
\begin{equation}
T(x,y,t,\phi) = \iint P(u,v,t)W_{\scriptscriptstyle T}(x-u,y-v,\phi)dudv
\end{equation}
where the convolution kernel $W_{\scriptscriptstyle T}(x,y,\phi)$ represents
\begin{equation}
\begin{split}
W_{\scriptscriptstyle T}(x,y,\phi) =& W_{\scriptscriptstyle A}(x+\alpha_2\cos\phi, y+\alpha_2\sin\phi) \\
                       &-W_{\scriptscriptstyle A}(x-\alpha_2\cos\phi, y-\alpha_2\sin\phi).
\end{split}
\end{equation}
Here $\phi$ belongs to $\{0, \frac{\pi}{4}, \frac{\pi}{2}, \frac{3\pi}{4}\}$, corresponding to four preferred directions of T1 neurons. It is worthy to note that the convolution kernel $W_{\scriptscriptstyle T}(x,y,\phi)$ is one of the directional derivative operators \cite{freeman1991design,zhang2015contour}, which can extract
anisotropic luminance variations (see Fig. \ref{Contrast-Pathway-T1-Kernel}).

\subsection{Mushroom Body}
In the proposed visual system, the mushroom body \cite{ardin2016using,webb2016neural} receives two types of neural outputs, including the output of STMDs $E(x,y,t,\theta)$ and the output of T1 neurons $T(x,y,t,\phi)$. These neural outputs are integrated to discriminate small targets from fake features via the following three procedures.

\begin{figure}[t]
	\centering
	\includegraphics[width=0.35\textwidth]{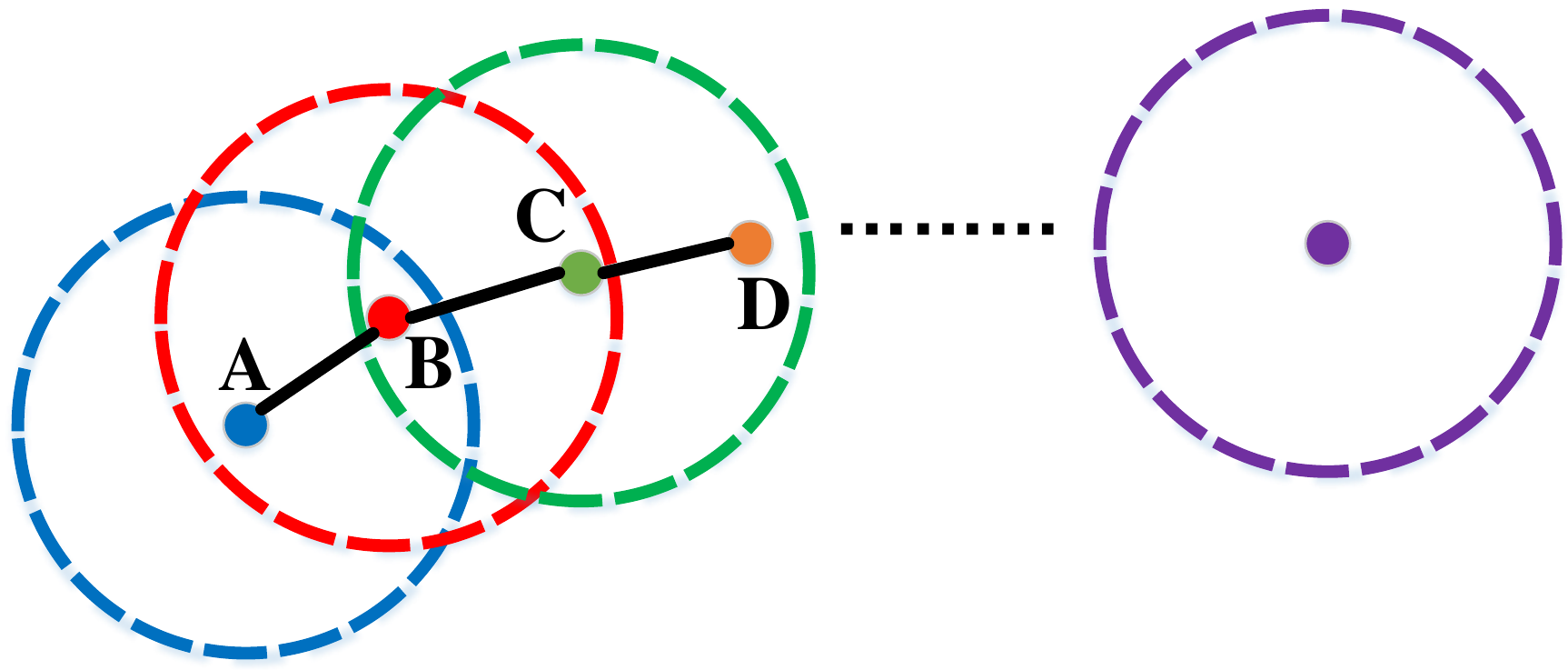}
	\caption{Motion trace recording. Each node denotes a detected pixel while each circle represents a small neighborhood. If pixel B is the nearest detected point to pixel A, and pixel B is in the neighborhood of pixel A, then we believe that pixels A and B belong to the same motion trace. Repeating this step, a motion trace could be recorded.}
	\label{Schematic-Illustration-Motion-Trail-Recording}
\end{figure}

\textbf{\textit{1) Motion Trace Recording}:} The output of STMDs $E(x,y,t,\theta)$ is employed to record motion traces of small objects. For a detection threshold $\beta$ and a starting time $t_0$, if there exists a pixel A $(x_{\scriptscriptstyle A},y_{\scriptscriptstyle A})$ and motion direction $\theta_{\scriptscriptstyle A}$ which satisfy $E(x_{\scriptscriptstyle A},y_{\scriptscriptstyle A},t_0,\theta_{\scriptscriptstyle A}) > \beta$, then we believe that a small object\footnote{The detected object could be a small target or a fake feature, which cannot be discriminated by the STMDs.} is detected at pixel $(x_{\scriptscriptstyle A},y_{\scriptscriptstyle A})$ and its motion direction is $\theta_{\scriptscriptstyle A}$. Similarly, at next time step $t_1$, another pixel B $(x_{\scriptscriptstyle B},y_{\scriptscriptstyle B})$ and motion direction $\theta_{\scriptscriptstyle B}$ can be detected. Especially, if pixel B $(x_{\scriptscriptstyle B},y_{\scriptscriptstyle B})$ is the nearest detected point to pixel A $(x_{\scriptscriptstyle A},y_{\scriptscriptstyle A})$, and pixel B is in the small neighborhood of pixel A, then we believe that pixels A and B belong to the same motion trace denoted by $TR$. Repeating the above steps, the motion trace $TR$ can be recorded during a time period, as shown in Fig. \ref{Schematic-Illustration-Motion-Trail-Recording}. The $TR$ can be described as,
\begin{equation}
	TR = {(x(t),y(t),\theta(t))}, t \in [t_0,t_n]
	\label{Motion-Trace}
\end{equation}
where $x(t)$ and $y(t)$ represent x and y coordinates at time $t$, $\theta(t)$ denotes motion direction, $t_0$ and $t_n$ are the starting time and current time.	
		
\textbf{\textit{2) Information Integration}:} Once motion traces are recorded, we can obtain their directional contrast by substituting (\ref{Motion-Trace}) into $T(x,y,t,\phi)$. That is,
\begin{equation}
Q(t,\phi) = T(x(t),y(t),t,\phi), t \in [t_0,t_n]
\label{T1-Neural-Output-Along-Target-Trace}
\end{equation}
where $Q(t,\phi)$ denotes the directional contrast along direction $\phi$ on the motion trace $TR$; $(x(t),y(t))$ stands for the point on the motion trace. To quantify the variation amount in the directional contrast, we calculate the standard deviation ($SD$) of the  $Q(t,\phi)$ during a time period $[t_{n-m},t_n]$, denoted by $SD(t_{n-m},t_n,\phi)$. Here $m$ represents the sample number for the $SD$ calculation. 

\textbf{\textit{3) Small Target Discrimination}:} We determine whether a detected object is a small target or a fake feature, using the standard deviations of the directional contrast on the object's motion trace, i.e., $SD(t_{n-m},t_n,\phi)$. If the $SD(t_{n-m},t_n,\phi)$ is smaller than a certain threshold, we believe that the detected object is a fake feature; Otherwise, it is a small target.

\subsection{Parameter Setting}
Parameters of the proposed visual system model are listed in Table \ref{Table-Parameter-FVS}, where the parameters of the motion pathway are determined by the analysis in \cite{wang2018directionally} while those of the contrast pathway are tuned empirically. These parameters are chosen to satisfy the functionality, which are mainly determined by the velocity and size ranges of the moving targets. They will not be changed in the following experiments unless stated.

The proposed visual system model is written in Matlab (The MathWorks, Inc., Natick, MA). The computer used in the experiments is a standard laptop with a $2.50$GHz Intel Core i7 CPU and $16$GB DDR3 memory. The source code can be found at https://github.com/wanghongxin/STMD-Plus.

\begin{table}[t!]
	\renewcommand{\arraystretch}{1.3}
	\caption{Parameters of the proposed visual system model.}
	\label{Table-Parameter-FVS}
	\centering
	\begin{tabular}{cc}
		\hline
		Eq. & Parameters \\	
		\hline
		(\ref{Photoreceptors-Gaussian-Blur}) & $\sigma_1 = 1$ \\

		(\ref{BPF-Para}) & $n_1 = 2, \tau_1= 3, n_2 = 6,\tau_2 = 9$\\
		
	    (\ref{DSTMD-Signal-Correlation-Distance}) & $\alpha_1 = 3$ \\

		(\ref{DSTMD-Signal-Correlation}) & $n_3 = 3, \tau_3 = 15, n_4 = 5, \tau_4 = 25, n_5 = 8, \tau_5 = 40$ \\
				
		(\ref{Inhibition-Kernel-W2-1}) & $A = 1, B = 3$ \\
		
		(\ref{DSTMD-Lateral-Inhibition-Kernel-W2-2}) & $\sigma_2 = 1.5, \sigma_3 = 3.0, e = 1, \rho = 0$ \\
		
		(\ref{AMC-1-Kernel}) & $\eta = 1.5$ \\

		(\ref{T1-Output}) & $\alpha_2 = 3$ \\
		\hline
	\end{tabular}
\end{table}	


\section{Results and Discussions}
\label{Results-and-Discussions}
The proposed visual system model is evaluated on a synthetic dataset \cite{straw2008vision} and a real dataset (STNS dataset) \cite{bagheri2017performance}. The synthetic dataset contains a number of image sequences which are synthesized by using real background images and a computer generated small target (a black block). These image sequences all display the motion of the small target against the cluttered moving backgrounds, which are different in the target sizes, target velocities, background velocities, background types and so on. The sampling frequencies of the synthetic videos are all equal to $1000$ Hz. The STNS dataset is a collection of $25$ real videos featuring various moving targets and environments. The scenarios include many kinds of challenges, such as heavy clutter, camera motion and changes in overall brightness. The STNS dataset (videos and manual ground truth annotations) is available at {https://figshare.com/articles/STNS\_Dataset/4496768}.

To quantitatively evaluate the detection performance, two metrics are defined as following \cite{gao2013infrared},
\begin{align}
D_R & = \frac{\text{number of true detections}}{\text{number of actual targets}} \\
F_A & = \frac{\text{number of false detections}}{\text{number of images}}
\end{align}
where $D_R$ and $F_A$ denote detection rate and false alarm rate, respectively. The detected result is considered correct if the pixel distance between the ground truth and the result is within a threshold ($5$ pixels).

\subsection{Signal Processing in the Motion Pathway}

\begin{figure}[!t]
	\centering
	\includegraphics[width=0.35\textwidth]{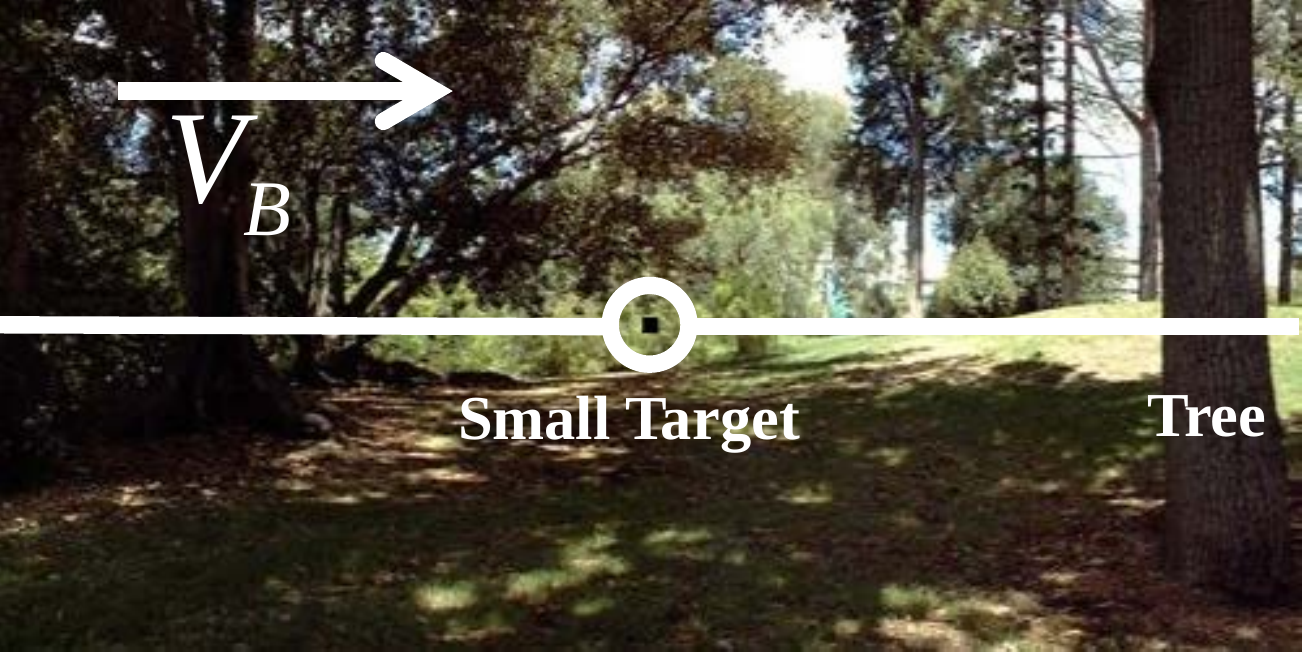}
	\caption{Input frame at time $t_0 = 1000$ ms whose resolution is $500$ pixels (in horizontal) by $250$ pixels (in vertical). The small target (the black block) and the cluttered background are moving from left to right. Their velocities are all equal to  $250$ pixel/s, where arrow $V_B$ denotes the motion direction of the background. The tree which is regarded as a large object, is also moving due to the background motion.}
	\label{Input-Image-Frame-Middle-Line-Highlighted}
\end{figure}

\begin{figure}[!t]
	\centering
	\includegraphics[width=3.5in]{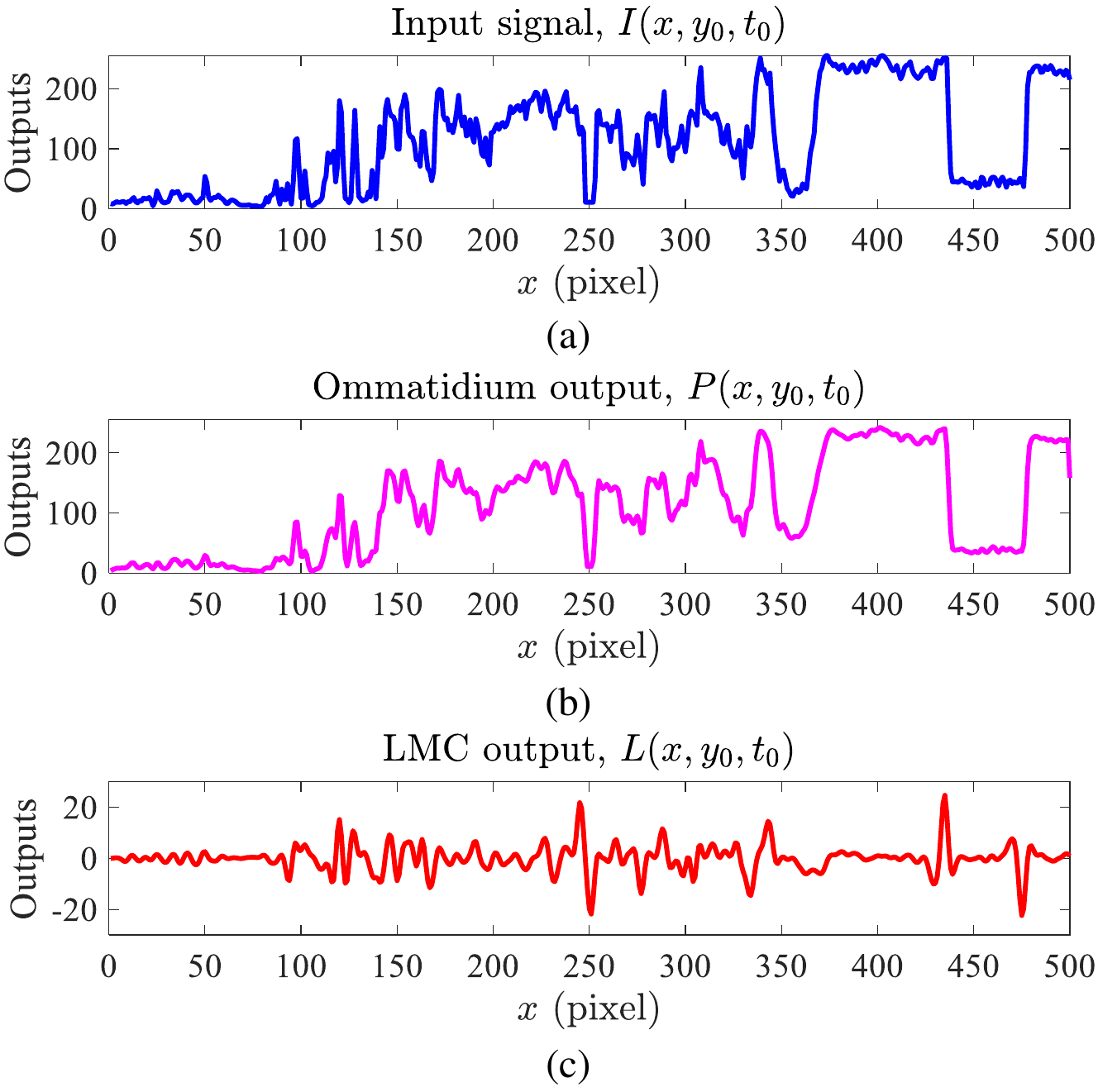}
	\caption{In each subplot, the horizontal axis denotes $x$ coordinate while the vertical axis represents neural outputs. (a) Input luminance signal $I(x,y_0,t_0)$. (b) Ommatidium output $P(x,y_0,t_0)$. (c) LMC output $L(x,y_0,t_0)$.}
	\label{Layer-Output-Input-Ommatidium-LMC}
\end{figure}

\begin{figure}[!t]
	\centering
	\includegraphics[width=3.5in]{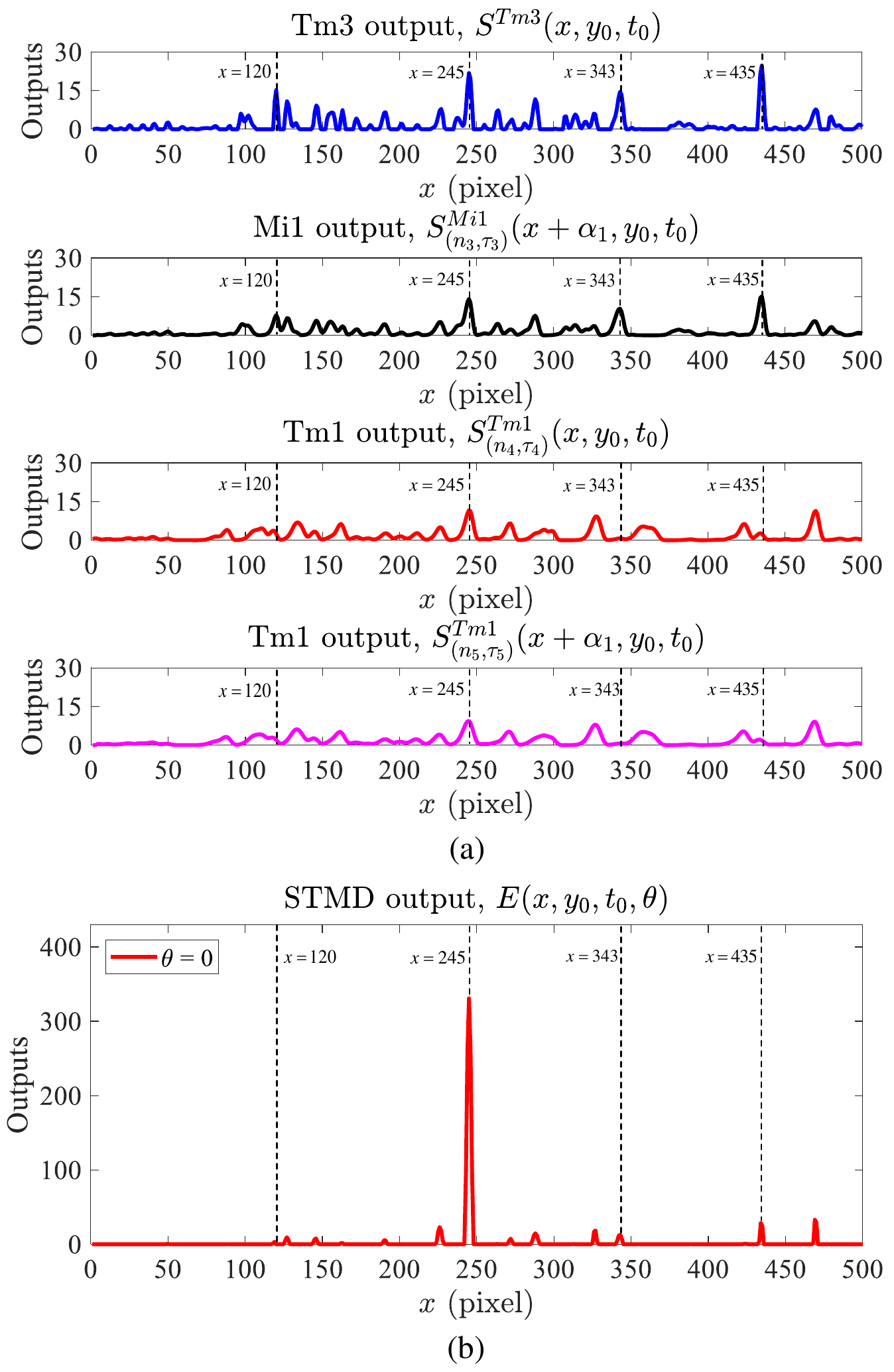}
	\caption{In each subplot, the horizontal axis denotes $x$ coordinate while the vertical axis represents neural outputs. (a) Four inputs of the STMDs when the preferred direction $\theta$ is set to $0$, i.e., $S^{\text{Tm3}}(x,y_0,t_0)$, $S^{\text{Mi1}}_{{(n_3,\tau_3)}}(x+\alpha_1,y_0,t_0)$, $S^{\text{Tm1}}_{{(n_4,\tau_4)}}(x,y_0,t_0)$  and $ S^{\text{Tm1}}_{{(n_5,\tau_5)}}(x+\alpha_1,y_0,t_0)$. (b) STMD output $E(x,y_0,t_0,\theta)$ when the preferred direction $\theta$ is equal to $0$.}
	\label{Layer-Output-Medulla-STMD}
\end{figure}

To intuitively illustrate the signal processing in the motion pathway, we observe the output of each neural layer with respect to $x$ by setting $y$ and $t$ as $y_0 = 125$ pixel and $t_0 = 1000$ ms. Fig. \ref{Input-Image-Frame-Middle-Line-Highlighted} shows the input frame at time $t_0 = 1000$ ms, where the luminance signal $I(x,y_0,t_0)$ on the middle line is presented in Fig. \ref{Layer-Output-Input-Ommatidium-LMC}(a). Its resulting  ommatidium output and LMC output are displayed in  Fig. \ref{Layer-Output-Input-Ommatidium-LMC}(b) and (c), respectively. The ommatidum output is a smoothed version of the input signal. The LMC output reveals the luminance changes of pixels, where the positive values correspond to luminance increase while the negative values suggest luminance decrease.

Fig. \ref{Layer-Output-Medulla-STMD}(a) demonstrates the four inputs of the STMDs when the preferred direction $\theta$ is set to $0$.
Specifically, the  $S^{\text{Tm3}}(x,y_0,t_0)$ is the positive part of the LMC output; the $S^{\text{Mi1}}_{{(n_3,\tau_3)}}(x+\alpha_1,y_0,t_0)$ denotes the delayed version of  the positive part of the LMC output with a shift of $\alpha_1$ pixels; the  $S^{\text{Tm1}}_{{(n_4,\tau_4)}}(x,y_0,t_0)$  stands for the delayed version of the negative part of the LMC output; and the $ S^{\text{Tm1}}_{{(n_5,\tau_5)}}(x+\alpha_1,y_0,t_0)$ represents the delayed version of negative part of the LMC output with a shift of $\alpha_1$ pixels. Fig. \ref{Layer-Output-Medulla-STMD}(b) further shows the output of STMDs, where a high response appears at the position of the small target ($x = 245$) while the responses at other positions are effectively suppressed. This is because the four peaks located at the position of the small target are aligned  (see Fig. \ref{Layer-Output-Medulla-STMD}(a)), which will produce a strong response after the multiplication, summation and lateral inhibition in the STMD (see Fig. \ref{Schematic-of-Signal-Process}). For other positions e.g., $x=120,343,435$, the peaks on the four curves exhibit a low aligning probability, hence producing a weak response. Note that the lateral inhibition is introduced to suppress the responses to large objects, such as the tree displayed in Fig. \ref{Input-Image-Frame-Middle-Line-Highlighted}.

\begin{figure}[!t]
	\centering
	\includegraphics[width=3.5in]{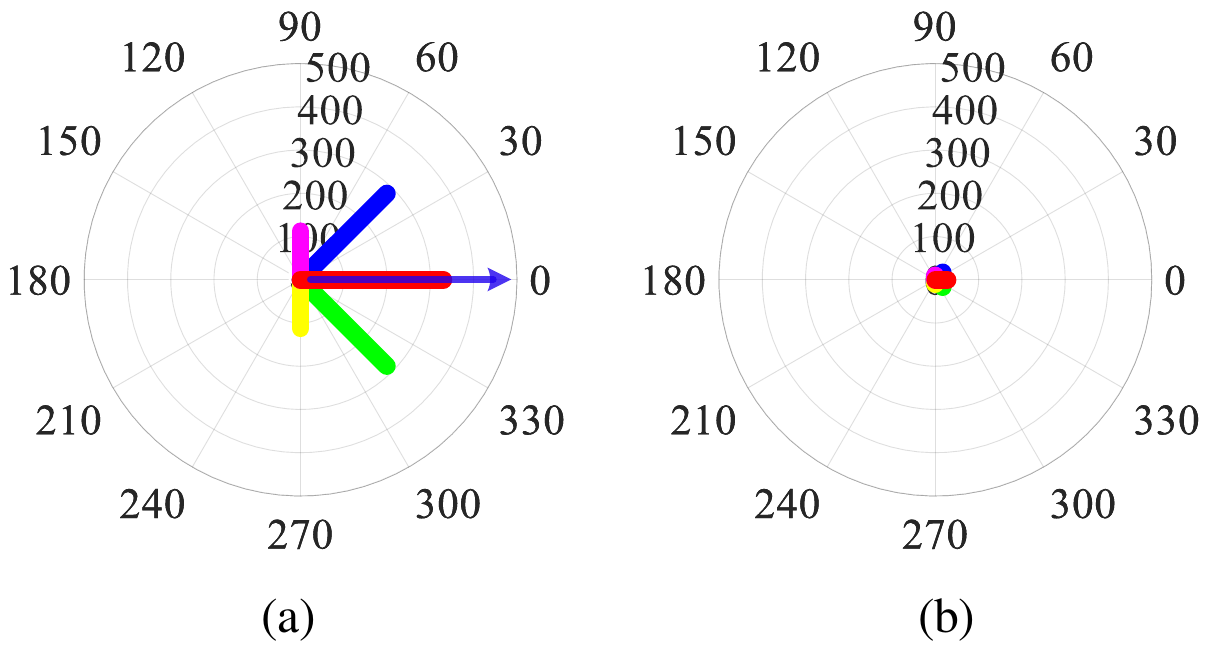}
	\caption{In the polar coordinate system, the angular coordinate represents the preferred direction $\theta$ while the radial coordinate denotes the STMD output. (a) STMD outputs at position $x=245$ along eight preferred directions $\theta \in \{0, \frac{\pi}{4}, \frac{\pi}{2}, \frac{3\pi}{4}, \pi, \frac{5\pi}{4}, \frac{3\pi}{2}, \frac{7\pi}{4}\}$. The blue arrow stands for the motion direction of the small target. (b) STMD outputs at position $x=435$ along eight preferred directions $\theta$.}
	\label{Layer-Output-STMD-Multi-Directions}
\end{figure}

It is worthy to note that the above analysis is based on the presetting of the preferred direction $\theta = 0$. When we change the preferred direction $\theta$, different STMD outputs can be calculated. Fig. \ref{Layer-Output-STMD-Multi-Directions} presents the STMD outputs at the positions $x=245$ and $x=435$ along eight preferred directions $\theta$, where $x=245$ is the position of the small target and $x=435$ corresponds to the position of the large tree.  As shown in Fig. \ref{Layer-Output-STMD-Multi-Directions}(a), for the small target, the STMD shows strong directional selectivity. As the preferred direction deviates from the motion direction of the small target, the STMD output will decrease correspondingly. On the other hand, the direction of the small target can be estimated by computing the summation of these output vectors \cite{wang2018directionally}. For the large tree (see Fig. \ref{Layer-Output-STMD-Multi-Directions}(b)), the outputs of the STMD along eight preferred directions are very low, suggesting that the STMD is not interested in large moving objects.

\subsection{Characteristics of the STMD}
\begin{figure}[t]
	\centering
	\includegraphics[width=0.18\textwidth]{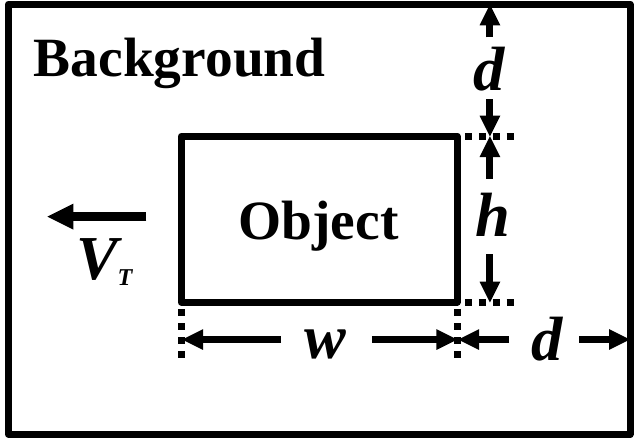}
	\caption{External rectangle and neighboring background rectangle of an object. Arrow $V_T$ denotes the motion direction of the object. $w$ represents object width while $h$ stands for object height.}
	\label{The-External-Rectangle-and-Neighboring-Background-Rectangle-of-a-Small-Target}
\end{figure}

\begin{figure}[t]
	\vspace{-10pt}
	\centering
	\subfloat[]{\includegraphics[width=0.22\textwidth]{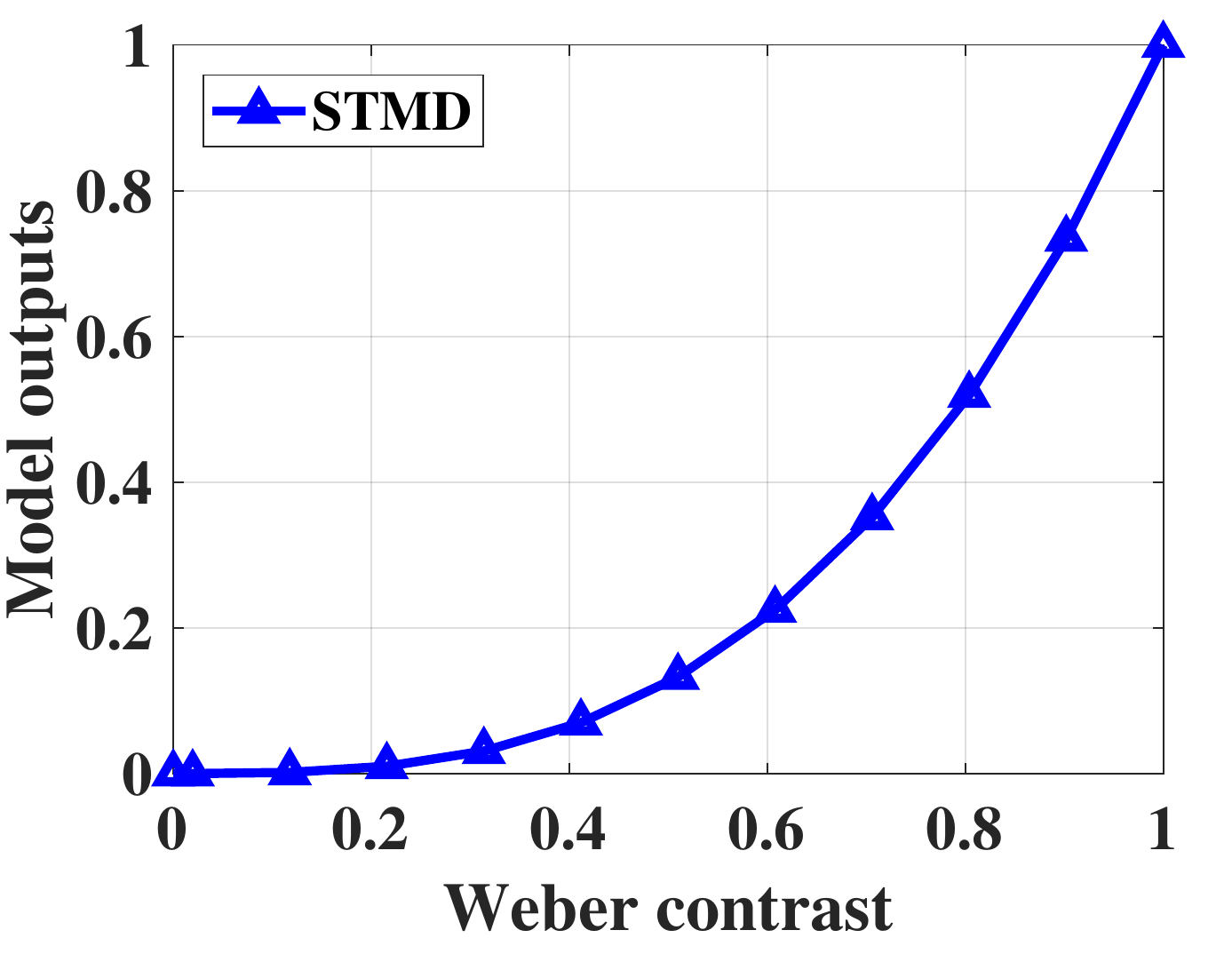}
		\label{Tuning-Properties-LDTB}}
	\hfil
	\subfloat[]{\includegraphics[width=0.22\textwidth]{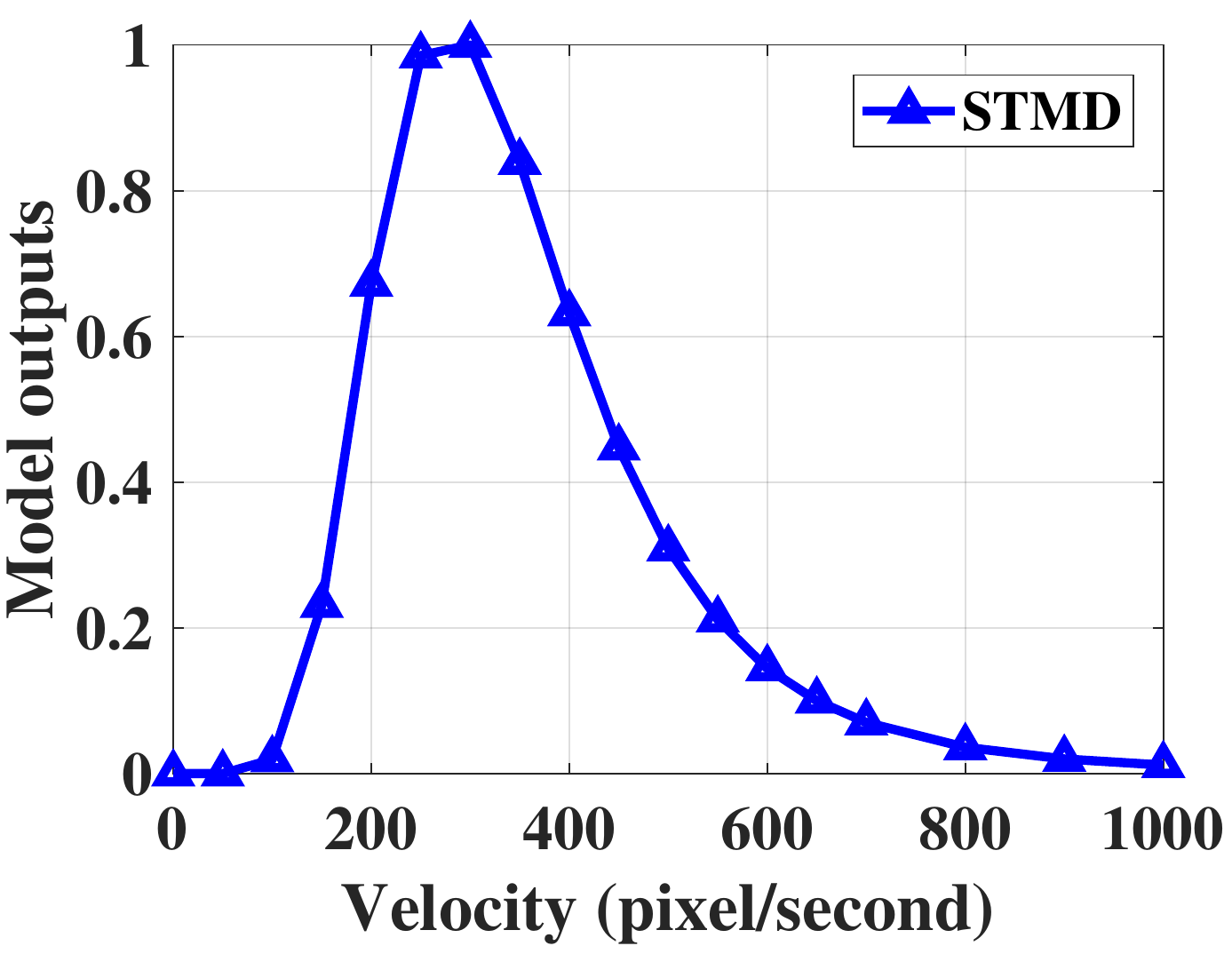}
		\label{Tuning-Properties-Velocity}}
	\hfil
	\subfloat[]{\includegraphics[width=0.22\textwidth]{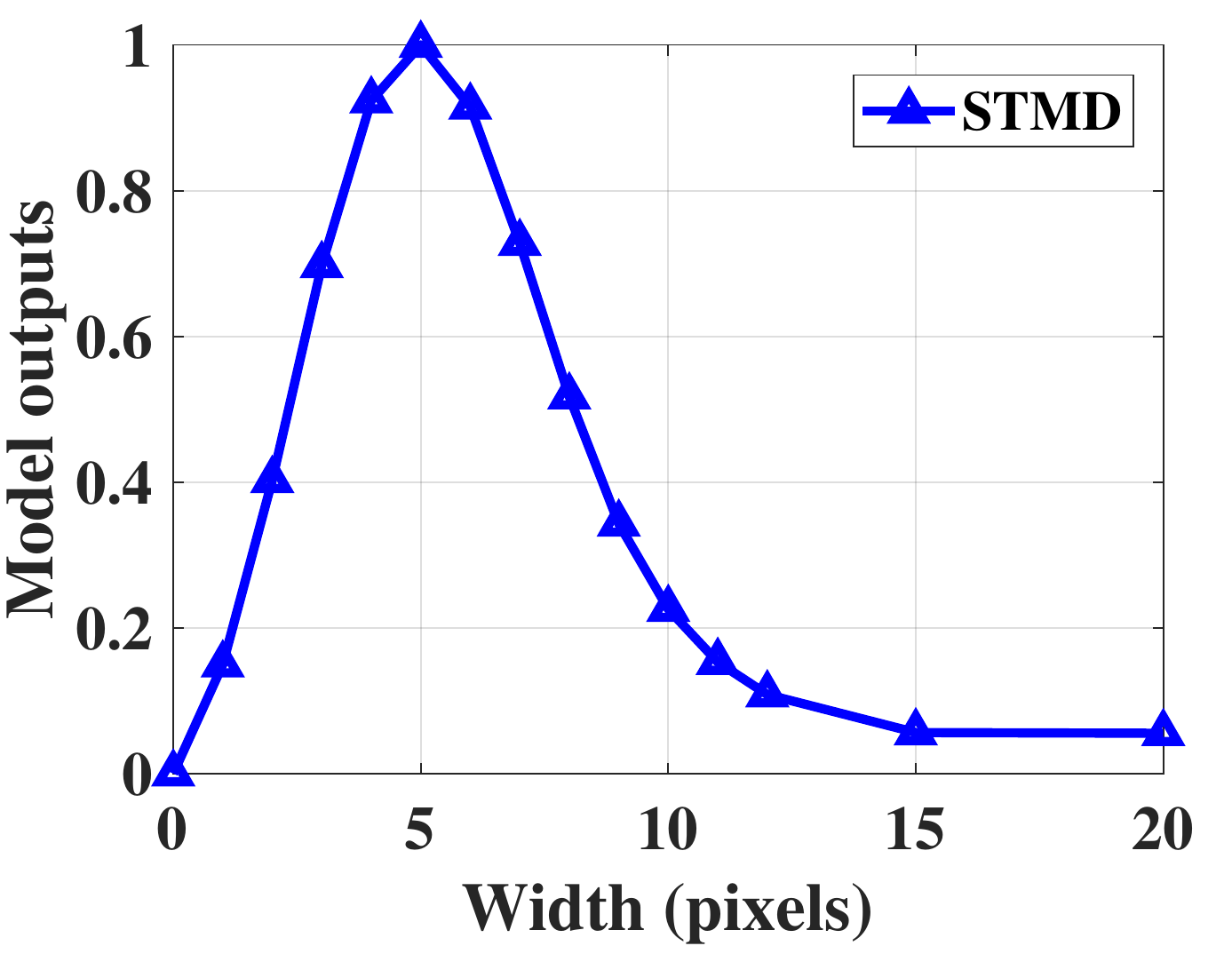}
		\label{Tuning-Properties-Width}}
	\hfil
	\subfloat[]{\includegraphics[width=0.22\textwidth]{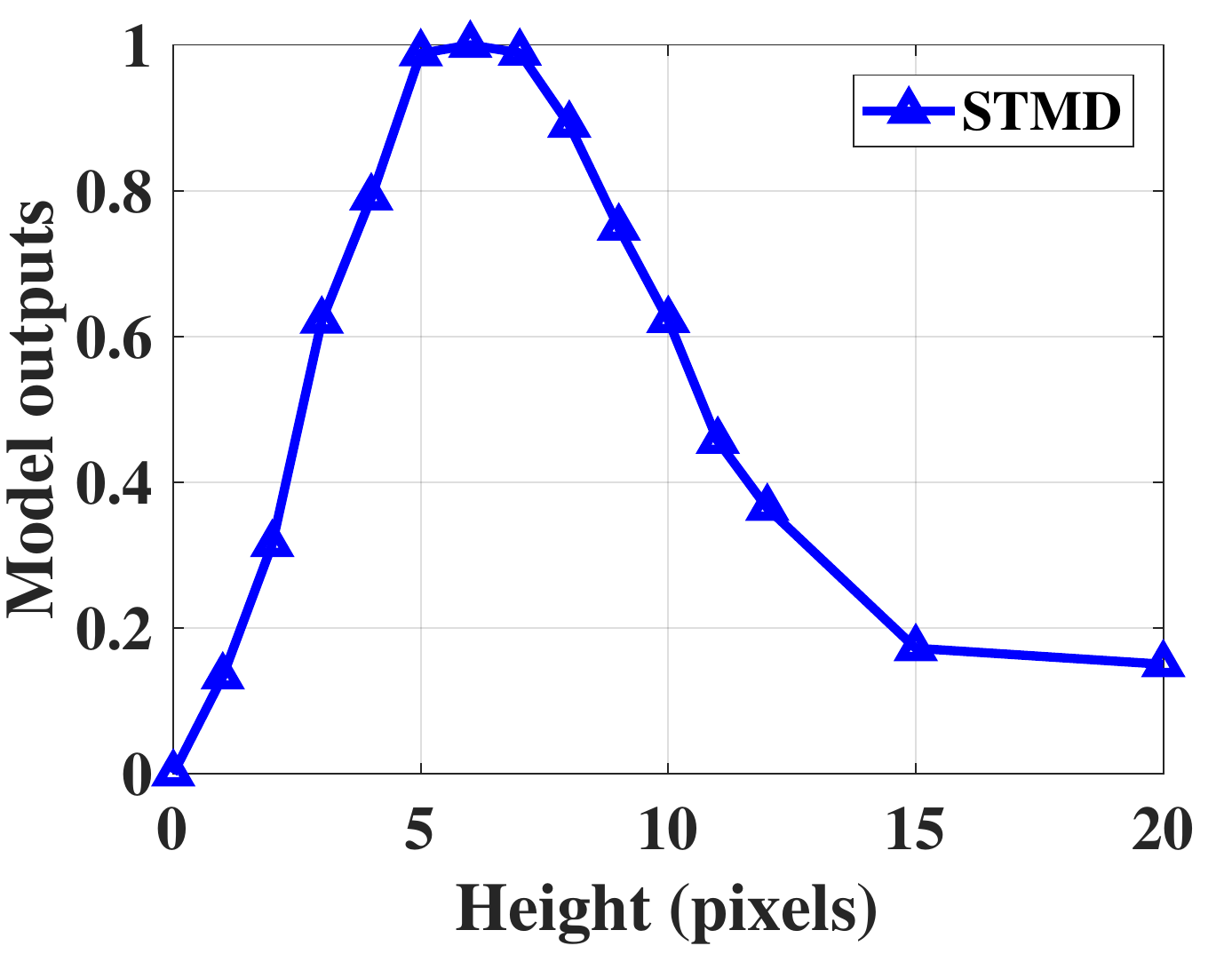}
		\label{Tuning-Properties-Height}}
	
	\caption{STMD outputs to moving objects with different Weber contrast, velocities, widths and heights. (a) Different Weber contrast. (b) Different velocities. (c) Different widths. (d) Different heights.}
	\label{Tuning-Properties-LPTC-STMD}
\end{figure}

To further demonstrate the characteristics of the STMD,  we compare its outputs to objects with different velocities, widths, heights and Weber contrast. As shown in Fig. \ref{The-External-Rectangle-and-Neighboring-Background-Rectangle-of-a-Small-Target}, width (or height) represents object length extended parallel (or orthogonal) to the motion direction. Weber contrast is defined by the following equation,
\begin{equation}
\text{Weber contrast} = \frac{|\mu_t - \mu_b|}{255}
\label{LDTB}
\end{equation}
where $\mu_t$ is the average pixel value of the object, while $\mu_b$ is the average pixel value in neighboring area around the object. If the size of a object is $w \times h$, the size of its background rectangle is $(w+2d)\times(h+2d)$, where $d$ is a constant which equals to $10$ pixels. The initial Weber contrast, velocity, width and height of the object are set as $1$, $250$ pixel/s, $5$ pixels and $5$ pixels, respectively. 

Fig. \ref{Tuning-Properties-LPTC-STMD}(a) shows the STMD output with respect to the  Weber contrast. As can be seen, the STMD output increases as the increase of Weber contrast, until reaches maximum at Weber contrast $=1$. This indicates that the higher Weber contrast of an object is, the easier it can be detected. Fig. \ref{Tuning-Properties-LPTC-STMD}(b) presents the STMD output with regard to the velocity of the moving object. Obviously, the STMD output peaks at an optimal velocity ($300$ pixel/s). The STMD also exhibits high responses to the objects whose velocities range from $100$ to $500$ pixel/s. Fig. \ref{Tuning-Properties-LPTC-STMD}(c) and (d) display the output of the STMD when changing the width and height of the object, which indicate that the STMD prefers moving objects whose widths and heights are smaller than $15$ pixels.

These characteristics of the STMD revealed in Fig. \ref{Tuning-Properties-LPTC-STMD}(a)-(d), are called Weber contrast sensitivity, velocity selectivity, width selectivity and height selectivity, respectively, which have been already found in the STMD neurons in biological research \cite{kelecs2017object,nordstrom2006insect,barnett2007retinotopic}.

\subsection{Effectiveness of the Contrast Pathway}

\begin{figure*}[!t]
    \vspace{-15pt}
	\centering
	\subfloat[]{\includegraphics[width=0.325\textwidth]{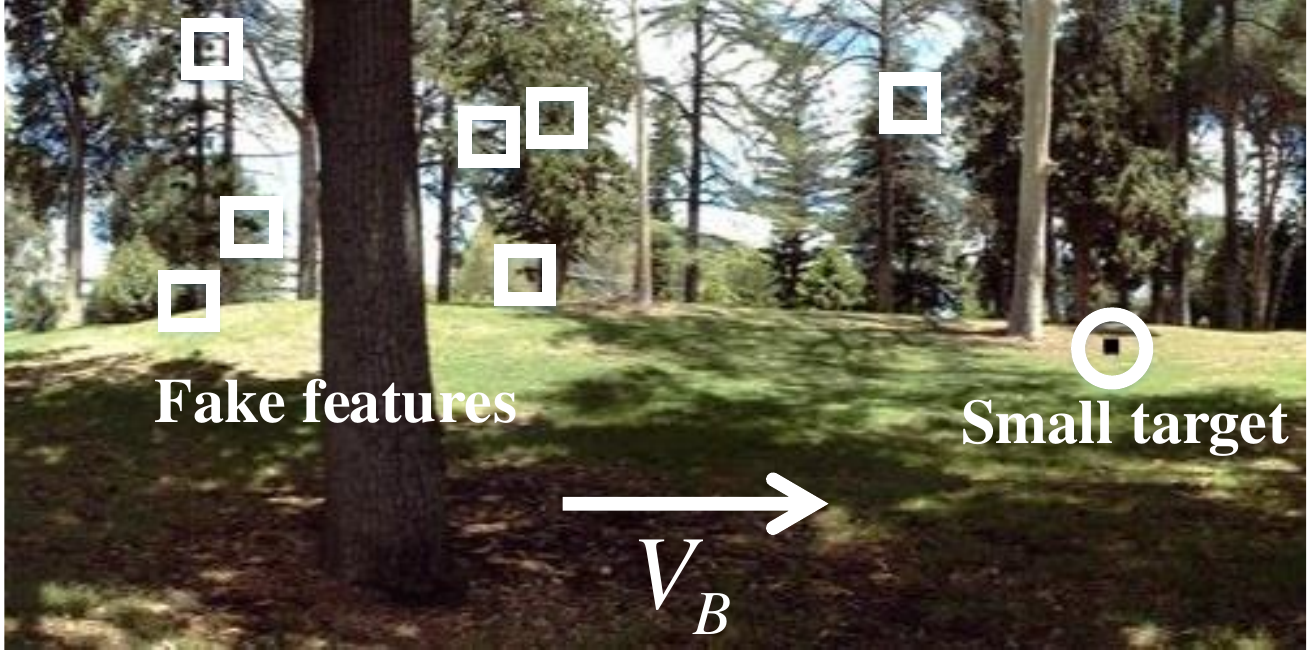}
		\label{Curvilinear-Motion-Original-Image}}
	\hspace{1em}
	\subfloat[]{\includegraphics[width=0.425\textwidth]{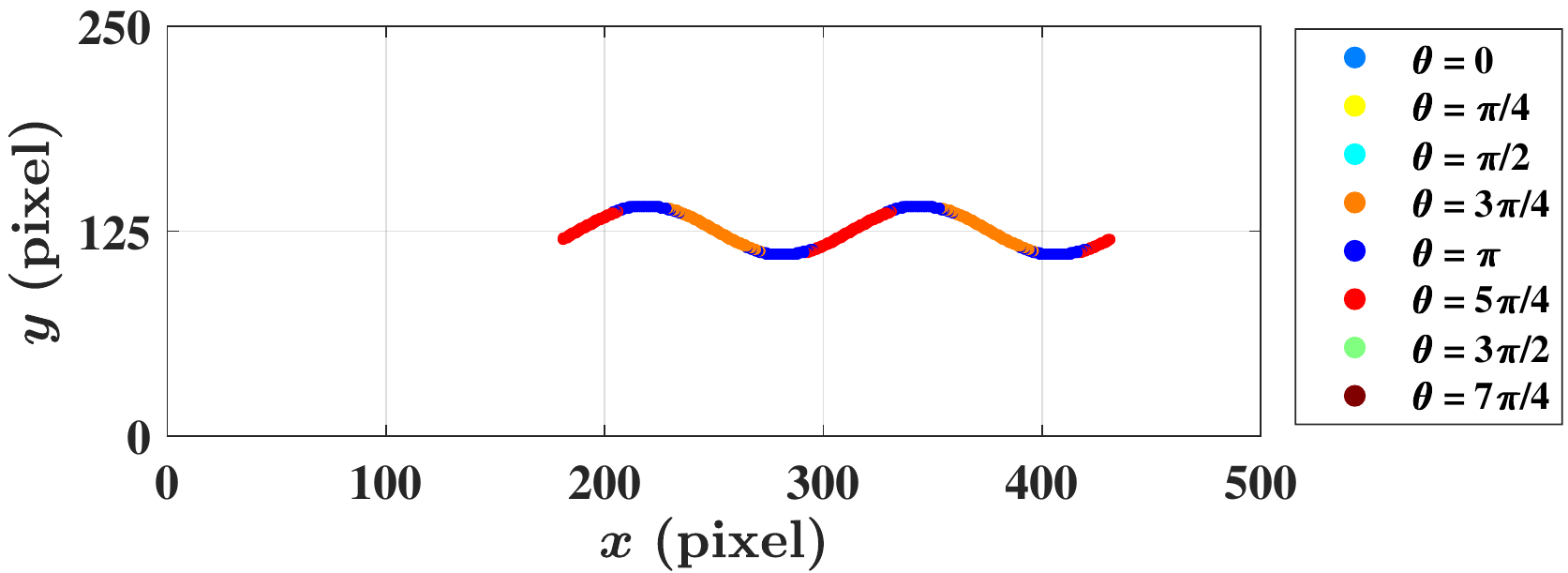}
		\label{Target-Trace-Threshold-350-Legend}}
	\caption{(a) Representative frame of the input image sequence. A small target (the small black block) highlighted by the circle, is moving against the cluttered background. The background which contains a number of fake features, is also moving from left to right where arrow $V_B$ denotes the background motion direction. (b) Motion trace of the small target during time period $[0,1000]$ ms, i.e., ground truth. In this subplot, color represents motion direction $\theta$ of the small target.}
	\label{Curvilinear-Motion-Original-Image-and-Target-Trace}
\end{figure*}

\begin{figure*}[!t]
	\vspace{-15pt}
	\centering
	\subfloat[]{\includegraphics[width=0.24\textwidth]{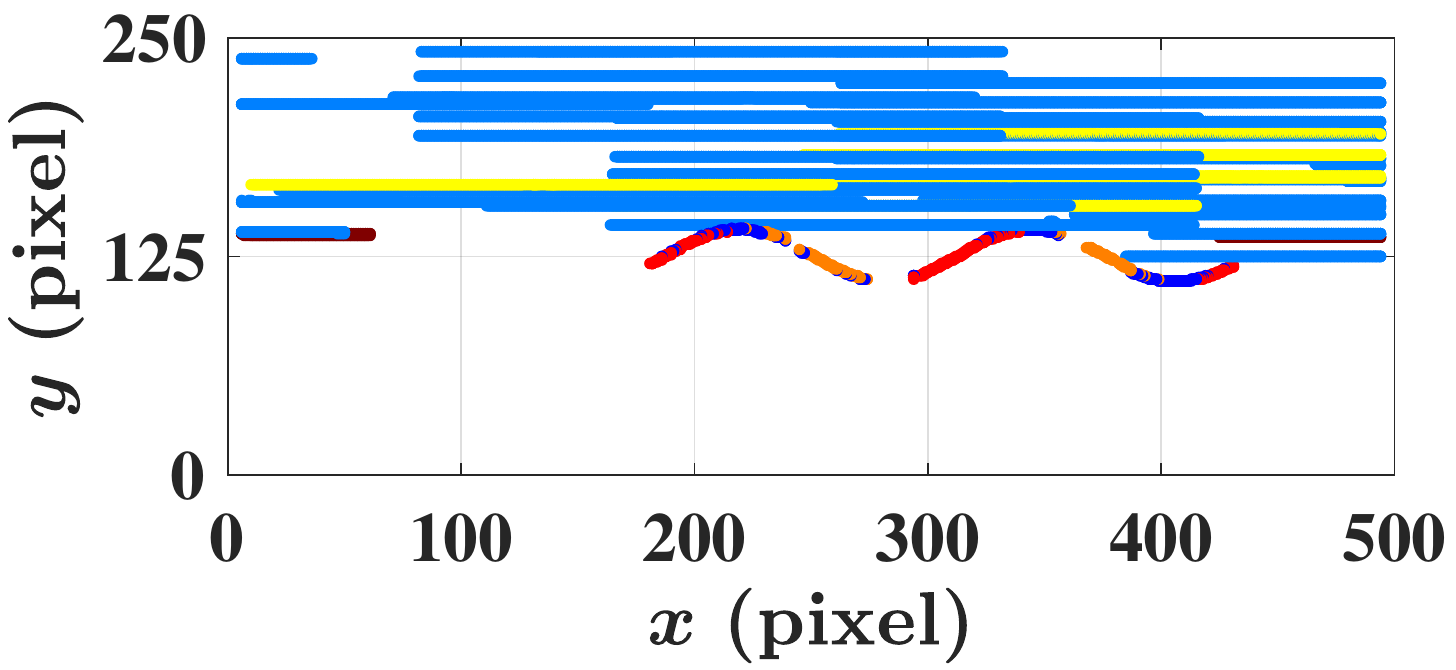}
		\label{CB-1-Target-Background-Noises-Trace-Threshold-150}}
	\hfill
	\subfloat[]{\includegraphics[width=0.24\textwidth]{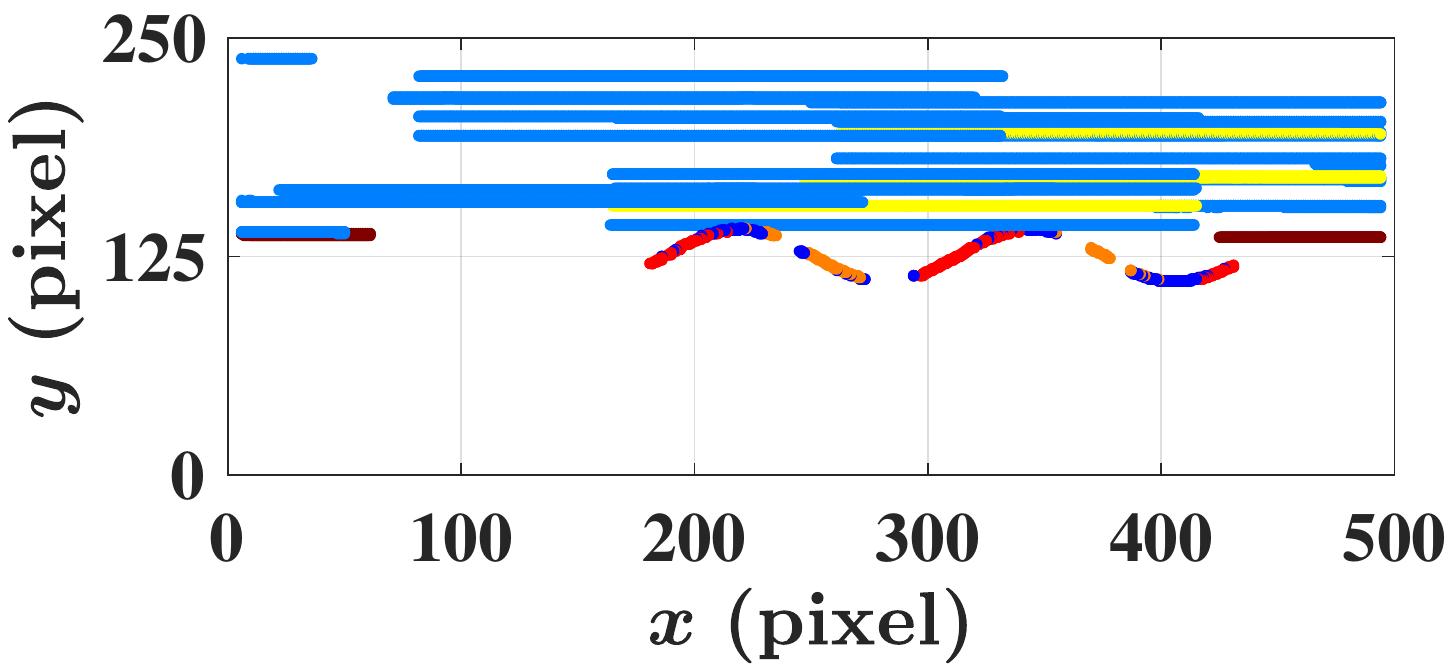}
		\label{CB-1-Target-Background-Noises-Trace-Threshold-250}}
	\hfill
	\subfloat[]{\includegraphics[width=0.24\textwidth]{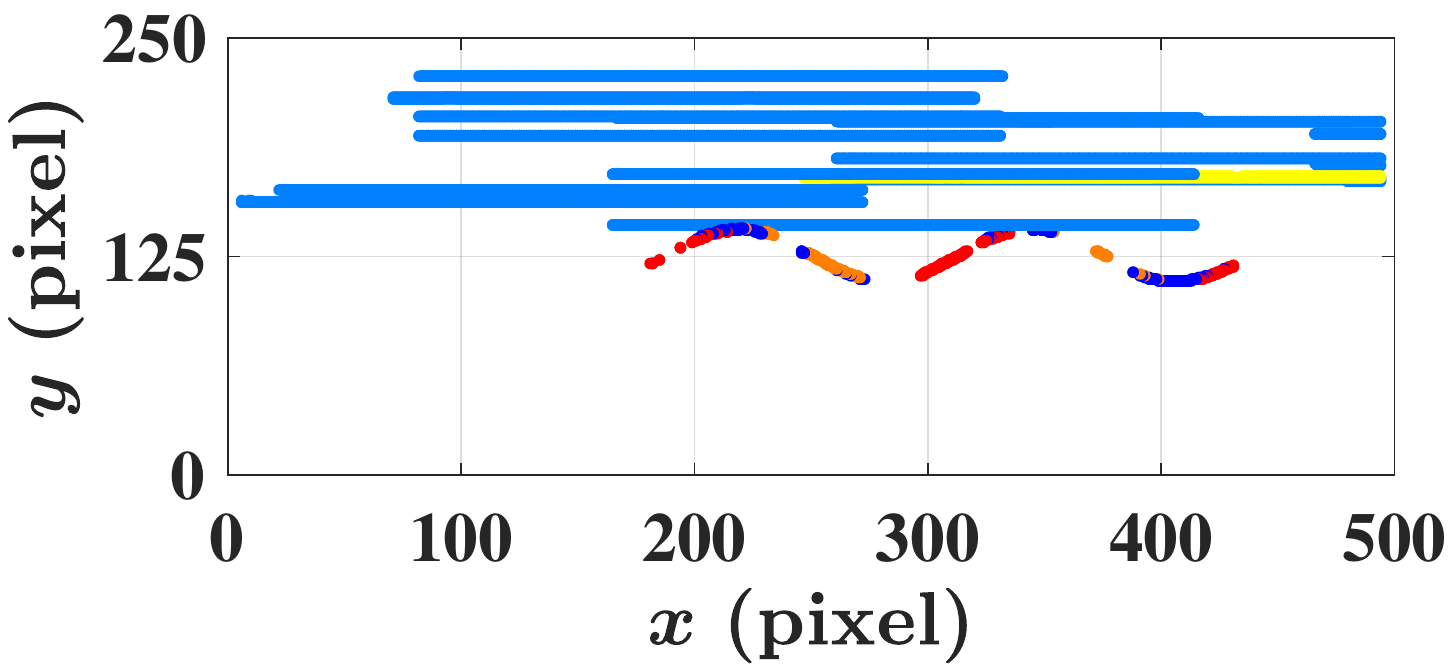}
		\label{CB-1-Target-Background-Noises-Trace-Threshold-350}}
	\hfill
	\subfloat[]{\includegraphics[width=0.24\textwidth]{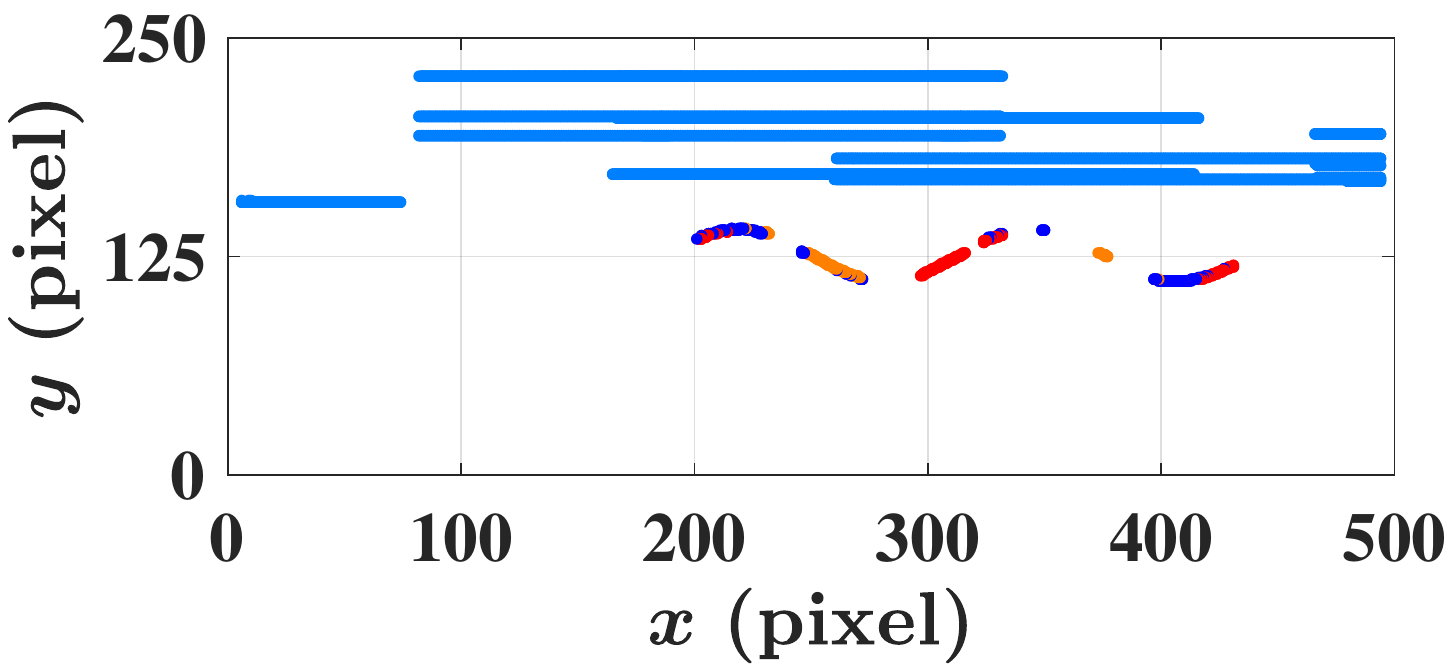}
		\label{CB-1-Target-Background-Noises-Trace-Threshold-450}}
	\hfill
	\subfloat[]{\includegraphics[width=0.24\textwidth]{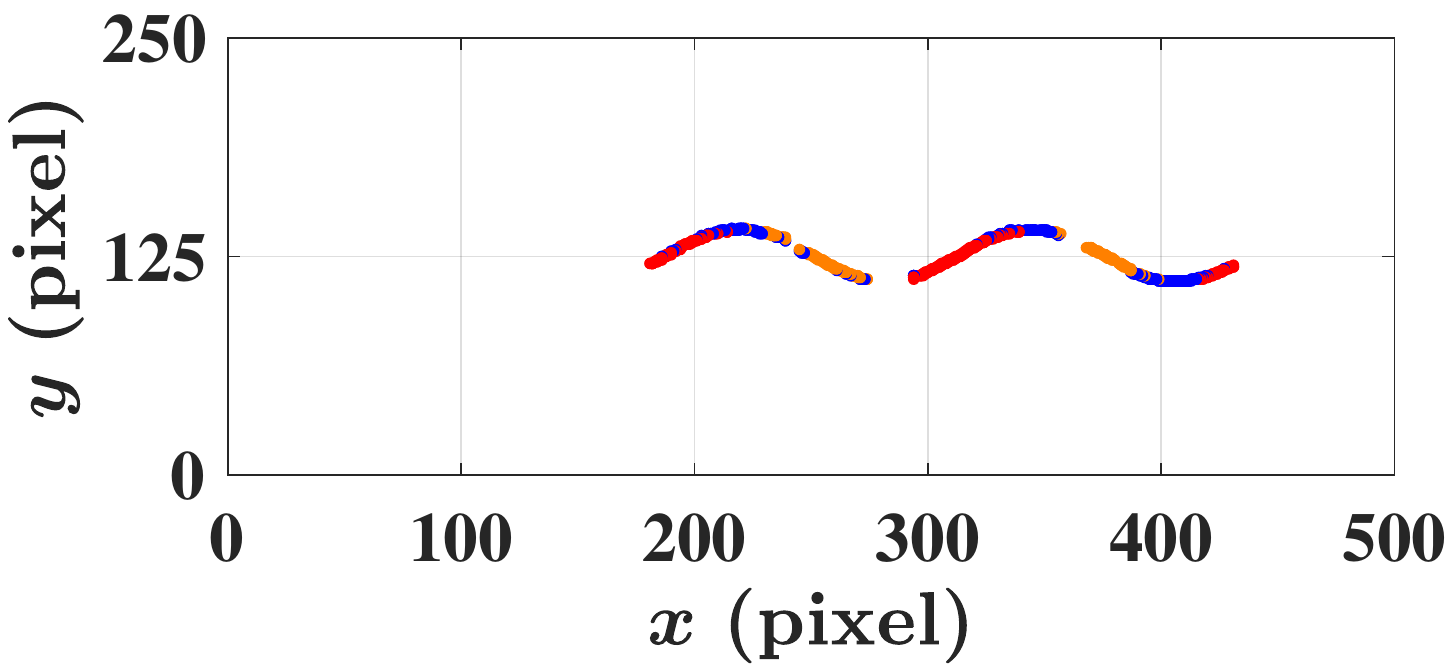}
		\label{CB-1-Target-Trace-Threshold-150}}
	\hfill
	\subfloat[]{\includegraphics[width=0.24\textwidth]{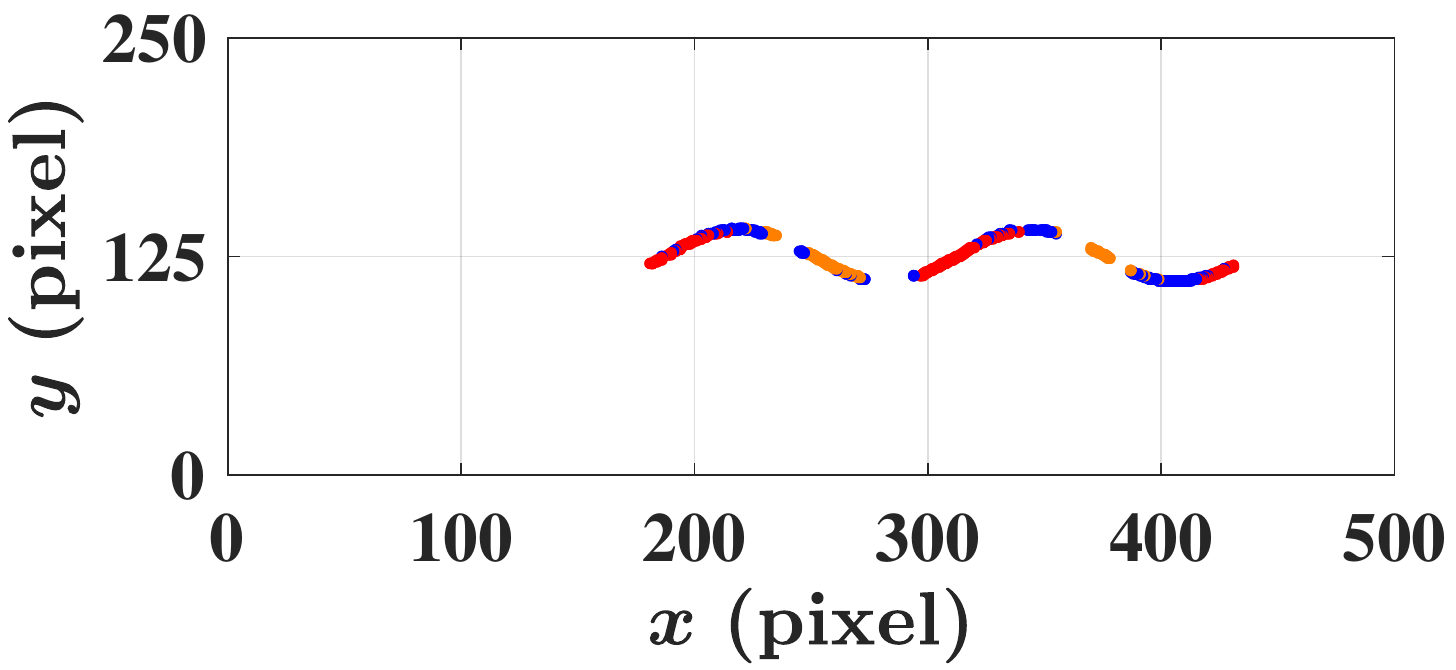}
		\label{CB-1-Target-Trace-Threshold-250}}
	\hfill
	\subfloat[]{\includegraphics[width=0.24\textwidth]{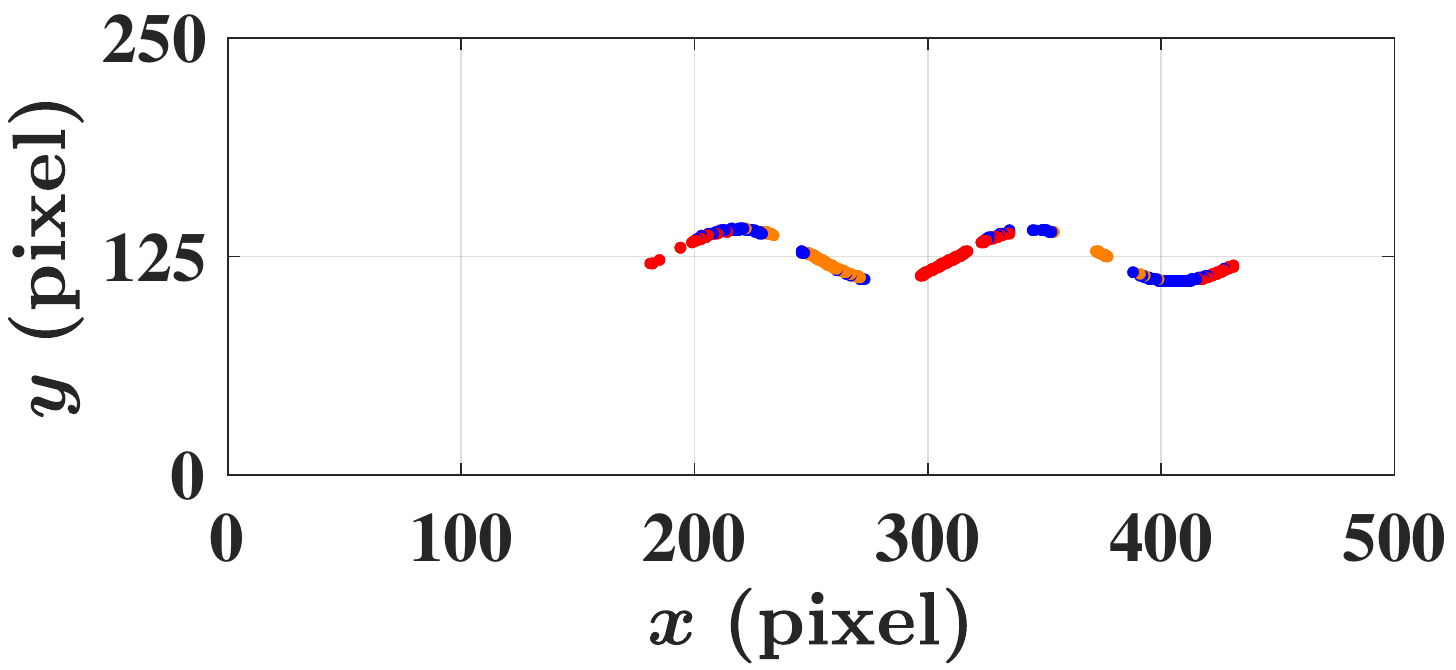}
		\label{CB-1-Target-Trace-Threshold-350}}
	\hfill
	\subfloat[]{\includegraphics[width=0.24\textwidth]{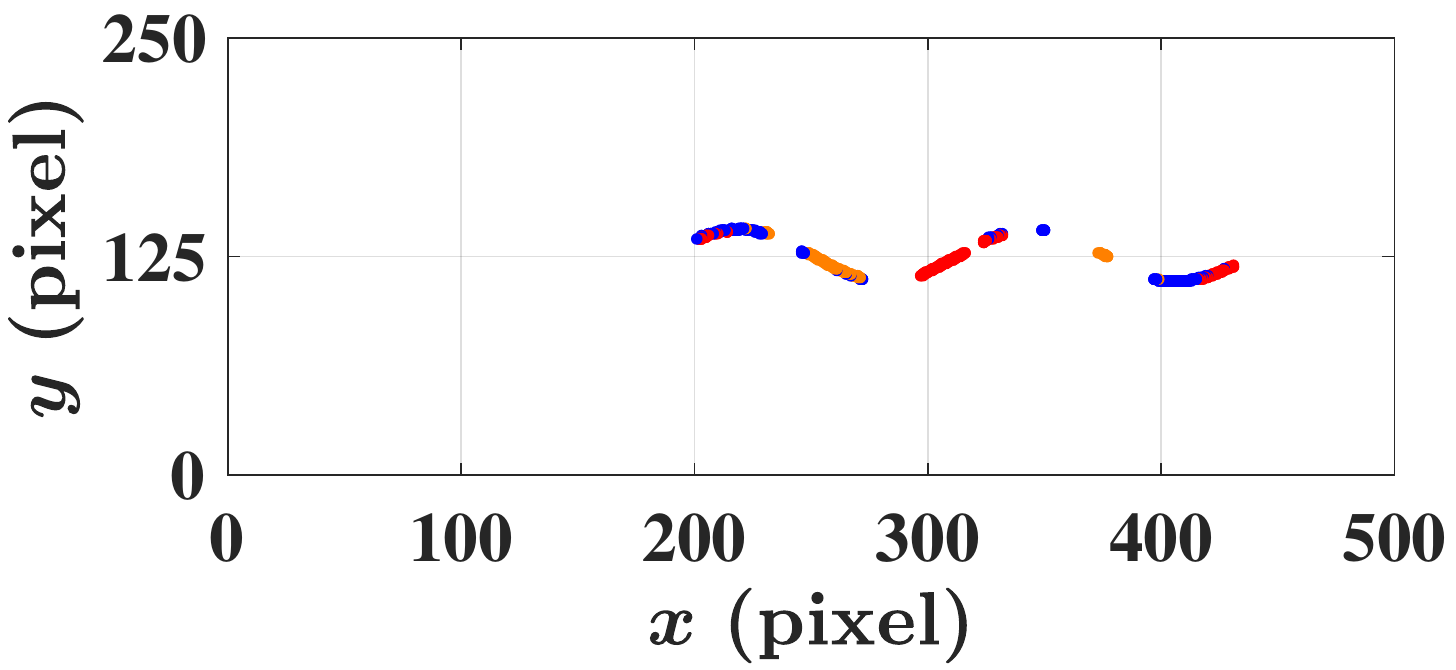}
		\label{CB-1-Target-Trace-Threshold-450}}
	\caption{(a)-(d) Motion traces detected by the STMD+ without the contrast pathway under different detection thresholds $\beta$ which are set as $150$, $250$, $350$ and $450$, respectively. (e)-(h) Motion traces detected by the STMD+ with the contrast pathway under different detection thresholds $\beta$ which are set as $150$, $250$, $350$ and $450$, respectively.}
	\label{CB-1-Target-Trace-STMD-Plus-Without-Contrast-Pathway}
\end{figure*}

\begin{figure*}[!t]
	\vspace{-15pt}
	\centering
	\subfloat[]{\includegraphics[width=0.24\textwidth]{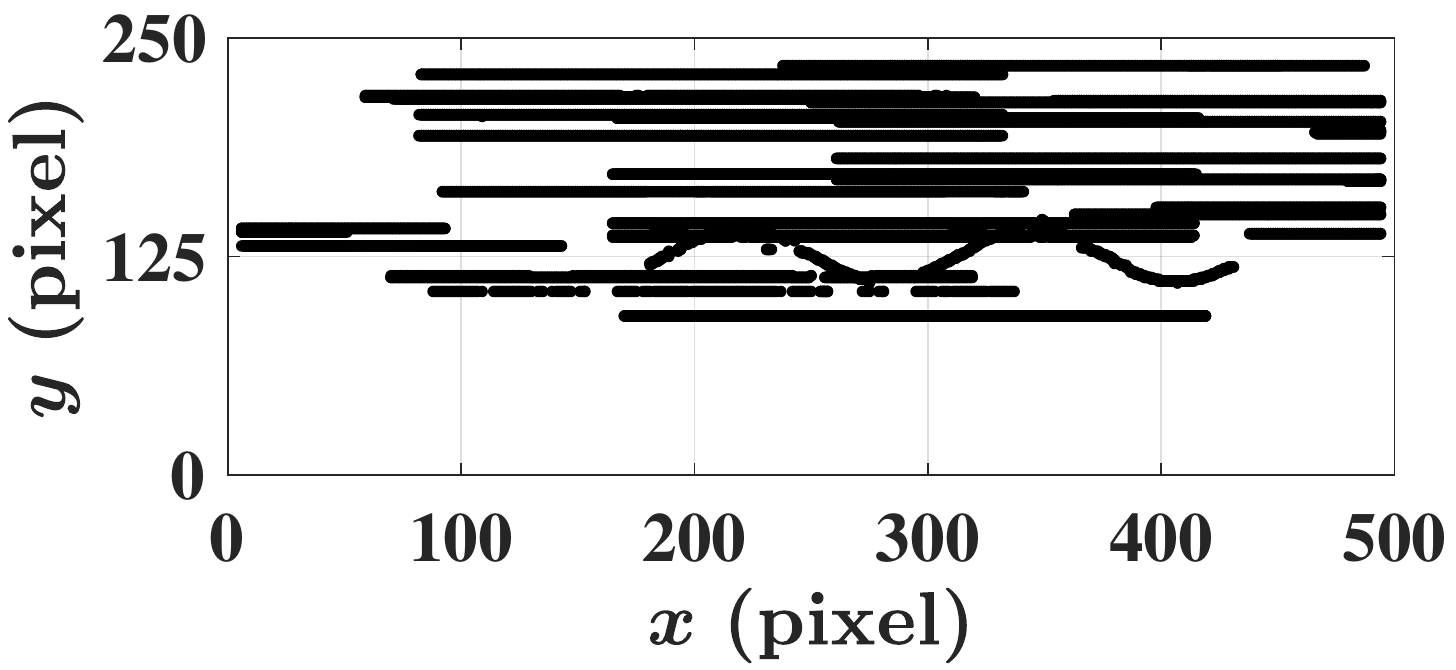}
		\label{CB-1-ESTMD-Traces-1-1000}}
	\hspace{1em}
	\subfloat[]{\includegraphics[width=0.24\textwidth]{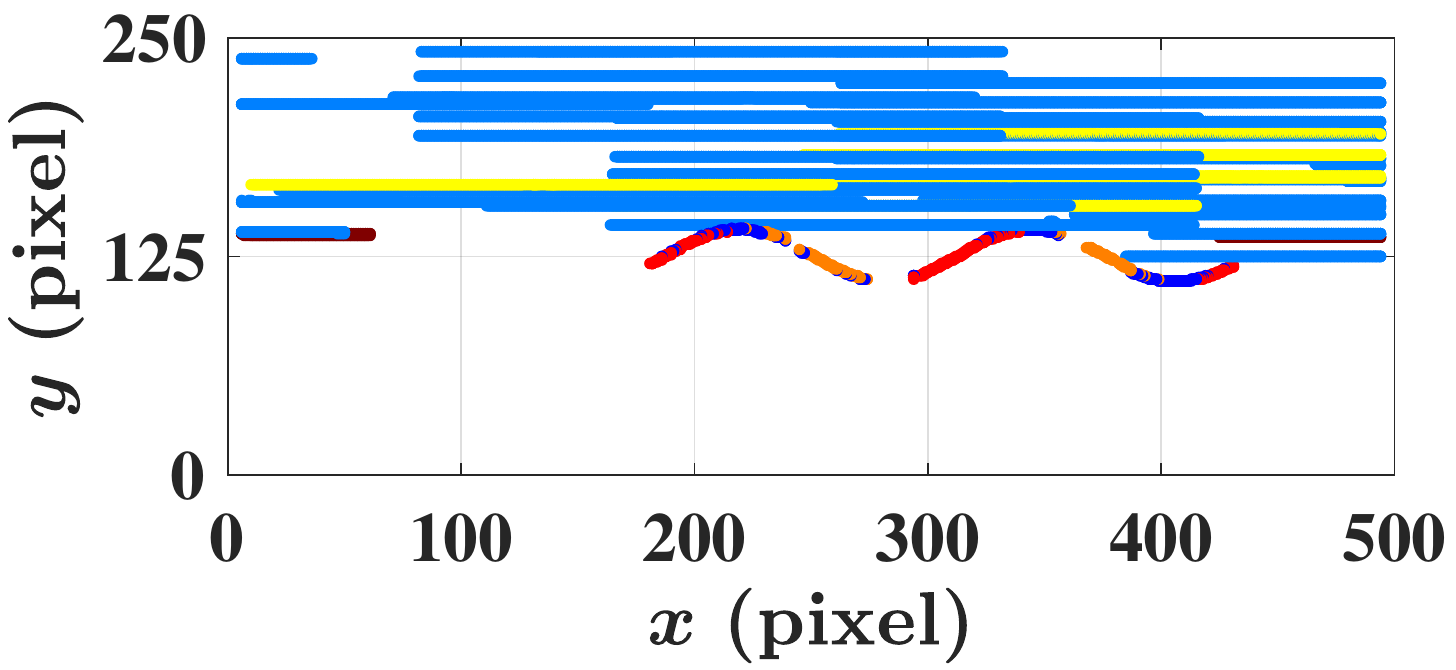}
		\label{CB-1-DSTMD-Traces-1-1000}}
	\hspace{1em}
	\subfloat[]{\includegraphics[width=0.24\textwidth]{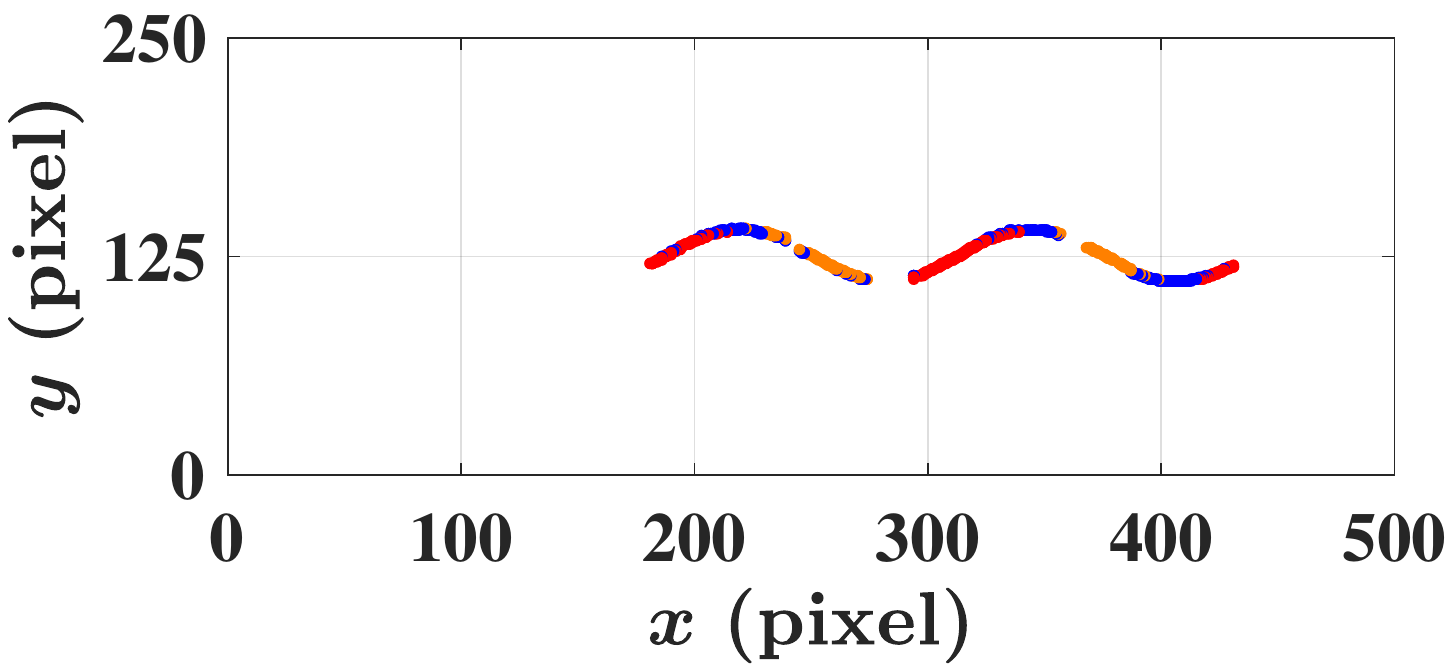}
		\label{CB-1-STMD-Plus-Traces-1-1000}}
	\caption{Motion traces detected by the ESTMD, DSTMD and STMD+. For fair comparison, the three models have fixed detection rates ($D_R = 0.85$). (a) ESTMD. (b) DSTMD. (c) STMD+. Since the ESTMD cannot detect motion direction, its outputs are all shown in black color.}
	\label{CB-1-Target-Trace-1-1000-Three-Models-Detection-results}
\end{figure*}

In the proposed visual system model, we design a contrast pathway and incorporate it with the motion pathway to discriminate small targets from fake features. To validate its effectiveness, we first compare the performance of the STMD+ with and without the contrast pathway. Then we conduct a performance comparison between the developed STMD+ and two baseline models including ESTMD \cite{wiederman2008model} and DSTMD \cite{wang2018directionally}. 
The testing setups are detailed as follows:  
the input image sequence is presented in Fig. \ref{Curvilinear-Motion-Original-Image-and-Target-Trace}(a), which displays a small target moving against the cluttered background; the background is moving from left to right and its velocity is $250$ pixel/s; the luminance, size and velocity of small target are equal to $0$, $5 \times 5$ pixels and $250$ pixel/s, respectively; the motion trace of the small target during time period $[0,1000]$ ms is illustrated in Fig. \ref{Curvilinear-Motion-Original-Image-and-Target-Trace}(b).

Fig. \ref{CB-1-Target-Trace-STMD-Plus-Without-Contrast-Pathway}(a)-(d) displays the motion traces detected by the STMD+ without the contrast pathway under different detection thresholds $\beta$. As can be seen, these detection results all contain numerous fake features. When increasing the detection threshold, the detected fake features will decrease while the detected motion trace becomes more incomplete. After applying the contrast pathway, the fake features are all filtered out even under different detection thresholds (see Fig. \ref{CB-1-Target-Trace-STMD-Plus-Without-Contrast-Pathway}(e)-(h)). The specific detection rate ($D_R$) and false alarm rate ($F_A$) are presented Table \ref{Table-Detection-Rate-False-Alarm-Rate-CB-1}. 

Fig. \ref{CB-1-Target-Trace-1-1000-Three-Models-Detection-results} demonstrates the motion traces detected by the ESTMD, DSTMD and STMD+, where the detection rates ($D_R$) of the three models are all set to $0.85$ for fair comparison. As can be seen, the detection results of the ESTMD and DSTMD are seriously contaminated by a number of fake features, whereas that of the STMD+ is noiseless.  
%

\begin{figure}[t]
	\vspace{-15pt}
	\centering
	\subfloat[]{\includegraphics[width=0.22\textwidth]{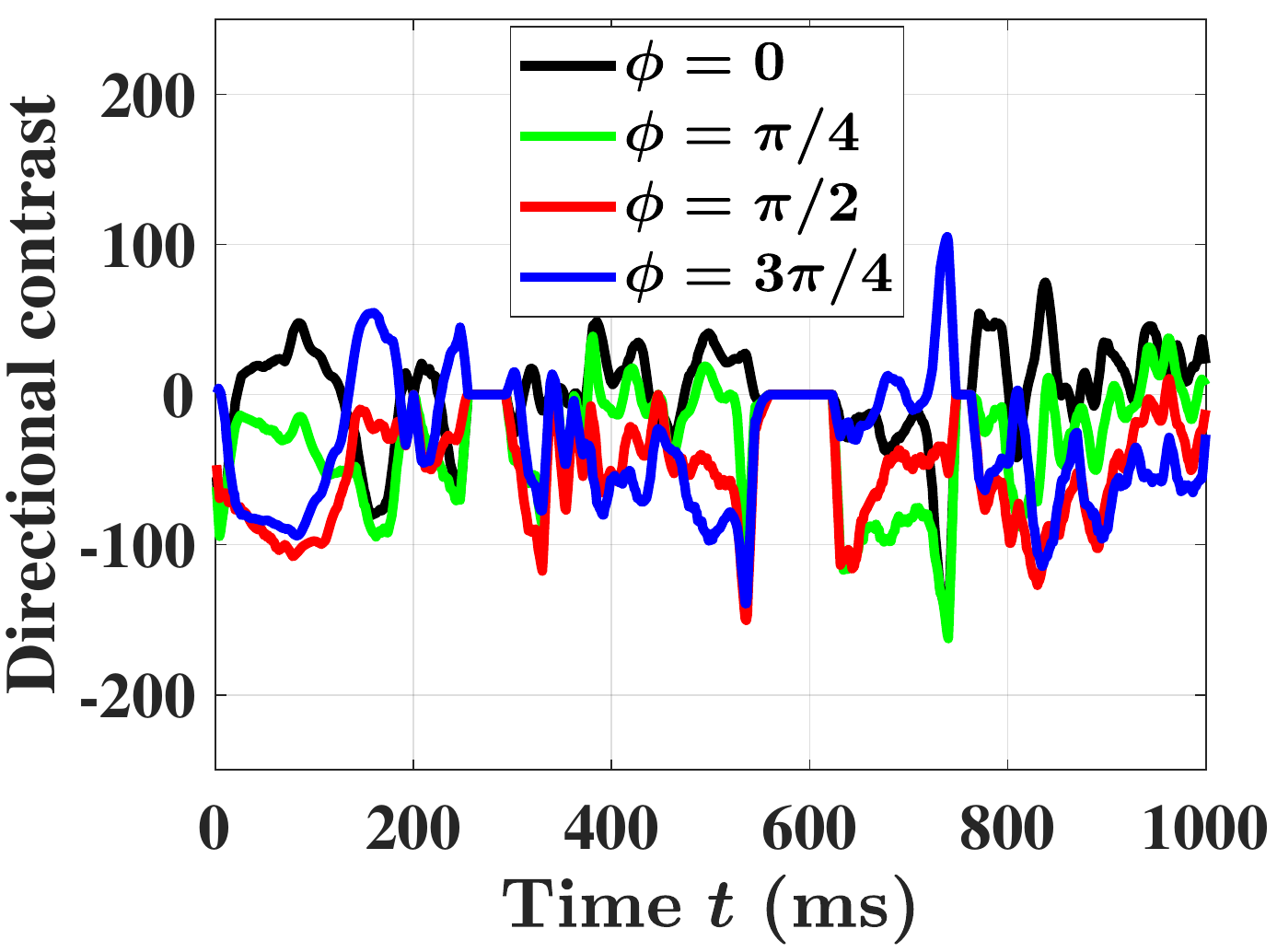}
		\label{CB-1-T1-Neural-Output-Target-Trace-Threshold-150}}
	\hfil
	\subfloat[]{\includegraphics[width=0.22\textwidth]{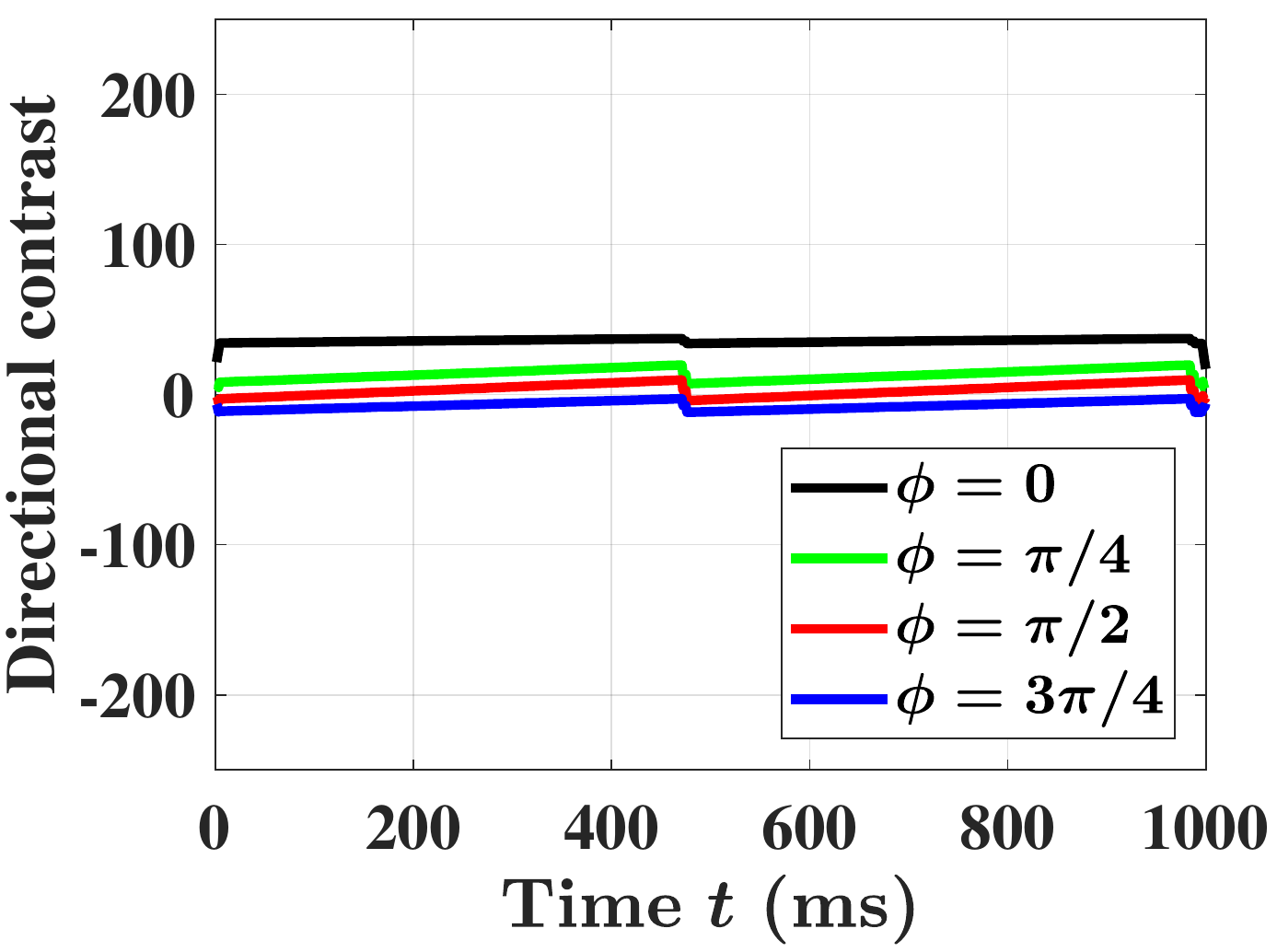}
		\label{CB-1-T1-Neural-Output-Trace-6-Threshold-150}}
	\caption{(a) Directional contrast on the motion trace caused by the {\bf small target}. (b) Directional contrast on the  motion trace caused by the {\bf fake feature}. In each subplot, the directional contrast along four directions $\phi \in \{0, \frac{\pi}{4}, \frac{\pi}{2}, \frac{3\pi}{4}\}$ is presented.}
	\label{CB-1-Spatial-Contrast-on-Motion-Trace}
\end{figure}

\begin{figure}[!t]
	\vspace{-15pt}
	\centering
	\subfloat[]{\includegraphics[width=0.22\textwidth]{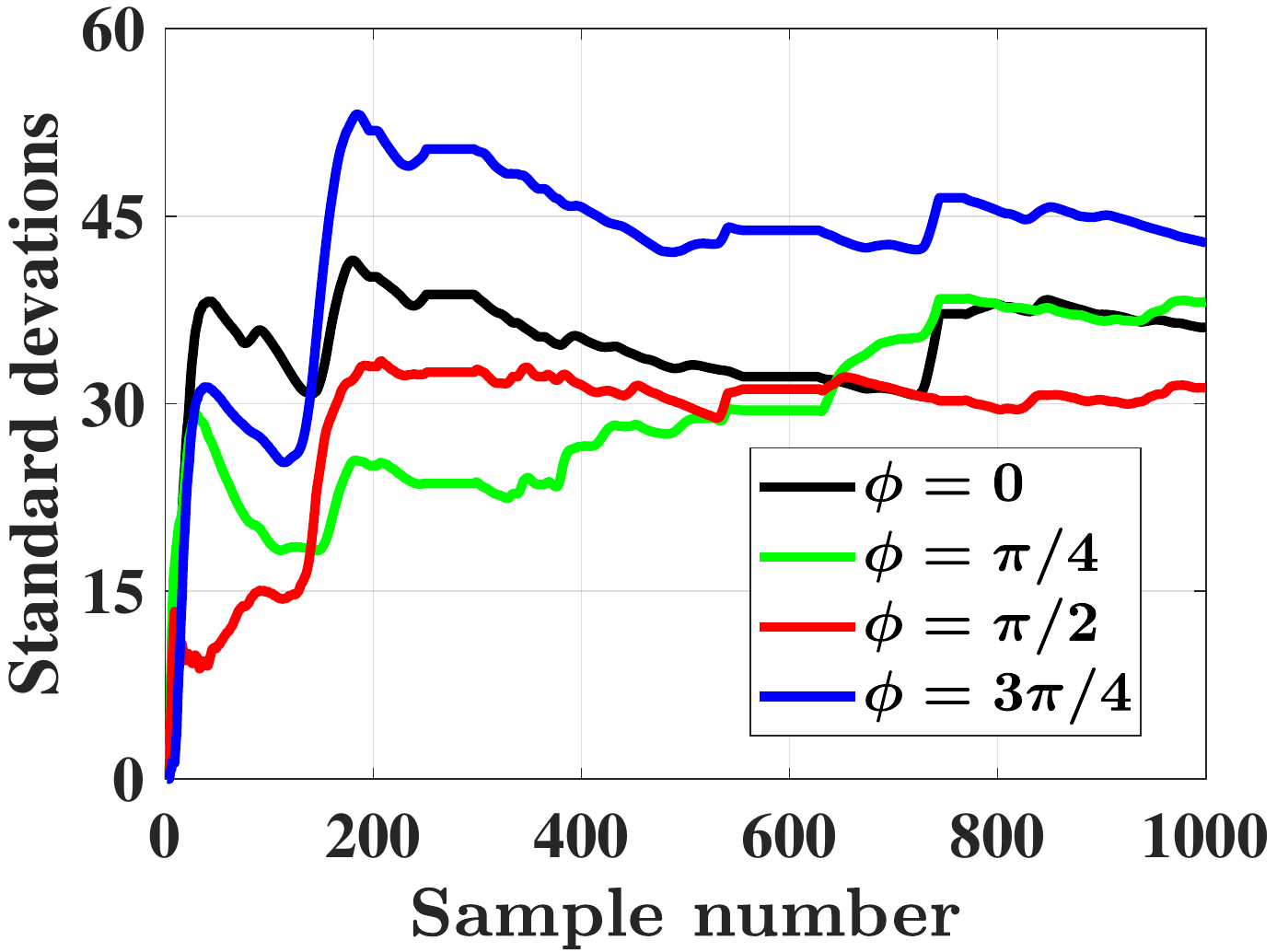}
		\label{CB-1-Frame-Length-T1-Output-STD-Target-Trace-Threshold-150}}
	\hfil
	\subfloat[]{\includegraphics[width=0.22\textwidth]{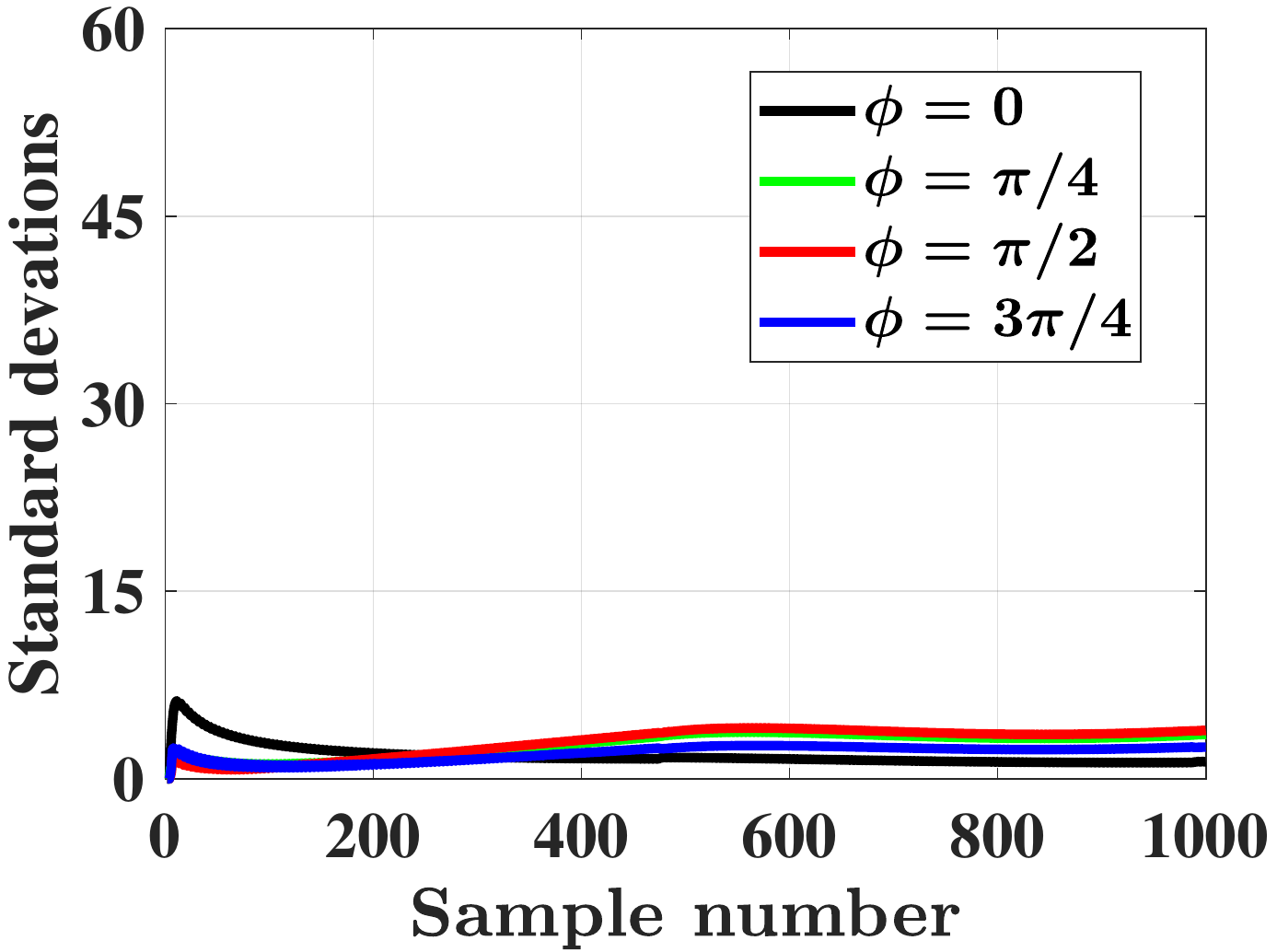}
		\label{CB-1-Frame-Length-T1-Output-STD-Trace-6-Threshold-150}}
	\caption{Standard deviations under different sample numbers. (a) Standard deviations of the {\bf small target}. (b) Standard deviations of the {\bf fake feature}.}
	\label{CB-1-T1-Neural-Outputs-STD-Frame-Length}
\end{figure}

\begin{table}[t]
	\renewcommand{\arraystretch}{1.3}
	\caption{Detection rate ($D_R$) and false alarm rate ($F_A$) of the STMD+ with and without the contrast pathway under different detection thresholds $\beta$.}
	\label{Table-Detection-Rate-False-Alarm-Rate-CB-1}
	\centering
	\begin{threeparttable}
		\begin{tabular}{|c|c|c|c|c|}
			\hline
			\multirow{2}{*}{Threshold $\beta$} &  
			\multicolumn{2}{c|}{Without\tnote{*}}& \multicolumn{2}{c|}{With\tnote{\#}} \cr \cline{2-5}
			&$D_R$ & $F_A$ &$D_R$ & $F_A$ \cr
			\hline  
			$150$ & $\bf{0.85}$ & $\bf{27.70}$ & $\bf{0.85}$ & $\bf{0}$ \cr\hline
			$250$ & $0.74$ & $19.11$ & $0.74$ & $0$  \cr\hline
			$350$ & $0.60$ & $12.87$ & $0.60$ & $0$  \cr\hline
			$450$ & $0.50$ & $6.88$  & $0.50$ & $0$  \cr\hline
		\end{tabular}
		
		\begin{tablenotes}
			\footnotesize
			\item[*] The STMD+ without the contrast pathway.
			\item[\#] The STMD+ with the contrast pathway.
		\end{tablenotes}
	\end{threeparttable}
	
\end{table}	

\begin{table}[t]
	\renewcommand{\arraystretch}{1.3}
	\caption{Standard deviations of the direction contrast.}
	\label{STD-Observation-Time-1000-ms}
	\centering
	\begin{tabular}{|c|c|c|c|c|}
		\hline
		\multirow{2}{*}{}&  
		\multicolumn{4}{c|}{Standard deviation} \cr \cline{2-5}
		& $\phi = 0$ & $\phi = \frac{\pi}{4}$ & $\phi = \frac{\pi}{2}$ & $\phi = \frac{3\pi}{4}$ \\	
		\hline
		Small target & $\bf{36.10}$ & $\bf{38.17}$ & $\bf{31.29}$ & $\bf{42.86}$\\
		\hline
		Fake feature  & $1.38$ & $3.56$ & $3.88$ & $2.54$\\
		\hline
	\end{tabular}
\end{table}

To reveal the role of the contrast pathway, we analyze the directional contrast on two motion traces chosen from Fig. \ref{CB-1-Target-Trace-STMD-Plus-Without-Contrast-Pathway}(a), where one is the small target motion trace, and the other is a randomly selected fake feature trace. Fig. \ref{CB-1-Spatial-Contrast-on-Motion-Trace} presents the directional contrast on these two motion traces. Note that  each motion trace has four directional contrast along four directions $\phi \in \{0, \frac{\pi}{4}, \frac{\pi}{2}, \frac{3\pi}{4}\}$. As shown in Fig. \ref{CB-1-Spatial-Contrast-on-Motion-Trace}(a), the directional contrast on the motion trace caused by the small target displays significant changes over time. In contrast, the directional contrast of the fake feature trace remains almost unchanged with respect to time (see Fig. \ref{CB-1-Spatial-Contrast-on-Motion-Trace}(b)). The calculated standard deviations of the directional contrast on these two motion traces are listed in Table \ref{STD-Observation-Time-1000-ms}, where the sample number $m$ is equal to $1000$. Obviously, the standard deviations of the small target are much larger than those of the fake feature, suggesting that the small target can be discriminated from fake features by comparing their standard deviations.

We further study the relationship of the standard deviations with regard to the sample number $m$ (see Fig. \ref{CB-1-T1-Neural-Outputs-STD-Frame-Length}). As it is shown, the standard deviations of the small target exhibit a sharp rise when the sample number increases from $0$ to $200$. With the continuous growth of the sample number,  the standard deviations tend to be stable. Similarly, the standard deviations of the fake feature become stable as the increase of the sample number. Above results indicate that a certain number of samples which is at least greater than $200$, is needed to obtain stable standard deviations.

\subsection{Comparison on Synthetic and Real Datasets}

\begin{table*}[t!]
	\renewcommand{\arraystretch}{1.3}
	\caption{Details of the initial image sequence and six groups of image sequences. Comparing to the initial image sequence, Group $1$ to $6$ are composed of image sequences with different parameters. }
	\label{Parameter-Seeting-of-Image-Sequence}
	\centering
	\begin{tabular}{|c|c|c|c|c|c|c|c|}
		\hline
		Parameter & Initial sequence & Group 1 & Group 2 & Group 3 & Group 4 & Group 5 & Group 6 \\
		\hline
		Target velocity (pixel/s)& 250 & $\mathbf{0 \sim 500}$& 250 & 250 & 250 & 250 & 250 \\	
		\hline
		Target size ($\text{pixel} \times \text{pixel}$) & $5 \times 5$ & $5 \times 5$ & $\mathbf{1 \times 1 \sim 12 \times 12}$ & $5 \times 5$ & $5 \times 5$ & $5 \times 5$ & $5 \times 5$ \\
		\hline
		Target luminance & $0$ & $0$ & $0$ & $\mathbf{0\sim 75}$ & $0$  & $0$ & $0$ \\
		\hline
		Background velocity (pixel/s) & $250$ & $250$ & $250$ & $250$ & $\mathbf{0\sim 500}$ & $\mathbf{0\sim 500}$ & $250$ \\
		\hline
		Background motion direction &rightward & rightward  & rightward & rightward & rightward  & \bf{leftward} & rightward \\
		\hline
		Background Image &  Fig.\ref{Curvilinear-Motion-Original-Image-and-Target-Trace}(a) &  Fig.\ref{Curvilinear-Motion-Original-Image-and-Target-Trace}(a) & Fig.\ref{Curvilinear-Motion-Original-Image-and-Target-Trace}(a) & Fig.\ref{Curvilinear-Motion-Original-Image-and-Target-Trace}(a) & Fig.\ref{Curvilinear-Motion-Original-Image-and-Target-Trace}(a) & Fig.\ref{Curvilinear-Motion-Original-Image-and-Target-Trace}(a) & \bf{Fig.\ref{Detection-Performance-Differnet-Backgrounds}(a) $\sim$ (c)} \\
		\hline
	\end{tabular}
\end{table*}	

\begin{figure*}[!t]
	\centering
	\subfloat[]{\includegraphics[width=0.275\textwidth]{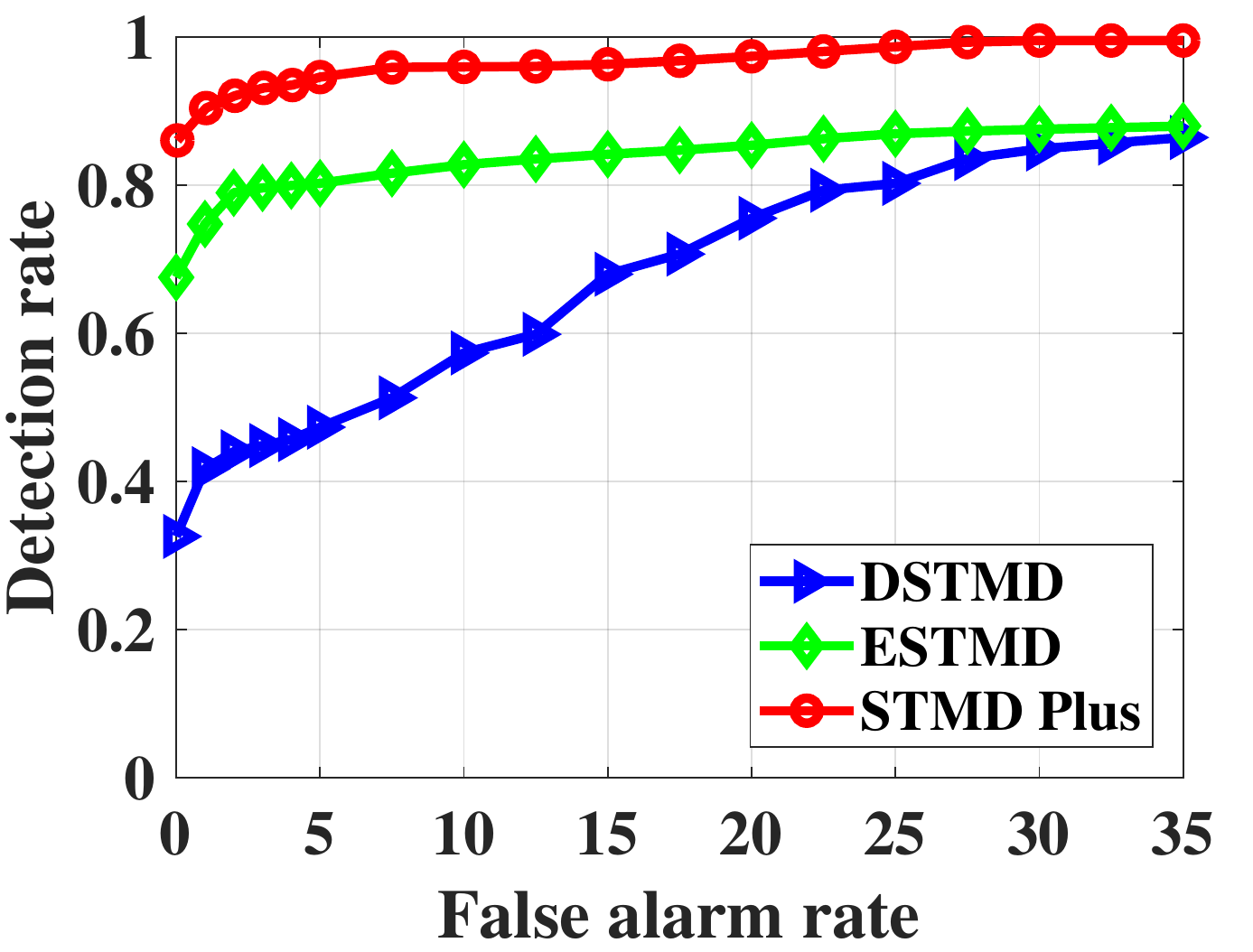}
		\label{CB-1-ROC-Curve-Three-Models-Initial-Parameters}}
	\hfil
	\subfloat[]{\includegraphics[width=0.275\textwidth]{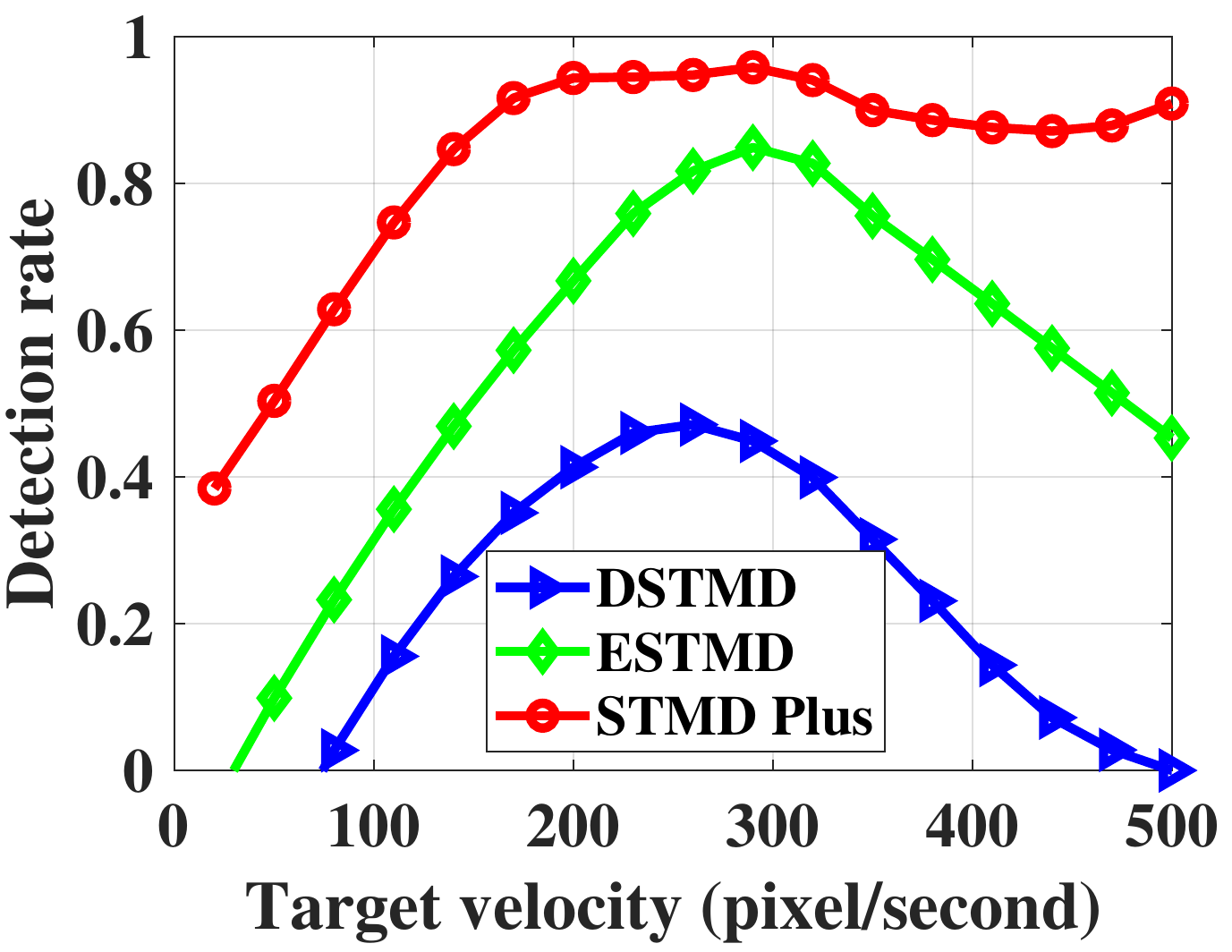}
		\label{CB-1-ROC-Curve-Three-Models-Varing-Target-Velocity}}
	\hfil
	\subfloat[]{\includegraphics[width=0.275\textwidth]{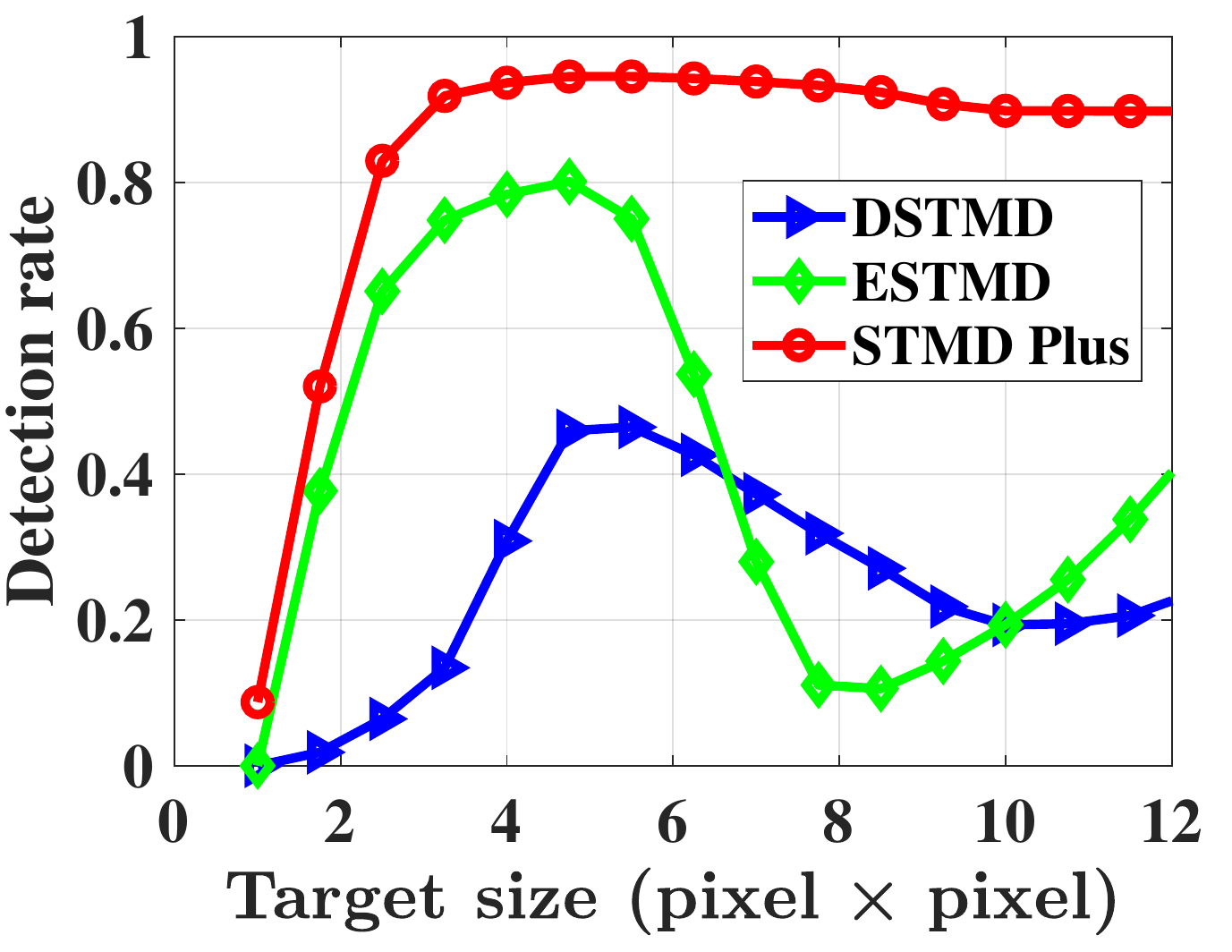}
		\label{CB-1-ROC-Curve-Three-Models-Varing-Target-Size}}
	\hfil
	\subfloat[]{\includegraphics[width=0.275\textwidth]{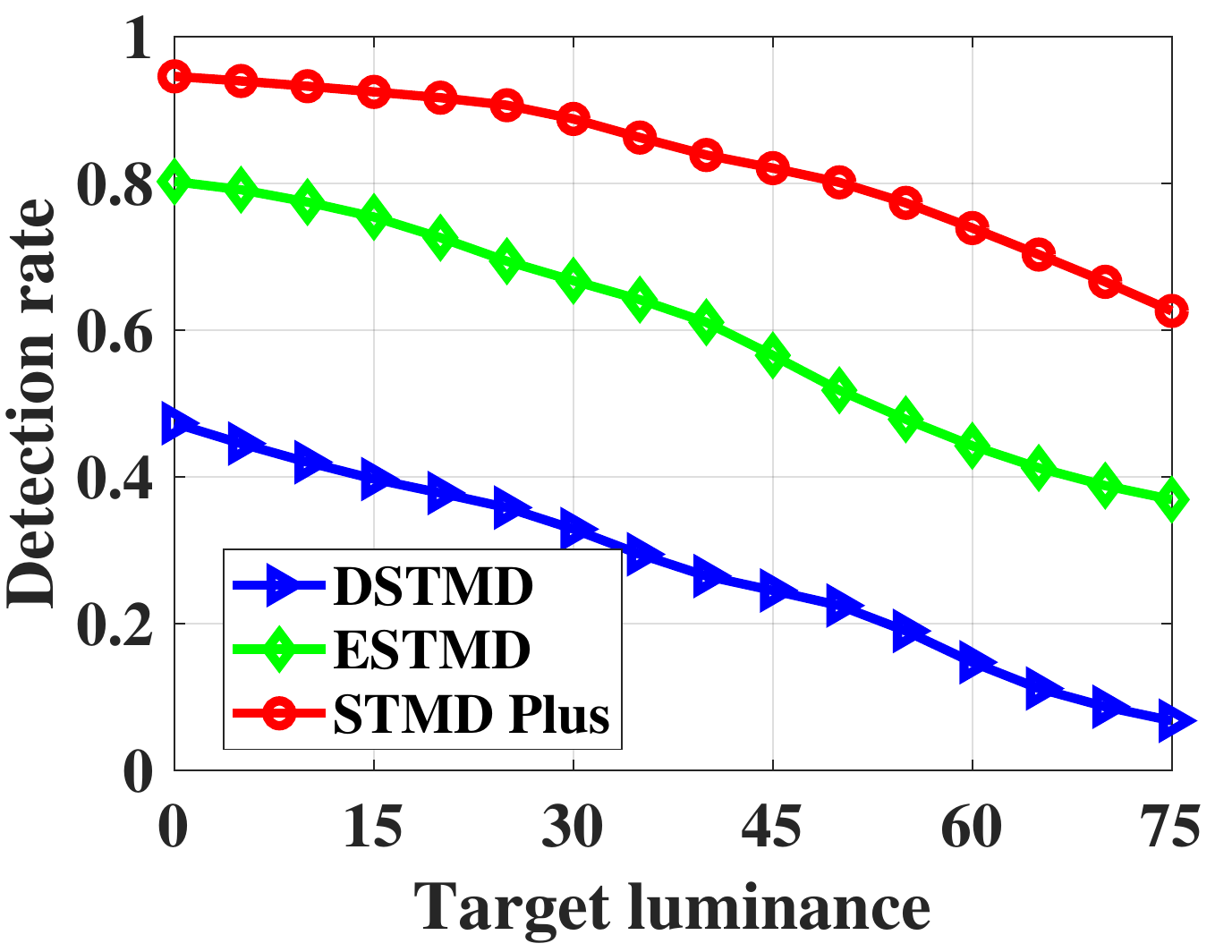}
		\label{CB-1-ROC-Curve-Three-Models-Varing-Target-Luminance}}
	\hfil
	\subfloat[]{\includegraphics[width=0.275\textwidth]{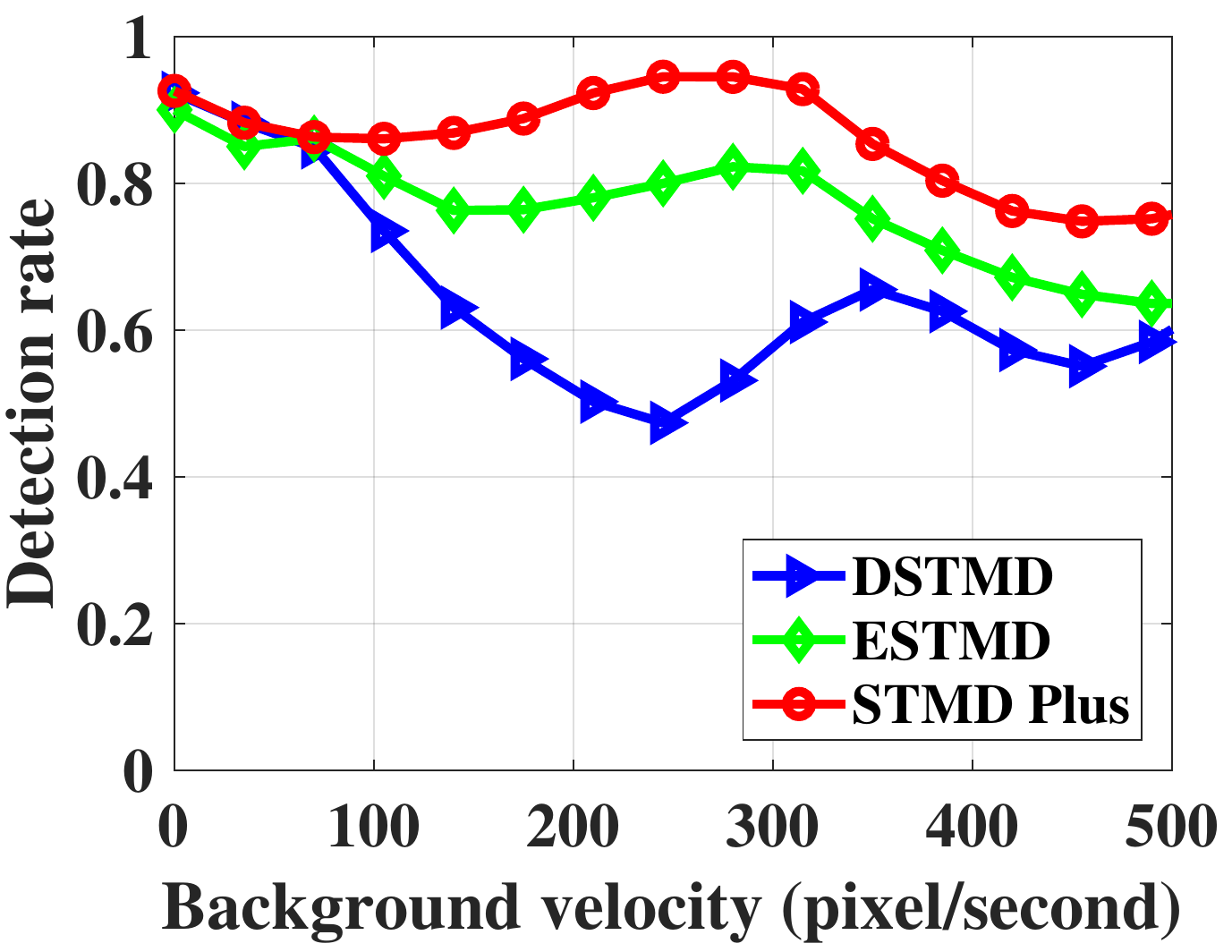}
		\label{CB-1-ROC-Curve-Three-Models-Varing-Background-Velocity-Opposite}}
	\hfil
	\subfloat[]{\includegraphics[width=0.275\textwidth]{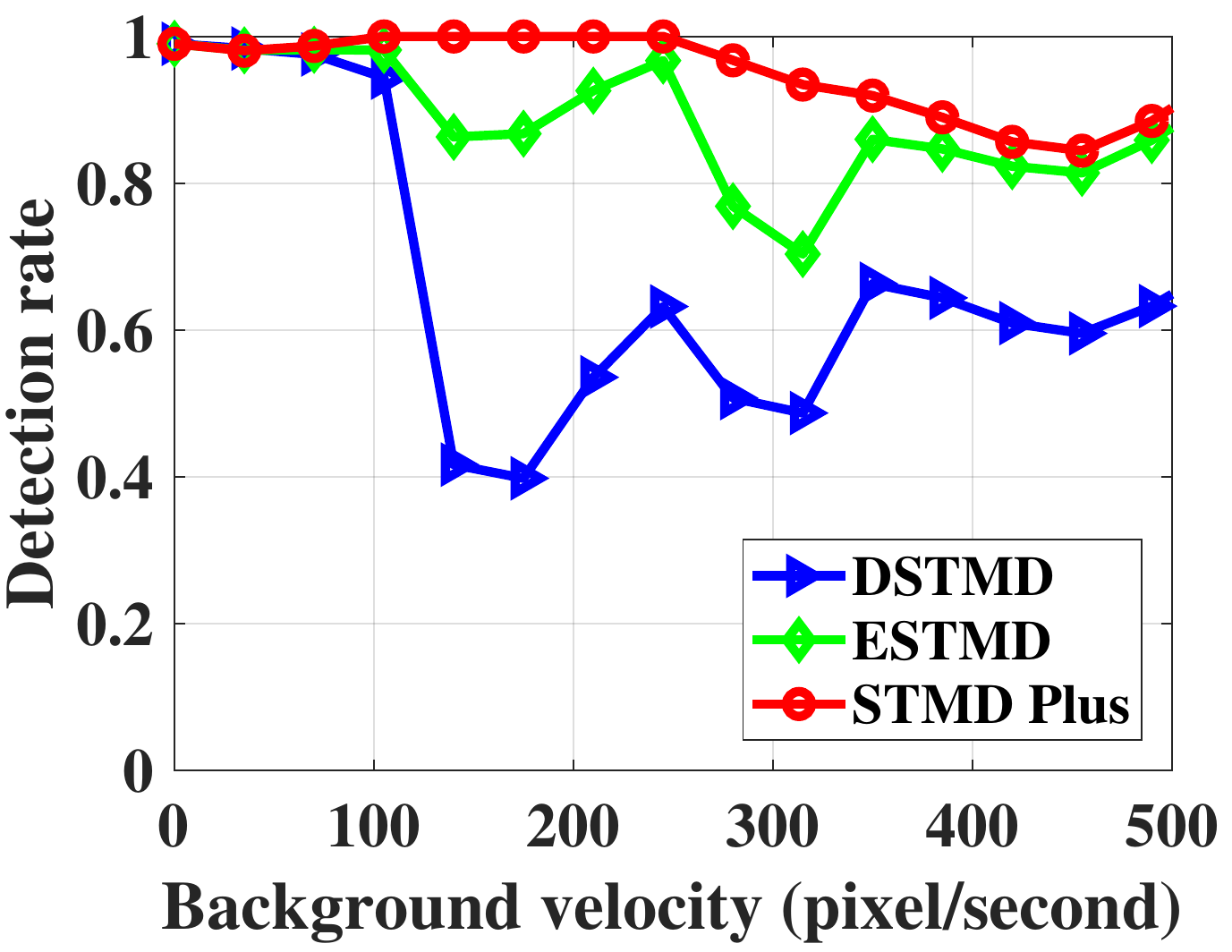}
		\label{CB-1-ROC-Curve-Three-Models-Varing-Background-Velocity-Same}}
	\caption{(a) Receiver operating characteristic (ROC) curves of the three models for the initial image sequence. (b)-(f) Detection rates of the three models for the Group 1-5. For fair comparison, the three models have fixed false alarm rate ($F_A = 5$). (b) Group 1, different target velocities. (c) Group 2, different target sizes. (d) Group 3, different target luminance. (e) Group 4, different background velocities (in rightward motion). (f) Group 5, different background velocities (in leftward motion).}
	\label{CB-1-LDTB-Size-Velocity-DR-FA}
\end{figure*}

In this section, six groups of synthetic image sequences are first utilized to evaluate the performance of the proposed model in terms of different target velocities, target sizes, target luminance, background velocities, background motion directions and background images. The details of the synthetic image sequences are listed in Table \ref{Parameter-Seeting-of-Image-Sequence}. Then the proposed model is further tested on the real dataset (STNS dataset \cite{bagheri2017performance}). The performance comparison between the proposed STMD+ and two baseline models (namely, ESTMD and DSTMD), is also conducted.

Fig. \ref{CB-1-LDTB-Size-Velocity-DR-FA}(a) shows the receiver operating characteristics (ROC) curves of the three models for the initial synthetic image sequence. It can be seen that the STMD+ has better performance than the baseline models. More precisely, the STMD+ has higher detection rates ($D_R$) compared to the baseline models while the false alarm rates ($F_A$) are low. Fig. \ref{CB-1-LDTB-Size-Velocity-DR-FA}(b)-(d) display the detection rates of the three models for the Group $1$ to $5$, where the false alarm rates of the three models are all equal to $5$ for fair comparison. From Fig. \ref{CB-1-LDTB-Size-Velocity-DR-FA}(b) and (c), we can see that the STMD+ significantly outperforms the baseline models. The STMD+ has higher detection rates than the baseline models for different target velocities and sizes. The detection rate of the STMD+ remains stable when the target velocity (or size) ranges from $200$ to $500$ pixel/s (or from $4 \times 4$ to $12 \times 12$ $\text{pixel} \times \text{pixel}$). In contrast, the detection rates of the two baseline models significantly decrease after reach the maximum points. As it shown in Fig. \ref{CB-1-LDTB-Size-Velocity-DR-FA}(d), the STMD+ consistently performs best under different target luminance. It is worthy to note that the detection rates of the three models all decrease with the increase of target luminance. In Fig. \ref{CB-1-LDTB-Size-Velocity-DR-FA}(e) and (f), we can see that the STMD+ has the better performance than the baseline models under different background velocities and directions.


\begin{figure*}[!t]
	\centering
	\subfloat[]{\includegraphics[width=0.285\textwidth]{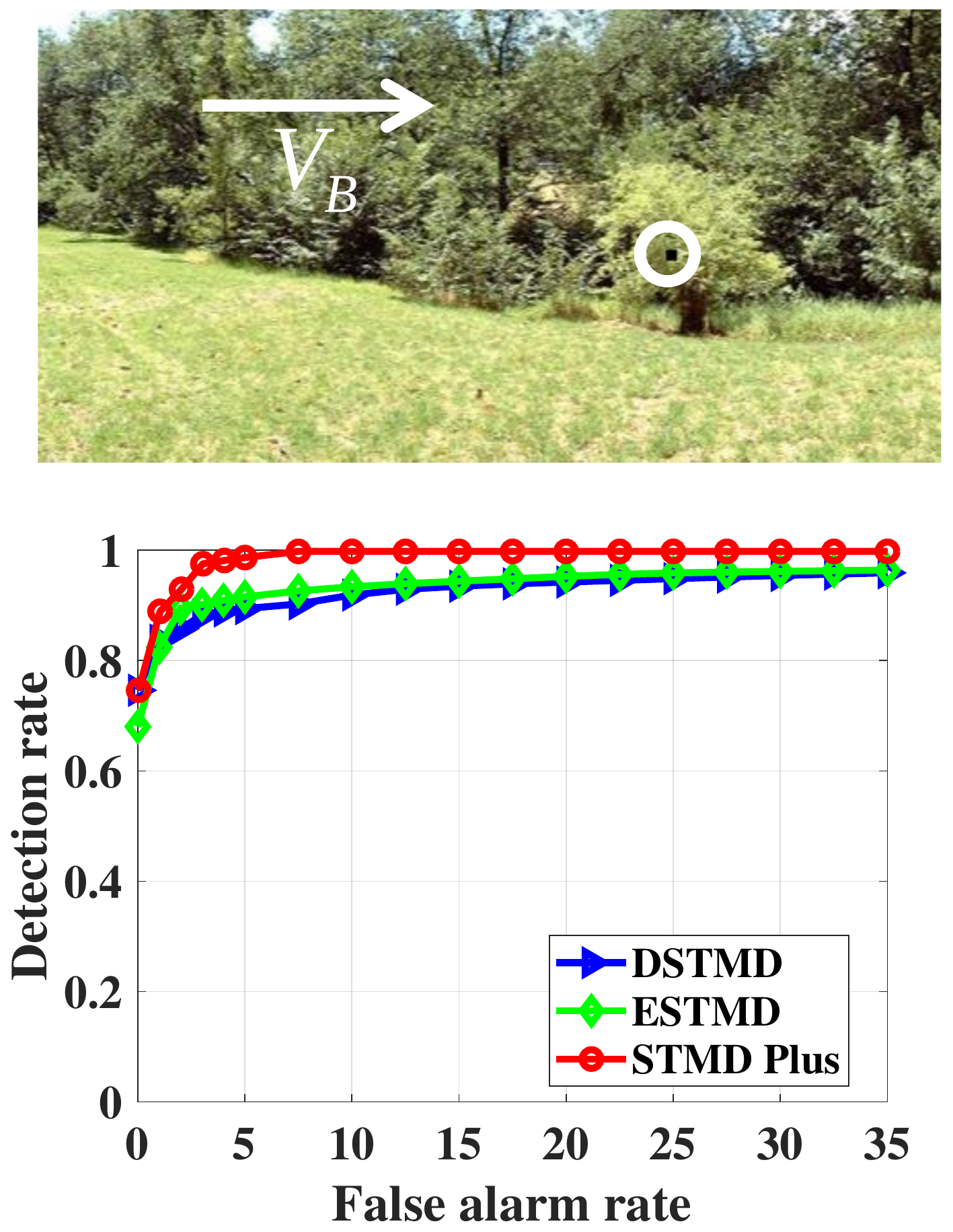}
		\label{CB-2-Frame-ROC}}
	\hfil
	\subfloat[]{\includegraphics[width=0.285\textwidth]{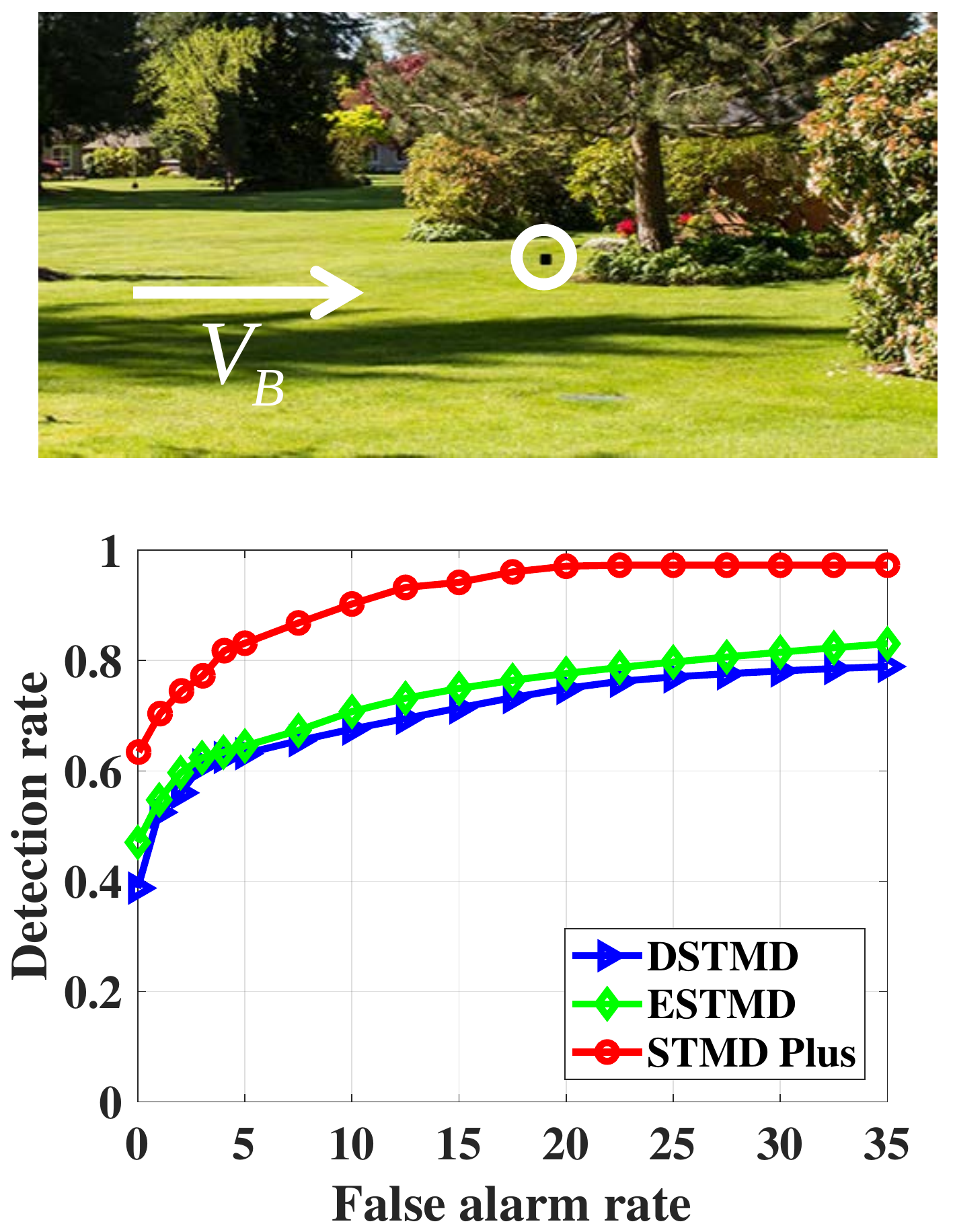}
		\label{CB-4-Frame-ROC}}
	\hfil
	\subfloat[]{\includegraphics[width=0.285\textwidth]{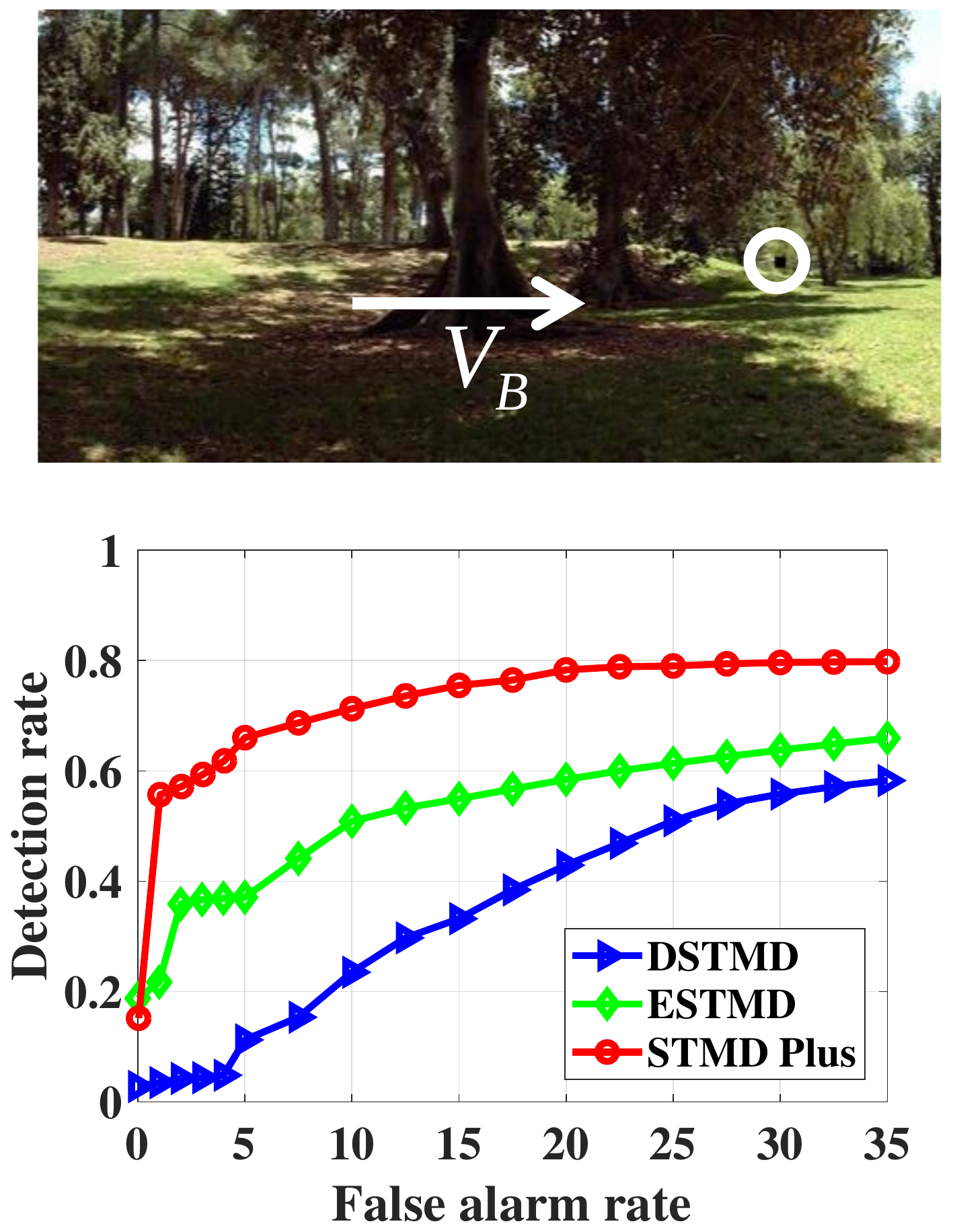}
		\label{CB-3-Frame-ROC}}
	\caption{Background images and receiver operating characteristic (ROC) curves of the three models for the Group 6, different backgrounds.}
	\label{Detection-Performance-Differnet-Backgrounds}
\end{figure*}

\begin{figure*}[t!]
	\centering
	\subfloat[]{\includegraphics[width=0.275\textwidth]{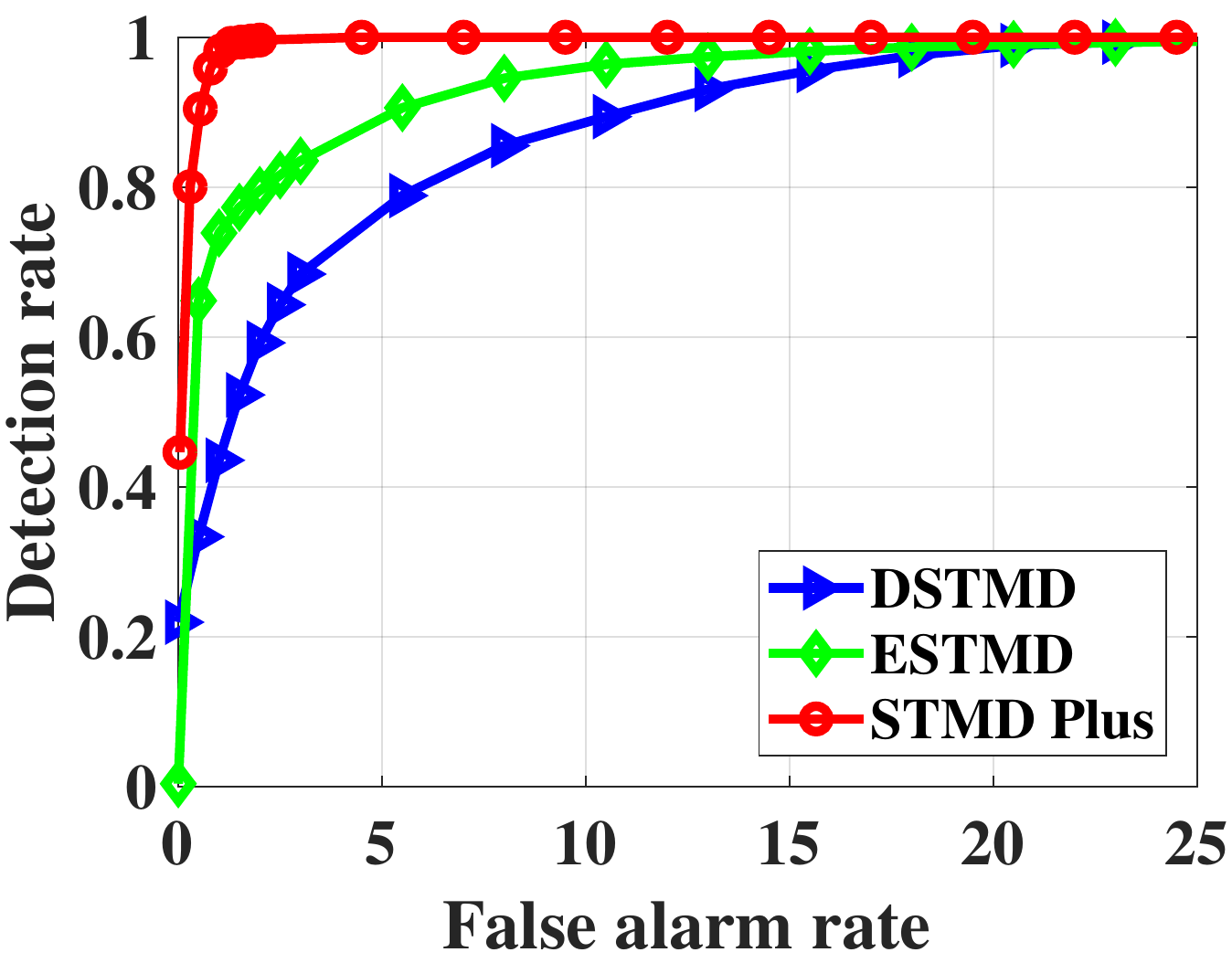}
		\label{Real-Dataset-STNS-4-ROC-Curve}}
	\hfil
	\subfloat[]{\includegraphics[width=0.275\textwidth]{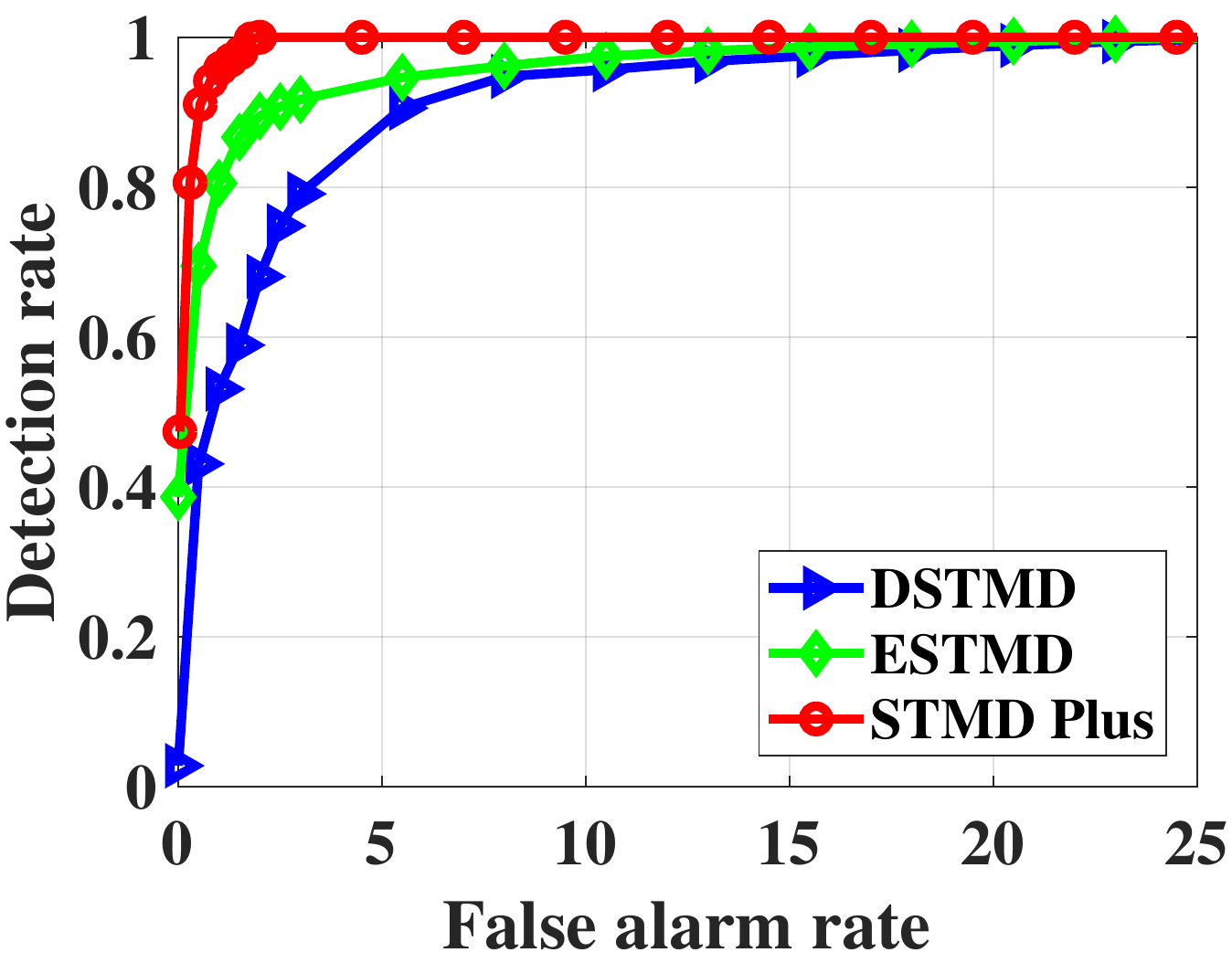}
		\label{Real-Dataset-STNS-15-ROC-Curve}}
	\hfil
	\subfloat[]{\includegraphics[width=0.275\textwidth]{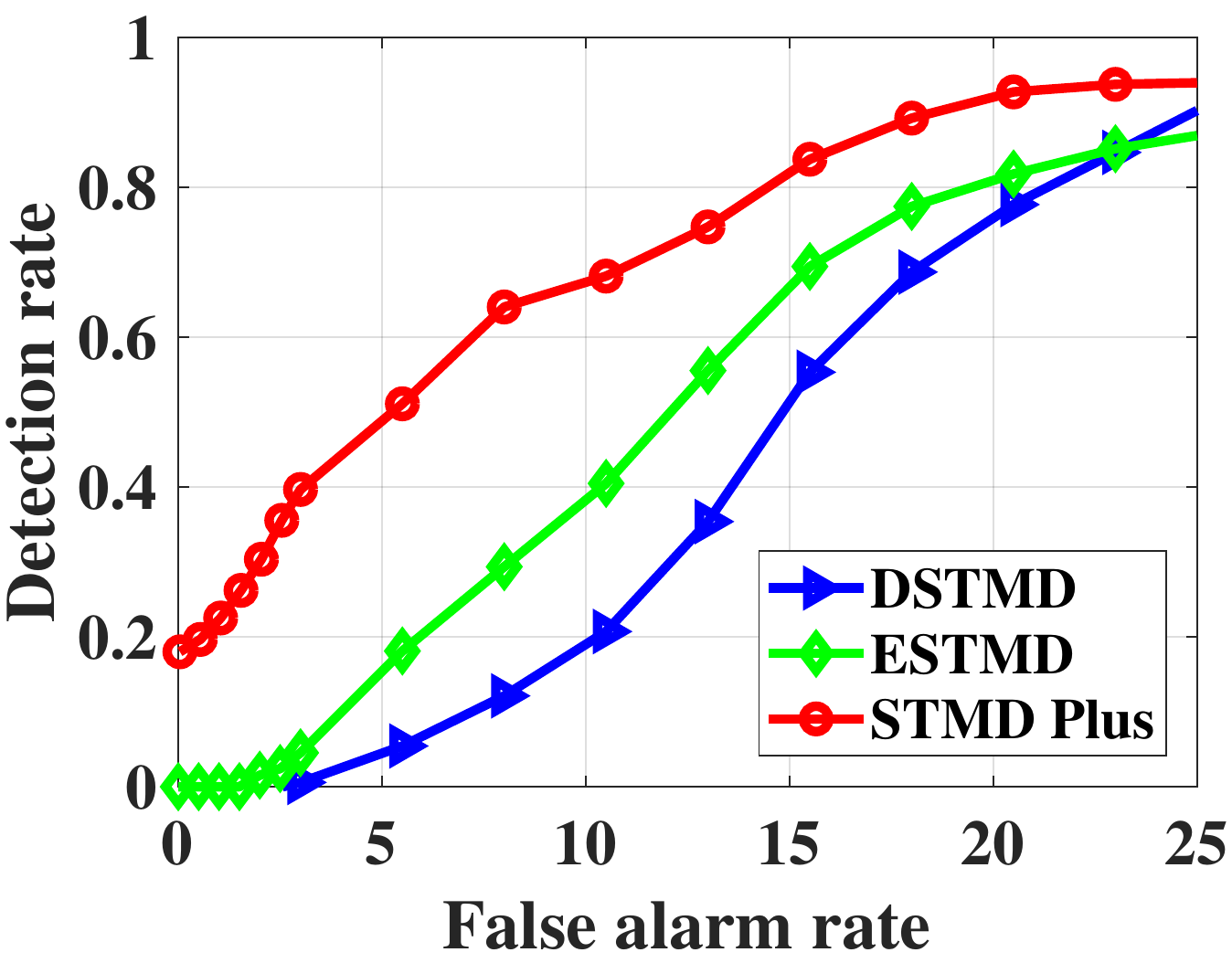}
		\label{Real-Dataset-STNS-16-ROC-Curve}}
	\hfil
	\subfloat[]{\includegraphics[width=0.275\textwidth]{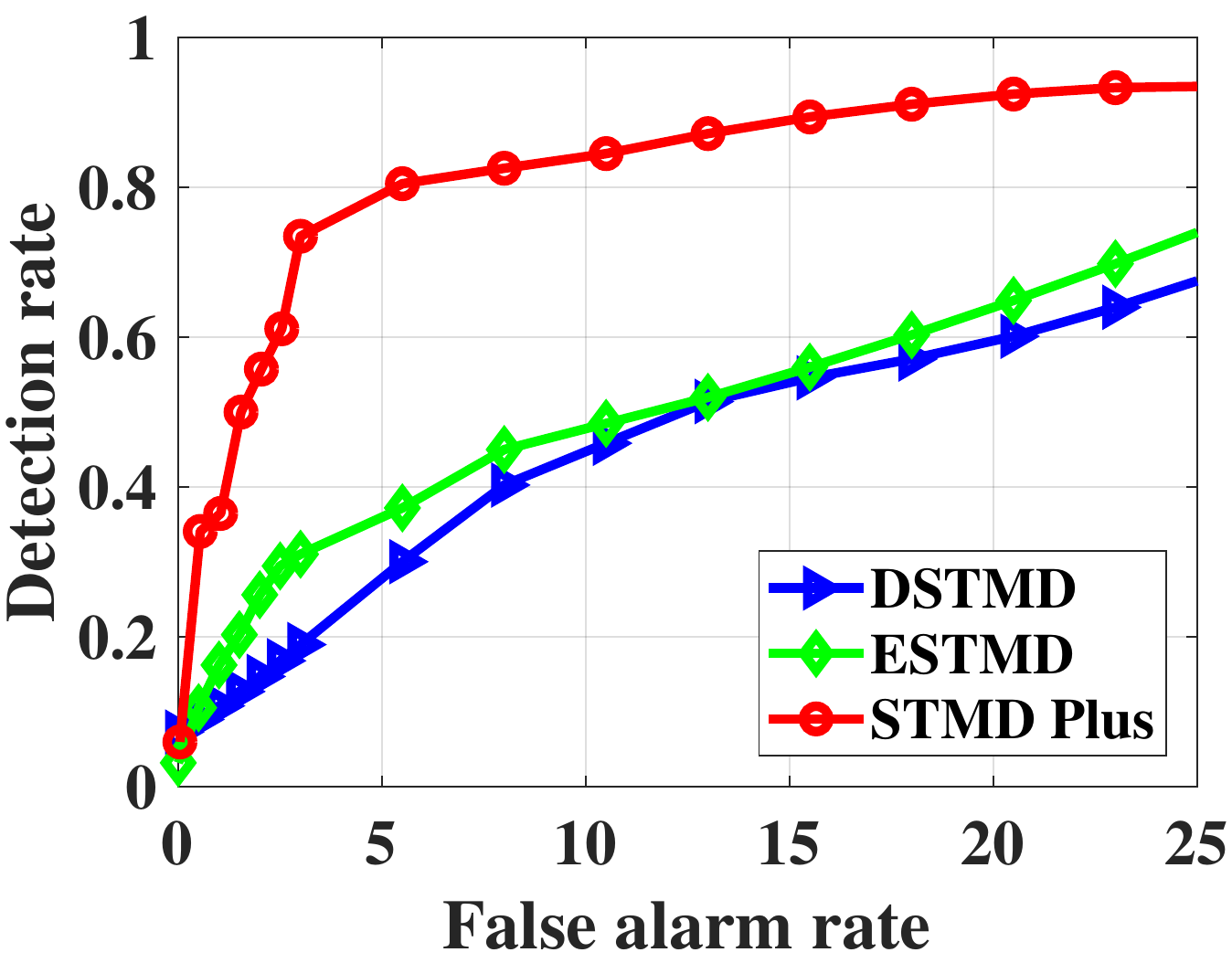}
		\label{Real-Dataset-STNS-18-ROC-Curve}}
	\hfil
	\subfloat[]{\includegraphics[width=0.275\textwidth]{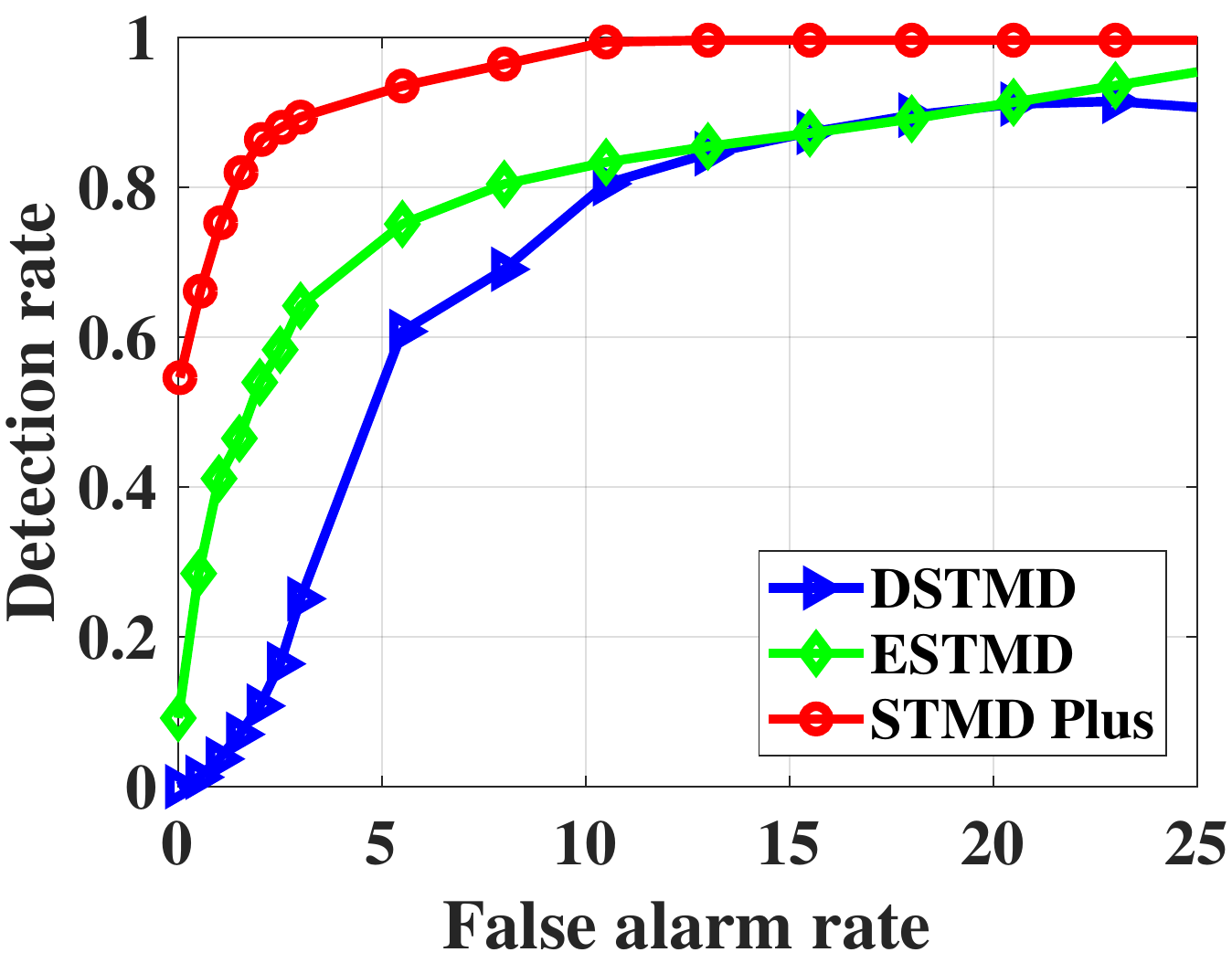}
		\label{Real-Dataset-STNS-22-ROC-Curve}}
	\hfil
	\subfloat[]{\includegraphics[width=0.275\textwidth]{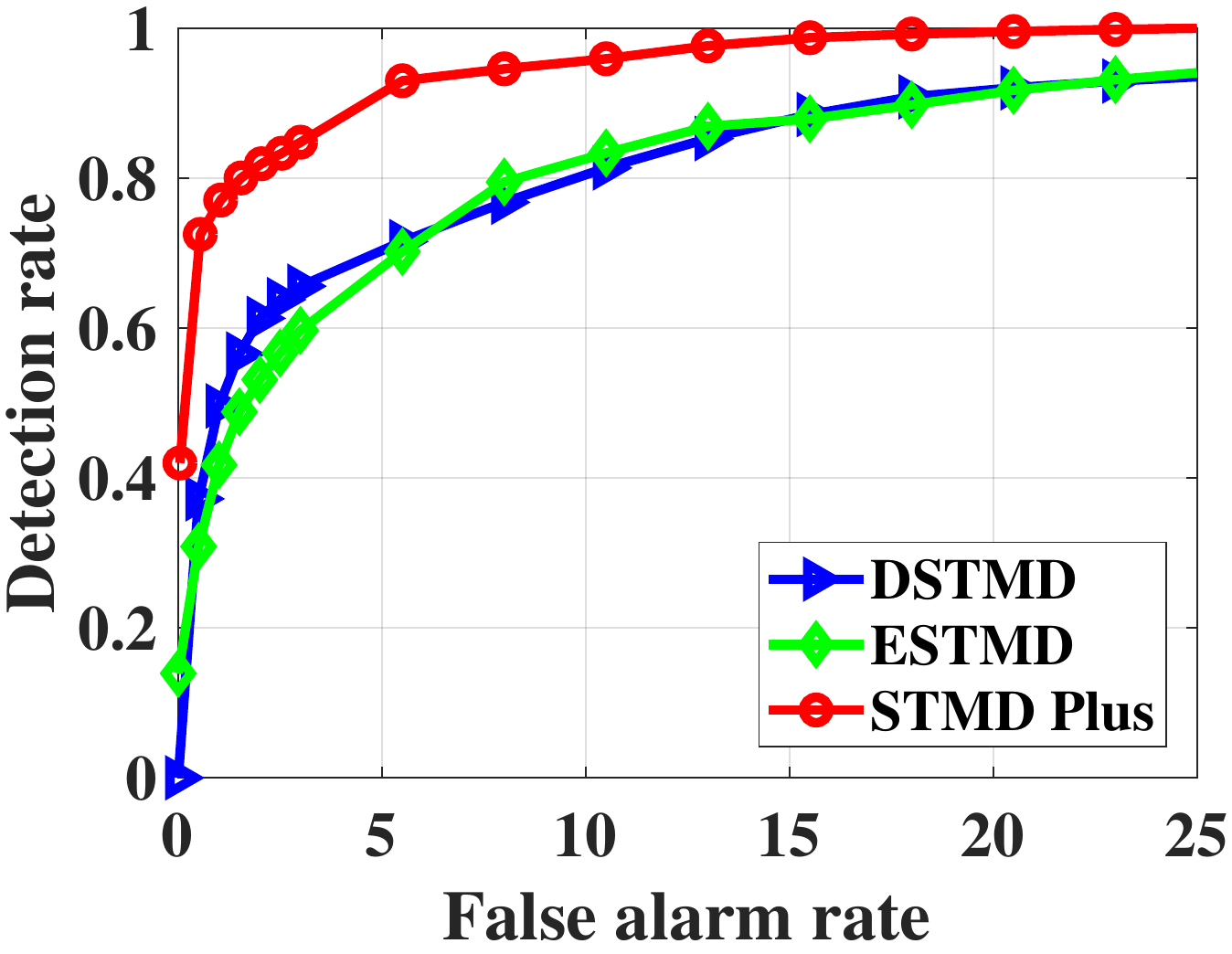}
		\label{Real-Dataset-STNS-25-ROC-Curve}}
	
	\caption{Receiver operating characteristic (ROC) curves of the three models for the six real image sequences. (a) Real image sequence $1$ (STNS-$4$). (b) Real image sequence $2$ (STNS-$15$). (c) Real image sequence $3$ (STNS-$16$). (d) Real image sequence $4$ (STNS-$18$). (e) Real image sequence $5$ (STNS-$22$). (f) Real image sequence $6$ (STNS-$25$).}
	\label{Real-Dataset-ROC-Curve}
\end{figure*}

Fig. \ref{Detection-Performance-Differnet-Backgrounds} presents the ROC curves of the three models for the Group $6$. As can be seen, the STMD+ outperforms the baseline models in different backgrounds. Note that the three models all perform well in Fig. \ref{Detection-Performance-Differnet-Backgrounds}(a). Their detection rates are all close to $1$ when the false alarm rates are low, and show small differences. This is because the background is much more homogeneous and contains less fake features. However, in more cluttered backgrounds such as Fig. \ref{Detection-Performance-Differnet-Backgrounds}(b) and (c), the STMD+ has a much better performance than the other two models.

We further tested the developed model on the publicly available STNS dataset \cite{bagheri2017performance}. Fig. \ref{Real-Dataset-ROC-Curve} illustrates the ROC curves of the three models for the six real image sequences, where the numbers of these six image sequences in the STNS dataset are $4$, $15$, $16$, $18$, $22$ and $25$, respectively. As it is shown in the six subplots, the detection rates of the STMD+ are higher than those of two baseline models when the false alarm rates are given. That is, the STMD+ obtains the best performance for all six real sequences, which means that the STMD+ can work more stably for different cluttered backgrounds and target types.

\section{Conclusion}
\label{Conclusion}
In this paper, we have proposed a visual system model (STMD+) for small target motion detection in cluttered backgrounds. The visual system contains two parallel information pathways and is capable of discriminating small targets from fake features. The first pathway called motion pathway, is intended to locate all small moving objects by calculating luminance changes over time at each pixel. The second pathway called contrast pathway, is designed to capture the directional contrast by computing luminance changes of each pixel along different directions. The mushroom body is introduced to fuse the two types of information from the two pathways. Finally, small targets are distinguished from fake features by comparing the standard deviations of the directional contrast on their motion traces. Comprehensive evaluation on the synthetic and real datasets, and comparisons with the existing STMD models demonstrate the effectiveness of the proposed visual system in filtering out fake features and improving detection rates. In the future, we will investigate the self-adaptability of the proposed visual system in different environments to further improve the robustness.

\section*{Appendix}

For demonstration of actual implementations, we attach pseudo-code form of the STMD Plus (see Algorithm \ref{alg:STMD_Plus}). We further briefly discuss the complexity of the proposed method for small target motion detection. As shown in Algorithm \ref{alg:STMD_Plus}, the computational time of our method mainly consists of four parts:  the ommatidia, the motion pathway, the contrast pathway and the mushroom body.

The computational complexity of the ommatidia is essentially determined by a 2-D spatial convolution of the input image with a Gaussian kernel (see Equation (\ref{Photoreceptors-Gaussian-Blur})), which can be implemented in $O(k^2mn)$ time for an $m \times n$ input image and a $k \times k$ kernel.

In the motion pathway, the LMC output can be regarded as the difference of two Gamma convolutions (see Equation (\ref{BPF-Para})-(\ref{LMCs-HPF})). Since the temporal Gamma convolution needs $O(lmn)$ cost where $l$ is the length of the Gamma kernel, the computational complexity of the LMC scales with $O(2lmn)$. Similarly, the total cost of the four medulla neurons is about $O(2lmn+2mn)$. 
According to (\ref{DSTMD-Signal-Correlation}), the computational complexity of the STMD is $O(2mn)$ for each preferred direction, so its entire cost grows like $O(2dmn)$ where $d$ denotes the number of the preferred directions. Finally, the lateral inhibition mechanism which is implemented by a 2-D convolution (see Equation (\ref{DS-STMD-Lateral-Inhibition})), needs $O(k^2mn)$ cost. Thus the entire computational complexity of the motion pathway is $O((k^2+4l+2d+2)mn)$.

In the contrast pathway, the directional contrast of each pixel along different spatial directions is calculated by convolving the ommatidium output with directional derivative operators (see Equation (\ref{T1-Output})). Since the 2-D spatial convolution needs $O(k^2mn)$ cost for each spatial direction, the entire computational complexity of the contrast pathway is $O(k^2dmn)$.

In the mushroom body, the nearest neighbor of each detected object is calculated for recording motion trace, which can be implemented in $O(p \log p)$ time \cite{mariello2018feature} where $p$ is the number of the detected objects. In addition, the cost of standard deviation calculation is around $O(rp)$ where $r$ represents the sample number. So the entire computational complexity of the mushroom body is around $O(p \log p +rp)$. 

Based on the aforementioned analysis, the entire computational complexity of the proposed STMD Plus is around $O(N(2k^2+k^2d+4l+2d+2)mn+N(\log p +r)p)$ where $N$ stands for the number of input images.

\begin{algorithm}[t!]
	\caption{Detection Process of the STMD Plus} 
	\label{alg:STMD_Plus}
	\begin{algorithmic}[1]
		\REQUIRE Image sequence $\{I_1,I_2,\cdots,I_N\}$, where $I_i \in \mathbb{R}^{m\times n}$.
		\ENSURE  Positions of small moving targets in each input image.
		\FOR {each input image}
		\STATE \COMMENT{Ommatidia}
		\STATE Calculate the output of the ommatidium via (\ref{Photoreceptors-Gaussian-Blur});
		\STATE \COMMENT{Motion Pathway}
		\STATE Calculate the output of the LMC via (\ref{LMCs-HPF});
		\STATE Calculate the outputs of the medulla neurons via (\ref{Tm3-Output})-(\ref{Tm1-Output});
		\STATE Calculate the output of the STMD via (\ref{DSTMD-Signal-Correlation});
		\STATE Calculate the laterally inhibited output  via (\ref{DS-STMD-Lateral-Inhibition}); 
		\STATE \COMMENT{Contrast Pathway}
		\STATE Calculate the output of the AMC via (\ref{AMC-Output});
		\STATE Calculate the output of the T1 neuron  via (\ref{T1-Output});
		\STATE \COMMENT{Mushroom Body}
		\STATE Calculate motion traces of the detected objects via (\ref{Motion-Trace});
		\FOR {each motion trace}
		\STATE Calculate the directional contrast of the motion trace via (\ref{T1-Neural-Output-Along-Target-Trace});
		\STATE Calculate the standard deviations ($SD$) of the directional contrast on the motion trace;
		\IF{$SD > \text{threshold}$}
		\STATE {the detected object is a small target};
		\ELSE \STATE {the detected object is a fake feature}.
		\ENDIF
		\ENDFOR
		\ENDFOR
	\end{algorithmic}
\end{algorithm}


\ifCLASSOPTIONcaptionsoff
  \newpage
\fi


\bibliographystyle{IEEEtran}

\bibliography{IEEEabrv,Reference}

\begin{thebibliography}{10}
\providecommand{\url}[1]{#1}
\csname url@samestyle\endcsname
\providecommand{\newblock}{\relax}
\providecommand{\bibinfo}[2]{#2}
\providecommand{\BIBentrySTDinterwordspacing}{\spaceskip=0pt\relax}
\providecommand{\BIBentryALTinterwordstretchfactor}{4}
\providecommand{\BIBentryALTinterwordspacing}{\spaceskip=\fontdimen2\font plus
\BIBentryALTinterwordstretchfactor\fontdimen3\font minus
  \fontdimen4\font\relax}
\providecommand{\BIBforeignlanguage}[2]{{%
\expandafter\ifx\csname l@#1\endcsname\relax
\typeout{** WARNING: IEEEtran.bst: No hyphenation pattern has been}%
\typeout{** loaded for the language `#1'. Using the pattern for}%
\typeout{** the default language instead.}%
\else
\language=\csname l@#1\endcsname
\fi
#2}}
\providecommand{\BIBdecl}{\relax}
\BIBdecl

\bibitem{yue2006bio}
S.~Yue, F.~C. Rind, M.~S. Keil, J.~Cuadri, and R.~Stafford, ``A bio-inspired
  visual collision detection mechanism for cars: Optimisation of a model of a
  locust neuron to a novel environment,'' \emph{Neurocomputing}, vol.~69,
  no.~13, pp. 1591--1598, Feb. 2006.

\bibitem{yue2006collision}
S.~Yue and F.~C. Rind, ``Collision detection in complex dynamic scenes using an
  lgmd-based visual neural network with feature enhancement,'' \emph{IEEE
  Trans. Neural Netw.}, vol.~17, no.~3, pp. 705--716, May 2006.

\bibitem{wozniak2018adaptive}
M.~Wo{\'z}niak and D.~Po{\l}ap, ``Adaptive neuro-heuristic hybrid model for
  fruit peel defects detection,'' \emph{Neural Netw.}, vol.~98, pp. 16--33,
  Feb. 2018.

\bibitem{Yan2018AFast}
C.~Yan, H.~Xie, J.~Chen, Z.~Zha, X.~Hao, Y.~Zhang, and Q.~Dai, ``A fast uyghur
  text detector for complex background images,'' \emph{IEEE Trans. Multimed.},
  vol.~20, no.~12, pp. 3389--3398, Dec. 2018.

\bibitem{yan2018effective}
C.~Yan, H.~Xie, S.~Liu, J.~Yin, Y.~Zhang, and Q.~Dai, ``Effective uyghur
  language text detection in complex background images for traffic prompt
  identification,'' \emph{IEEE Trans. Intell. Transp. Syst.}, vol.~19, no.~1,
  pp. 220--229, Jan. 2018.

\bibitem{yan2018supervised}
C.~Yan, H.~Xie, D.~Yang, J.~Yin, Y.~Zhang, and Q.~Dai, ``Supervised hash coding
  with deep neural network for environment perception of intelligent
  vehicles,'' \emph{IEEE Trans. Intell. Transp. Syst.}, vol.~19, no.~1, pp.
  284--295, Jan. 2018.

\bibitem{barnett2007retinotopic}
P.~D. Barnett, K.~Nordstr{\"o}m, and D.~C. O'Carroll, ``Retinotopic
  organization of small-field-target-detecting neurons in the insect visual
  system,'' \emph{Curr. Biol.}, vol.~17, no.~7, pp. 569--578, Apr. 2007.

\bibitem{mischiati2015internal}
M.~Mischiati, H.-T. Lin, P.~Herold, E.~Imler, R.~Olberg, and A.~Leonardo,
  ``Internal models direct dragonfly interception steering,'' \emph{Nature},
  vol. 517, no. 7534, pp. 333--338, Jan. 2015.

\bibitem{kelecs2017object}
M.~F. Kele{\c{s}} and M.~A. Frye, ``Object-detecting neurons in drosophila,''
  \emph{Curr. Biol.}, vol.~27, no.~5, pp. 680--687, Mar. 2017.

\bibitem{nordstrom2006insect}
K.~Nordstr{\"o}m, P.~D. Barnett, and D.~C. O'Carroll, ``Insect detection of
  small targets moving in visual clutter,'' \emph{PLoS Biol.}, vol.~4, no.~3,
  p. e54, Feb. 2006.

\bibitem{nordstrom2012neural}
K.~Nordstr{\"o}m, ``Neural specializations for small target detection in
  insects,'' \emph{Curr. Opin. Neurobiol.}, vol.~22, no.~2, pp. 272--278, Apr.
  2012.

\bibitem{wiederman2008model}
S.~D. Wiederman, P.~A. Shoemaker, and D.~C. O'Carroll, ``A model for the
  detection of moving targets in visual clutter inspired by insect
  physiology,'' \emph{PLoS One}, vol.~3, no.~7, pp. 1--11, Jul. 2008.

\bibitem{wang2018directionally}
H.~Wang, J.~Peng, and S.~Yue, ``A directionally selective small target motion
  detecting visual neural network in cluttered backgrounds,'' \emph{IEEE Trans.
  Cybern.}, to be published, doi: 10.1109/TCYB.2018.2869384.

\bibitem{freifeld2013gabaergic}
L.~Freifeld, D.~A. Clark, M.~J. Schnitzer, M.~A. Horowitz, and T.~R. Clandinin,
  ``Gabaergic lateral interactions tune the early stages of visual processing
  in drosophila,'' \emph{Neuron}, vol.~78, no.~6, pp. 1075--1089, Jun. 2013.

\bibitem{behnia2014processing}
R.~Behnia, D.~A. Clark, A.~G. Carter, T.~R. Clandinin, and C.~Desplan,
  ``Processing properties of on and off pathways for drosophila motion
  detection,'' \emph{Nature}, vol. 512, no. 7515, p. 427, Aug. 2014.

\bibitem{fu2018shaping}
Q.~Fu, C.~Hu, J.~Peng, and S.~Yue, ``Shaping the collision selectivity in a
  looming sensitive neuron model with parallel on and off pathways and spike
  frequency adaptation,'' \emph{Neural Netw.}, vol. 106, pp. 127--143, Oct.
  2018.

\bibitem{Hu2016A}
B.~Hu, S.~Yue, and Z.~Zhang, ``A rotational motion perception neural network
  based on asymmetric spatiotemporal visual information processing,''
  \emph{IEEE Trans. Neural Netw. Learn. Syst.}, vol.~28, no.~11, pp.
  2803--2821, Nov 2016.

\bibitem{yue2013redundant}
S.~Yue and F.~C. Rind, ``Redundant neural vision systems-competing for
  collision recognition roles,'' \emph{IEEE Trans. Auton. Mental Develop.},
  vol.~5, no.~2, pp. 173--186, Apr. 2013.

\bibitem{bagheri2017performance}
Z.~M. Bagheri, S.~D. Wiederman, B.~S. Cazzolato, S.~Grainger, and D.~C.
  O’Carroll, ``Performance of an insect-inspired target tracker in natural
  conditions,'' \emph{Bioinspir. \& Biomim.}, vol.~12, no.~2, p. 025006, Feb.
  2017.

\bibitem{St2016Adaptations}
A.~L. Stöckl, W.~A. Ribi, and E.~J. Warrant, ``Adaptations for nocturnal and
  diurnal vision in the hawkmoth lamina,'' \emph{J. Comp. Neurol.}, vol. 524,
  no.~1, p. 160, Jul. 2016.

\bibitem{Riveraalvidrez2011A}
Z.~Riveraalvidrez, I.~Lin, and C.~M. Higgins, ``A neuronally based model of
  contrast gain adaptation in fly motion vision,'' \emph{Visual Neurosci.},
  vol.~28, no.~5, p. 419, Aug. 2011.

\bibitem{lessios2018multiple}
N.~Lessios, R.~L. Rutowski, J.~H. Cohen, M.~E. Sayre, and N.~J. Strausfeld,
  ``Multiple spectral channels in branchiopods. i. vision in dim light and
  neural correlates,'' \emph{J. Exp. Biol.}, pp. jeb--165\,860, Apr. 2018.

\bibitem{lee2015spatio}
Y.-J. Lee, H.~O. J{\"o}nsson, and K.~Nordstr{\"o}m, ``Spatio-temporal dynamics
  of impulse responses to figure motion in optic flow neurons,'' \emph{PLoS
  One}, vol.~10, no.~5, pp. 1--16, May 2015.

\bibitem{li2017local}
J.~Li, J.~Lindemann, and M.~Egelhaaf, ``Local motion adaptation enhances the
  representation of spatial structure at emd arrays,'' \emph{PLOS Comput.
  Biol.}, vol.~13, no.~12, pp. 1--23, Dec. 2017.

\bibitem{wiederman2013biologically}
S.~D. Wiederman and D.~C. O’Carroll, ``Biologically inspired feature
  detection using cascaded correlations of off and on channels,'' \emph{J.
  Artif. Intell. Soft Comput. Res.}, vol.~3, no.~1, pp. 5--14, Dec. 2013.

\bibitem{hassenstein1956systemtheoretische}
B.~Hassenstein and W.~Reichardt, ``Systemtheoretische analyse der zeit-,
  reihenfolgen-und vorzeichenauswertung bei der bewegungsperzeption des
  r{\"u}sselk{\"a}fers chlorophanus,'' \emph{Zeitschrift f{\"u}r Naturforschung
  B}, vol.~11, no. 9-10, pp. 513--524, Oct. 1956.

\bibitem{eichner2011internal}
H.~Eichner, M.~Joesch, B.~Schnell, D.~F. Reiff, and A.~Borst, ``Internal
  structure of the fly elementary motion detector,'' \emph{Neuron}, vol.~70,
  no.~6, pp. 1155--1164, Jun. 2011.

\bibitem{wang2018improved}
H.~Wang, J.~Peng, and S.~Yue, ``An improved lptc neural model for background
  motion direction estimation,'' in \emph{Proc. IEEE Conf. ICDL-EpiRob}, 2017,
  pp. 47--52.

\bibitem{clark2011defining}
D.~A. Clark, L.~Bursztyn, M.~A. Horowitz, M.~J. Schnitzer, and T.~R. Clandinin,
  ``Defining the computational structure of the motion detector in
  drosophila,'' \emph{Neuron}, vol.~70, no.~6, pp. 1165--1177, Jun. 2011.

\bibitem{fortun2015optical}
D.~Fortun, P.~Bouthemy, and C.~Kervrann, ``Optical flow modeling and
  computation: a survey,'' \emph{Comput. Vis. Image Underst.}, vol. 134, pp.
  1--21, May 2015.

\bibitem{yong2017robust}
H.~Yong, D.~Meng, W.~Zuo, and L.~Zhang, ``Robust online matrix factorization
  for dynamic background subtraction,'' \emph{IEEE Trans. Pattern Anal. Mach.
  Intell.}, vol.~40, no.~7, pp. 1726--1740, Jul. 2017.

\bibitem{li2016rotation}
Z.~Li, G.~Zhao, S.~Li, H.~Sun, R.~Tao, X.~Huang, and Y.~J. Guo, ``Rotation
  feature extraction for moving targets based on temporal differencing and
  image edge detection.'' \emph{IEEE Geosci. Remote Sens. Lett.}, vol.~13,
  no.~10, pp. 1512--1516, Oct. 2016.

\bibitem{ren2003motion}
Y.~Ren, C.-S. Chua, and Y.-K. Ho, ``Motion detection with nonstationary
  background,'' \emph{Mach. Vis. Appl.}, vol.~13, no. 5-6, pp. 332--343, Mar.
  2003.

\bibitem{gao2013infrared}
C.~Gao, D.~Meng, Y.~Yang, Y.~Wang, X.~Zhou, and A.~G. Hauptmann, ``Infrared
  patch-image model for small target detection in a single image,'' \emph{IEEE
  Trans. Image Process.}, vol.~22, no.~12, pp. 4996--5009, Sep. 2013.

\bibitem{wei2016multiscale}
Y.~Wei, X.~You, and H.~Li, ``Multiscale patch-based contrast measure for small
  infrared target detection,'' \emph{Pattern Recognit.}, vol.~58, pp. 216--226,
  Oct. 2016.

\bibitem{bai2018derivative}
X.~Bai and Y.~Bi, ``Derivative entropy-based contrast measure for infrared
  small-target detection,'' \emph{IEEE Trans. Geosci. Remote Sens.}, vol.~56,
  no.~4, pp. 2452--2466, Apr. 2018.

\bibitem{hu2016bio}
C.~Hu, F.~Arvin, C.~Xiong, and S.~Yue, ``Bio-inspired embedded vision system
  for autonomous micro-robots: the lgmd case,'' \emph{IEEE Trans. Cogn.
  Develop. Syst.}, vol.~9, no.~3, pp. 241--254, Sep. 2016.

\bibitem{indiveri2000neuromorphic}
G.~Indiveri and R.~Douglas, ``Neuromorphic vision sensors,'' \emph{Science},
  vol. 288, no. 5469, pp. 1189--1190, May 2000.

\bibitem{rind2003locust}
F.~C. Rind, R.~D. Santer, J.~M. Blanchard, and P.~F. Verschure, ``Locust’s
  looming detectors for robot sensors,'' in \emph{Sensors and sensing in
  biology and engineering}.\hskip 1em plus 0.5em minus 0.4em\relax Springer,
  2003, pp. 237--250.

\bibitem{he2018adaptive}
W.~He and Y.~Dong, ``Adaptive fuzzy neural network control for a constrained
  robot using impedance learning,'' \emph{IEEE Trans. Neural Netw. Learn.
  Syst.}, vol.~29, no.~4, pp. 1174--1186, Apr. 2018.

\bibitem{polap2018multi}
D.~Po{\l}ap, M.~Wo{\'z}niak, W.~Wei, and R.~Dama{\v{s}}evi{\v{c}}ius,
  ``Multi-threaded learning control mechanism for neural networks,''
  \emph{Future Gener. Comput. Syst.}, vol.~87, pp. 16--34, Oct. 2018.

\bibitem{ardin2016using}
P.~Ardin, F.~Peng, M.~Mangan, K.~Lagogiannis, and B.~Webb, ``Using an insect
  mushroom body circuit to encode route memory in complex natural
  environments,'' \emph{PLoS Comput. Biol.}, vol.~12, no.~2, pp. 1--12, Feb.
  2016.

\bibitem{webb2016neural}
B.~Webb and A.~Wystrach, ``Neural mechanisms of insect navigation,''
  \emph{Curr. Opin. Insect Sci.}, vol.~15, pp. 27--39, Jun. 2016.

\bibitem{warrant2016matched}
E.~J. Warrant, ``Matched filtering and the ecology of vision in insects,'' in
  \emph{The Ecology of Animal Senses}.\hskip 1em plus 0.5em minus 0.4em\relax
  Springer, Dec. 2016, pp. 143--167.

\bibitem{takemura2013visual}
S.-y. Takemura, A.~Bharioke, Z.~Lu, A.~Nern, S.~Vitaladevuni, P.~K. Rivlin,
  W.~T. Katz, D.~J. Olbris, S.~M. Plaza, P.~Winston \emph{et~al.}, ``A visual
  motion detection circuit suggested by drosophila connectomics,''
  \emph{Nature}, vol. 500, no. 7461, p. 175, Aug. 2013.

\bibitem{behnia2015visual}
R.~Behnia and C.~Desplan, ``Visual circuits in flies: beginning to see the
  whole picture,'' \emph{Curr. Opin. Neurobiol.}, vol.~34, pp. 125--132, Oct.
  2015.

\bibitem{de1991theory}
B.~De~Vries and J.~C. Pr{\'\i}ncipe, ``A theory for neural networks with time
  delays,'' in \emph{Proc. NIPS}, 1990, pp. 162--168.

\bibitem{yamaguchi2011photoreceptors}
S.~Yamaguchi and M.~Heisenberg, ``Photoreceptors and neural circuitry
  underlying phototaxis in insects,'' \emph{Fly}, vol.~5, no.~4, pp. 333--336,
  Dec. 2011.

\bibitem{rogers2015differential}
S.~M. Rogers and S.~R. Ott, ``Differential activation of serotonergic neurons
  during short-and long-term gregarization of desert locusts,'' \emph{Proc.
  Royal Soc. B}, vol. 282, no. 1800, p. 20142062, Feb. 2015.

\bibitem{freeman1991design}
W.~T. Freeman, E.~H. Adelson \emph{et~al.}, ``The design and use of steerable
  filters,'' \emph{IEEE Trans. Pattern Anal. Mach. Intell.}, vol.~13, no.~9,
  pp. 891--906, Sep. 1991.

\bibitem{zhang2015contour}
W.-C. Zhang and P.-L. Shui, ``Contour-based corner detection via angle
  difference of principal directions of anisotropic gaussian directional
  derivatives,'' \emph{Pattern Recognit.}, vol.~48, no.~9, pp. 2785--2797, Sep.
  2015.

\bibitem{straw2008vision}
A.~D. Straw, ``Vision egg: an open-source library for realtime visual stimulus
  generation.'' \emph{Front. Neuroinf.}, vol.~2, no.~4, Nov. 2008.

\bibitem{mariello2018feature}
A.~Mariello and R.~Battiti, ``Feature selection based on the neighborhood
  entropy,'' \emph{IEEE Trans. Neural Netw. Learn. Syst.}, vol.~29, no.~12, pp.
  6313--6322, Dec. 2018.

\end{thebibliography}

\end{document}